\documentclass[10pt,twocolumn,letterpaper]{article}

\usepackage[final,pagenumbers]{cvpr} %

\usepackage{color}
\usepackage[normalem]{ulem}
\usepackage{tabularray}
\definecolor{Gray}{rgb}{0.501,0.501,0.501}
\usepackage[accsupp]{axessibility}

\usepackage{soul}
\def\X#1{
        \raisebox{.9pt}{\textcircled{\raisebox{-.9pt}{#1}}}%
}
\usepackage{array}
\usepackage{graphicx}
\usepackage[frozencache,cachedir=minted-cache]{minted}
\usepackage[dvipsnames]{xcolor}

\newcommand{\rev}[1]{{#1}}
\newcommand{\revv}[1]{{#1}}

\usepackage{amsmath}
\usepackage{amssymb}
\usepackage{booktabs}
\usepackage{tabularray}
\usepackage{graphicx}
\usepackage{fancyhdr}
\usepackage{float}
\usepackage{multicol}
\usepackage{multirow}
\usepackage{makecell}
\usepackage{ulem}
\usepackage{xcolor}
\usepackage{bm}
\usepackage[pagebackref=true,breaklinks=true,letterpaper=true,colorlinks,citecolor=citecolor,bookmarks=false]{hyperref}
\usepackage{textpos}
\usepackage[table]{colortbl}
\definecolor{citecolor}{RGB}{30,102,235}

\newcommand{\MYhref}[3][blue]{\href{#2}{\color{#1}{#3}}}%

\usepackage[capitalize]{cleveref}
\crefname{section}{Sec.}{Secs.}
\Crefname{section}{Section}{Sections}
\Crefname{table}{Table}{Tables}
\crefname{table}{Tab.}{Tabs.}

\def\paperID{10962} 
\def\confName{CVPR}
\def\confYear{2024}
\begin{document}

\newcommand{\sx}[1]{\textcolor{cyan}{sx: #1}}
\newcommand{\bp}[1]{\textcolor{cyan}{bp: #1}}
\definecolor{deemph}{gray}{0.6}
\newcommand{\gb}{\rowcolor{gray!20}}

\newtheorem{lemma}{Lemma}
\newtheorem{sublemma}{Lemma}[lemma]

\newcommand\blfootnote[1]{\begingroup\renewcommand\thefootnote{}\footnote{#1}\addtocounter{footnote}{-1}\endgroup}

\title{Brain Decodes Deep Nets}

\author{%
  \textbf{Huzheng Yang} \quad \textbf{{James Gee}*} \quad \textbf{{Jianbo Shi}*}\\
  University of Pennsylvania\\
\href{https://huzeyann.github.io/brain-decodes-deep-nets}{https://huzeyann.github.io/brain-decodes-deep-nets}
}

\twocolumn[{%
\renewcommand\twocolumn[1][]{#1}%
\maketitle
\vspace{-8mm}
\begin{center}
    \centering
    \captionsetup{type=figure}
    \includegraphics[width=\linewidth]{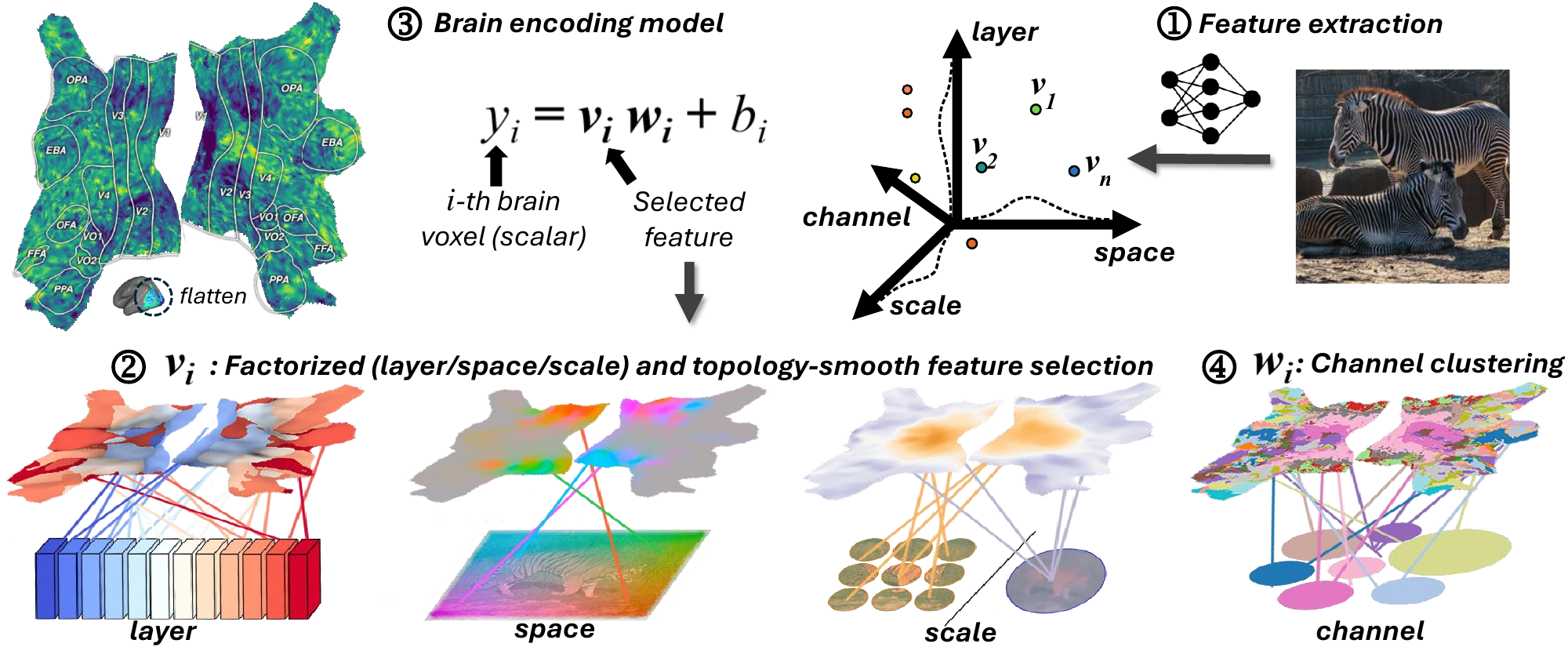}
    \vspace{-6mm}
    \captionof{figure}{\rev{\textbf{{Visualize Deep Networks in the Brain}}. The training objective of the brain encoding model is to predict the brain’s fMRI signal in response to an image stimulus.  3D visual brain surface is flattened into 2D for better visualization. \X1  Image features are extracted from a pre-trained network. \X2 Feature selection for each voxel is randomly initialized and learned using the brain encoding training objective. The selection is \textbf{\textit{factorized}} in the layer/space/scale axis; the \textbf{\textit{topological constraint}} improves selection smoothness and confidence. \X3 Linearized brain encoding model. \X4 After training, linear weights are used to cluster channels. We use the  resulting brain-to-network mapping together with the known knowledge of the brain to answer the question \textit{``how do deep networks work?''.}
    }
    }\label{fig:one}
    \vspace{-1mm}
\end{center}%
}]

\vspace{-1mm}
\begin{abstract}
\vspace{-5mm}
We developed a tool for visualizing and analyzing large pre-trained vision models by mapping them onto the brain, thus exposing their hidden inside. Our innovation arises from a surprising usage of brain encoding: predicting brain fMRI measurements in response to images. We report two findings. First, explicit mapping between the brain and deep-network features across dimensions of space, layers, scales, and channels is crucial. This mapping method, FactorTopy, is plug-and-play for any deep-network; with it, one can paint a picture of the network onto the brain (literally!). Second, our visualization shows how different training methods matter: they lead to remarkable differences in hierarchical organization and scaling behavior, growing with more data or network capacity. It also provides insight into fine-tuning: how pre-trained models change when adapting to small datasets. 
\rev{We found brain-like hierarchically organized network suffer less from catastrophic forgetting after fine-tuned.}
\end{abstract}


\vspace{-5mm}
\section{Introduction}
\label{sec:intro}


\blfootnote{*: Equal advising.}

The brain is massive, and its enormous size hides within it a mystery: how it efficiently organizes many specialized modules with distributed representation and control. One clue it offers is its feed-forward hierarchical organization \rev{(\Cref{fig:background})}.  This hierarchical structure facilitates efficient computation, continuous learning, and adaptation to dynamic tasks. 

Deep networks are enormous, containing billions of parameters.   Performances keep improving with more training data and larger size.  It doesn't seem to matter if the network is trained under the supervision of labels, weakly supervised with image captions, or even self-supervised without human-provided guidance.   Its sheer size also hides another mystery: as its size increases, it can be fine-tuned successfully to many unseen tasks.    

\emph{What can these two massive systems, the brain and deep network, tell about each other?} \revv{By identifying `what' deep features are most relevant for each brain voxel fMRI prediction, we can obtain a picture of deep features mapped onto a brain (literally), as shown by the brain-to-network mapping in Figure \ref{fig:one}.  }



\rev{The key insight is that deep networks trained with the same architecture, but different objectives and data, produce drastically different computation layouts of intermediate layers, even if they can produce similar brain encoding scores and other downstream task scores.  For example, we found intermediate layers of CLIP align hierarchically to the visual brain. However, there are unexpected non-hierarchical bottom-up and top-down structure in supervised classification and segmentation-trained models. Moreover, for many models, when scaling up in parameters and training data, they tend to lose hierarchical alignment to the brain, except CLIP, which improved hierarchical alignment to the brain after scaling up.}


\revv{Suppose the brain's hierarchical organization is a template for efficient, modular, and generalizable computation; an ideal computer vision model should align with the brain: the first layer of the deep network matches the early visual cortex, and the last layer best matches high-level regions. Our fine-tuning results show that networks with more hierarchy organization tend to (qualitatively) maintain their hidden layers better after fine-tuning on small datasets, thus suffering less (quantitatively) from catastrophic forgetting. We conjecture that better alignment to the brain is one way to find a robust model that adapts to dynamic tasks and scales better with larger models and more data.}

Our analysis crucially depends on a robust mapping between deep 4D features: spatial, layer, channel, and scale (class token vs local token) to the brain.  Our fundamental assumption is that this mapping should be: a) \emph{brain-topology constrained}, and b) \emph{factorized} in feature dimensions of space, layer, channel, and scale.
This is important because independent 4D image features to brain mapping are highly unconstrained, and learning a shared mapping across images, with brain-topology constraint and factorized representation, is statistically more stable.  

  
\blfootnote{$^1$: The Algonauts 2023 competition: \MYhref{http://algonauts.csail.mit.edu/}{http://algonauts.csail.mit.edu/}} 

\vspace{-4mm}
Our contribution is summarized as the following:
\begin{enumerate}[noitemsep, nolistsep]
    \item We introduce a \emph{factorized, brain-topological smooth} selection that produces an explicit mapping between deep features: space, layer, channel, and scale (class token vs local token) to the brain.
    \item We pioneer a new network visualization by coloring the brain using layer-selectors, exposing the inner workings of the network.
    \item \rev{We found that brain-like hierarchically organized networks suffer less from catastrophic forgetting after fine-tuning.}
\end{enumerate}

\section{Background and Related Work}

\begin{figure}[h]
    \centering
    \vspace{-2mm}
    \includegraphics[width=\linewidth]{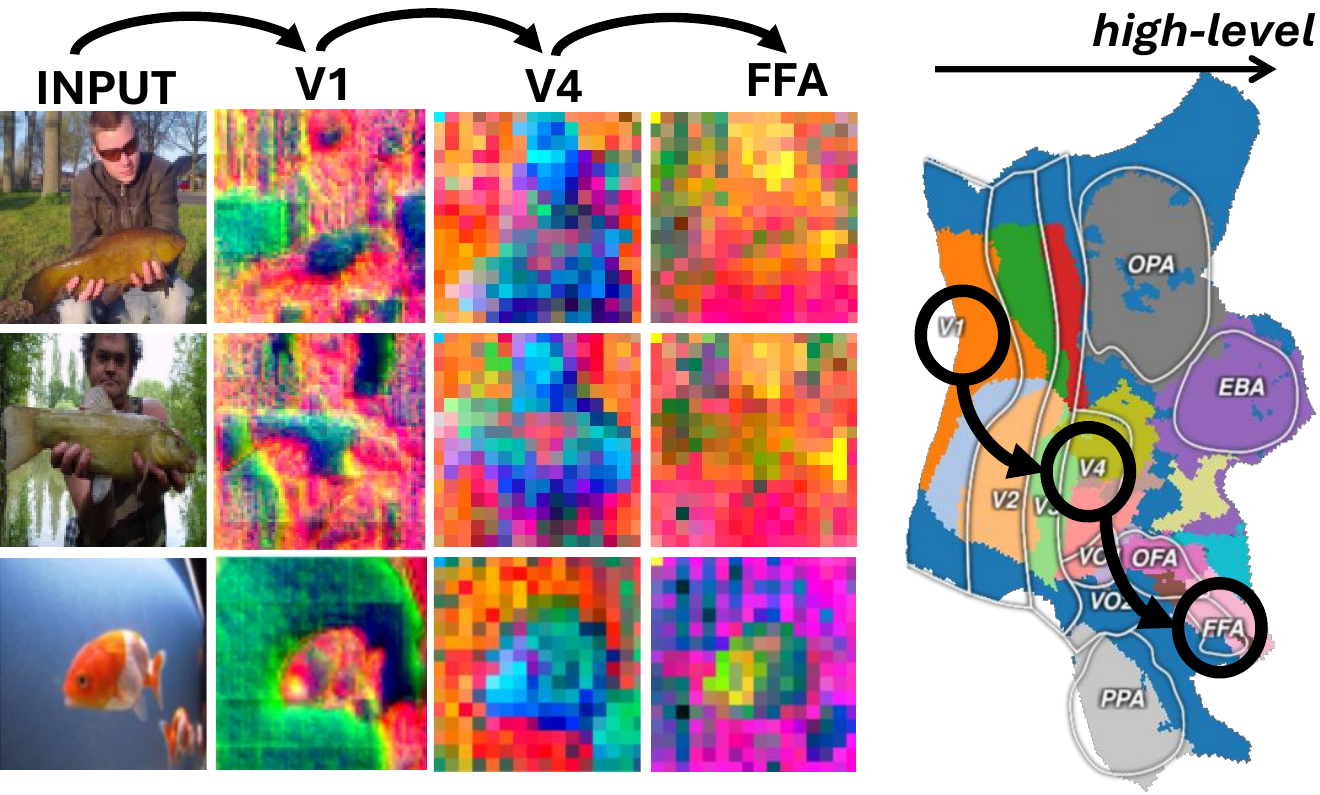}
    \vspace{-6mm}
    \caption{\textbf{Image features (selected channels) for brain ROIs}. V1 is orientation filtering, V4 segmentation, FFA face-selective. }
    \label{fig:background}
    \vspace{-4mm}
\end{figure}

\paragraph{Hierarchy of the Visual Brain} 
In Figure \ref{fig:background}, visual brain is organized into regions, each region has specialized functions. Image processing in visual brain is organized in a hierarchical and feed-forward fashion. Starting from region V1 to V4, neurons were found to have increasing receptive field size and represent more abstract concepts \cite{dumoulin_population_2008, dicarlo_how_2012, yamins_using_2016}, the late visual brain has semantic regions such as face (FFA), body (EBA), and place (OPA, PPA).

\vspace{-4mm}
\paragraph{Brain Encoding Benchmarks} 
Open challenge and competitions on brain encoding model have generated broad interests \cite{gifford_algonauts_2023, cichy_algonauts_2021, cichy_algonauts_2019, schrimpf_brain-score_2018, schrimpf_integrative_2020, willeke_sensorium_2022, turishcheva_dynamic_2023}. Large-scale open-source datasets are growing rapidly in both quantity and quality \cite{lahner_bold_2023, chang_bold5000_2019, allen_massive_2022, hebart_things-data_2023, gifford_large_2022}. The Algonauts\textcolor{red}{$^1$} 2023 competition \cite{gifford_algonauts_2023} is the first to use a massive high-quality 7-Tesla fMRI dataset \cite{allen_massive_2022}.
The high-quality and large-scale of this datasets enabled models that can recover brain-to-space mapping from naturalistic image stimuli \cite{roth_natural_2022}, which was only possible with synthetic stimuli \cite{dumoulin_population_2008}. \rev{Our brain encoding model methods is a direct extension of the Algonauts 2023 competition winning method \textit{Memory Encoding Model} \cite{yang_memory_2023}. In this work, we added a scale axis for feature selection.}

\vspace{-4mm}
\paragraph{Explain Brain by Deep Networks} 
After fitting brain encoding models to predict brain response, gradient-based methods have been used to explain how brain works: orientation-selective neurons in V1 \cite{roth_natural_2022, franke_state-dependent_2022, lurz_generalization_2021}, category-selective regions in late visual brain \cite{ratan_murty_computational_2021, jain_selectivity_2023, sarch_brain_2023, luo_brainscuba_2023, prince_contrastive_2023, luo_neural_2023}. Gradient-based methods can also generate maximum-excited images \cite{bashivan_neural_2019, gu_neurogen_2022, kneeland_second_2023, takagi_high-resolution_2022, yamins_performance-optimized_2014}. Meanwhile, studies try to find the best performance pre-trained model for each brain ROI \cite{conwell_what_2023, zhuang_unsupervised_2021, oconnell_approaching_2023, sarch_3d_2023, wang_incorporating_2022} from a zoo of supervised \cite{radford_learning_2021, singh_revisiting_2022, kirillov_segment_2023, jain_bottom_2022}, self-supervised \cite{he_masked_2021, chen_empirical_2021, oquab_dinov2_2023, gupta_patchgame_2021, lal_coconets_2021}, image generation \cite{rombach_high-resolution_2022}, and 3D \cite{mildenhall_nerf_2020, sarch_tidee_2022, tung_learning_2019} models. Features can be efficiently cached and are plug-in-and-play \cite{walmer_teaching_2023, gwilliam_beyond_2022, taylor_extracting_2023}. Different from the main-stream study that use deep networks to explain the brain's functionality. In this work, we use existing knowledge of the brain's functionality to explain feature computation in deep networks.

\begin{figure*}[!ht]
    \vspace{-8mm}
    \centering
    \captionsetup{type=figure}
    \includegraphics[width=0.99\linewidth]{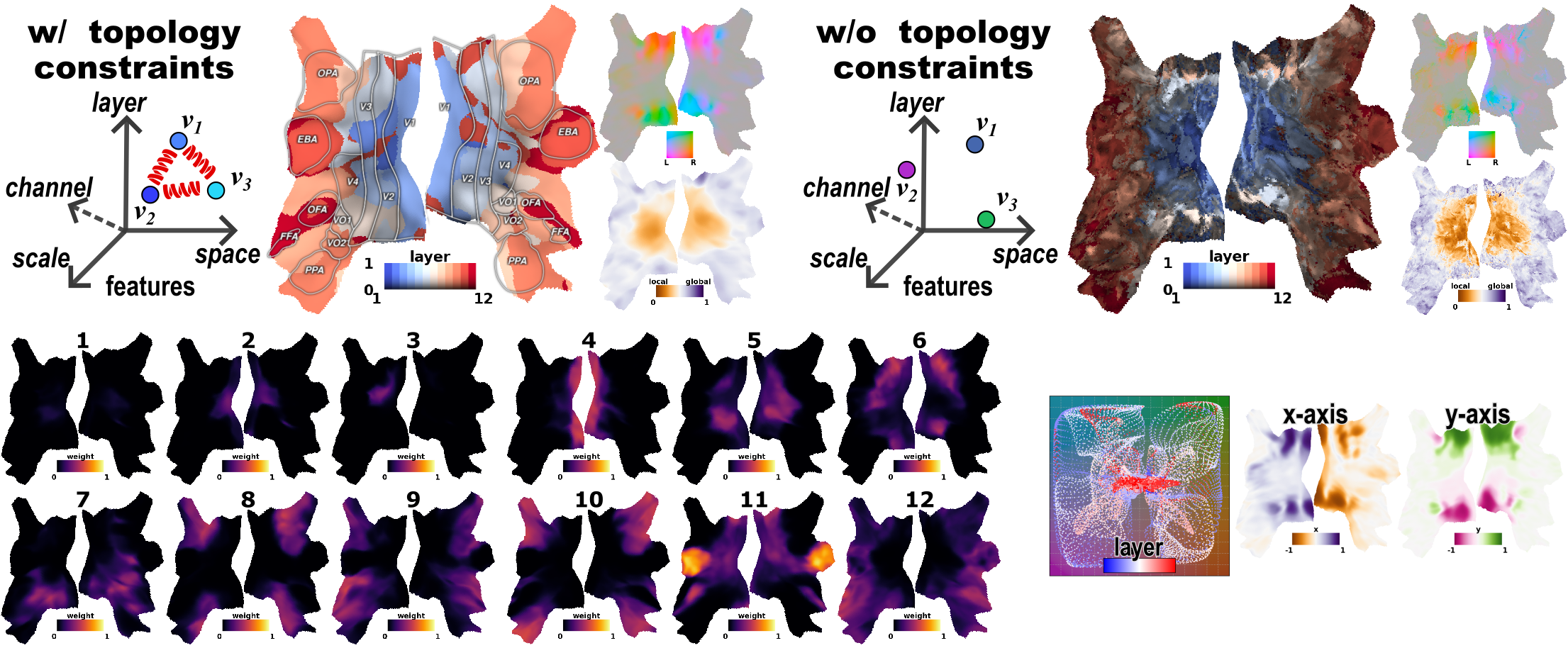}
    \vspace{-3mm}
    \captionof{figure}{\textbf{Topological Constrained, Factorized, Brain-to-Network Selectors for CLIP}. \textbf{\textit{Top}}: factorized-selectors trained with topological constraints improved confidence of the mapping (color brightness) and mapping smoothness (colored as Section \ref{sec:color}). \textbf{\textit{Bottom left}}: individual layer-selector weight $\bm{\hat{\omega}}^{layer}$, note layer 4 is mostly aligned with V1, and the last two are aligned with the body (EBA) and face (FFA) region. \textit{\textbf{Bottom right}}: space-selector $\bm{\hat{u}}^{space}$:
    3D voxels, dots, are mapped to the image space with color dots indicating the layers.  For later layers, only center image regions are selected.   }\label{fig:selectors}
    \vspace{-2mm}
\end{figure*}

\section{Methods: Brain Encoding Model}

Figure \ref{fig:one} presents an overview of our methods.  In the brain encoding task, one needs to predict a large number voxels (vertices), of visual cortex's fMRI responses as a function of the observed image.   This encoding task is under-constrained: since each subject has her/his unique mental process, a successful brain encoding model needs to be highly individualized, thus significantly reducing the training example per voxel.  Most of the current approaches treat each brain voxel independently.  This leads to a major reduction in signal-to-noise ratio, particularly for our analysis. 

Our fundamental innovations are two-fold.  First, we enforce brain-and-network \emph{\textbf{topology-constrained}} prediction.   Brain voxels are not independent but are organized locally into similar ``tasks'', and globally into diverse functional regions.  Similarly, Neural networks show local feature similarity across adjacent layers while ensuring diversity for far-away ones.  The local smoothness constraints significantly reduce uncertainties in network-to-brain mapping.  

Second, we propose a \emph{\textbf{factorized}} feature selection across three independent dimensions of space, layers, and scales (local vs global token).  This factorized representation leads to a more robust estimation because feature selection in each dimension is more straightforward, and learning can be more efficient across training samples.   For example, the spatial feature selection only needs to find the center of the pixel region for each brain voxel, similar to retinotopy.  The layer or scale selection estimates the size of the pixel region: the early layer typically has a smaller receptive field size. Note that the factorized feature selection is \emph{soft}: multiple layers or spatial locations can be selected, as determined by the brain prediction training target. 




\subsection{Factorized, Topological Smooth, Brain-to-network Selection (FactorTopy)}

We used a pre-trained image backbone model (ViT) to process input image $\bm{X}$ into features $\bm{V}$. The entire feature $\bm{V}$ is organized along four dimensions: space, layer, scale (class token and local tokens), and channels.  



The current state-of-the-art methods \cite{allen_massive_2022} compute a layer-specific, scale-specific, 2D spatial feature selection mask to pool features $\bm{V}  \in \mathbb{R}^{ L\times C \times H \times W}$ along spatial dimension $H\times W$ into a vector of $\mathbb{R}^{L \times C}$, where $L$ denotes layer and $C$ is channel.   Instead, we propose a \textit{\textbf{factorized}} feature selection method where, for each voxel, we select the corresponding space, layer, and scale in each dimension.

Essentially, a voxel asks: `What is the best x-factor for my brain prediction?' where the x-factor is one of the layer, space, scale, or channel dimensions.


\noindent
\textbf{1) space selector}.  $selSpace: \enspace  \mathbb{R}^{N \times 3} \rightarrow \bm{\hat{u}}^{space} \in \mathbb{R}^{N \times 2}$, maps brain voxels' \rev{3D coordinates} into 2D image coordinates, \rev{where $N$ is number of voxels.}  We used linear interpolation $\texttt{Interp}$ to extract $\bm{\bar{\nu}}_{i,l}^{local} \in \mathbb{R}^{1 \times C}$.

\noindent
\textbf{2) layer selector}. $selLayer: \enspace  \mathbb{R}^{N \times 3} \rightarrow \bm{\hat{\omega}}^{layer} \in \mathbb{R}^{N \times L}$, produces $\hat{\omega}_{i,l}^{layer} \in [0, 1]$ weight for each layer $l$, such that $\sum_{l=1}^{L}\hat{\omega}_{i,l}^{layer} = 1$.  We take a weighted channel-wise average of feature vectors $\bm{\bar{\nu}}_{i,l}$ across all layers. 

\noindent
\textbf{3) scale selector}: $selScale: \enspace  \mathbb{R}^{N \times 3} \rightarrow \bm{\hat{\alpha}}^{scale} \in \mathbb{R}^{N \times 1}$, computes a scalar $\hat{\alpha}_i^{scale} \in [0, 1]$ as the weight for local $\bm{\bar{\nu}}_{i,l}^{local}$ vs global token $\bm{\bar{\nu}}_{*,l}^{global}$. Note that $\bm{\bar{\nu}}^{local}_{i,l}$ is unique for each voxel, $\bm{\bar{\nu}}_{*,l}^{global}$ is same for all voxels.


Taking weighted averages over channels across layers could be problematic because channels in each layer represent different information. We need to preemptively align the channels into a shared $D$ dimension channel space.  Let $B_l$ be 
a layer-unique channel transformation: 

\noindent
\textbf{\textit{channel align}}. $ B_l(\bm{V}_l) : \enspace \mathbb{R}^{C \times M}  \rightarrow \mathbb{R}^{D \times M} $,
 where $M=(H \times W +1)$. 
\label{sec:channel_align}

\rev{The brain encoding prediction target $\bm{Y} \in \mathbb{R}^{N \times 1}$ is beta weights (amplitude) of hemodynamic response (pulse) function \cite{prince_improving_2022}.  Denote scalar $y_i$  the individual voxel $i \in  \{ 1,2,\dots,N \}$ response.} To obtain the final brain prediction scalar $y_i$, we apply feature selection across the channels:

\noindent
\textbf{4) channel selector}.  $\bm{w}_i : \enspace \mathbb{R}^{D} \rightarrow \mathbb{R}^{1}$, where $\bm{w}_i$ answers, `Which is the best channel for predicting this brain voxel?'   Putting it all together, we have 
 


\vspace{-4mm}
\begin{equation*}
\begin{aligned}
    \bm{V} &= \texttt{ViT}(\bm{X}) \\
    \bm{\bar{\nu}}^{local}_{i,l} &= \texttt{Interp}(\bm{\hat{u}}_i^{space}; B_l(\bm{V}_l)) \quad\quad\quad\quad\quad\quad\quad\quad \text{(1)}\\
    \bm{v}_i &= \sum_{l=1}^{L} \hat{\omega}_{i,l}^{layer} ((1-{\hat{\alpha}_i}^{scale})\bm{\bar{\nu}}^{local}_{i,l} + {\hat{\alpha}_i}^{scale} \bm{\bar{\nu}}_{*,l}^{global}) \\
    y_i &= \bm{v}_i \bm{w}_i + b_i
\end{aligned}
\label{eq:select}
\vspace{-2mm}
\end{equation*}


\begin{figure*}[ht]
    \vspace{-8mm}
    \centering
    \includegraphics[width=0.9\linewidth]{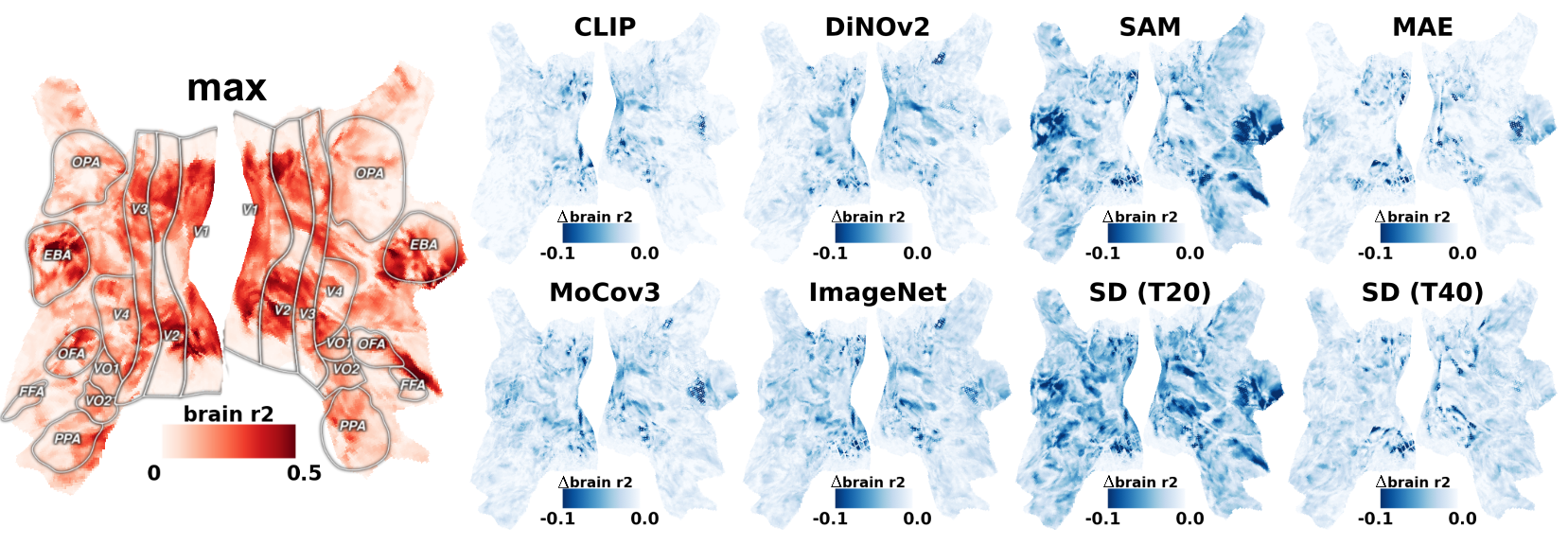}
    \vspace{-3mm}
    \caption{\textbf{Brain Score}. \textit{\textbf{Left}}: raw brain score $R^2$. \textit{\textbf{Right}}: difference of score to the model-wise max score (left). \textbf{\emph{Insights}}: \textbf{1)} CLIP and DiNOv2 predict semantic regions better but relatively weak for early visual, \textbf{2)} SAM and MAE are better at early visual region but weaker for body (EBA) and face (FFA) region, \textbf{3)}  Stable Diffusion (SD) shows a good prediction in all regions overall.  }
    \label{fig:brain_score}
    \vspace{-2mm}
\end{figure*}

\vspace{-4mm}
\paragraph{\emph{Topological Smooth}.} The factorized selector explicitly maps the brain and the network.  The topological structure of the corresponding brain voxels should also constrain this mapping.   The smoothness constraint can be formulated as Lipschitz continuity \cite{arjovsky_wasserstein_2017}: nearby brain voxels should have similar space, layer, and scale selection values.    We apply sinusoidal position encoding \cite{mildenhall_nerf_2020} to brain voxel. 
\subsection{Visualization and Coloring} 
\label{sec:color}
To visualize layer-to-brain mapping, we assign each voxel a color cue value associated with the layer with the highest layer selection value: $\texttt{argmax}_{L}(\bm{\hat{\omega}}^{layer}) \in \mathbb{R}^{N \times 1}$.  We assign voxel color brightness with a confidence measure $\bm{s} \in \mathbb{R}^{N \times 1}$ of $\bm{\hat{\omega}}^{layer}$:

\vspace{-6mm}
\begin{equation*}
\vspace{-2mm}
\begin{aligned}
\hspace{15mm} s_i = 1 - \frac{{\sum_{l=1}^{L} \hat{\omega}_{i,l}^{layer} \log \hat{\omega}_{i,l}^{layer}}}{{\sum_{l=1}^{L} \frac{1}{L} \log \frac{1}{L}}}  \hspace{15mm} \text{(2)} \hfilneg
\end{aligned}
\label{eq:constraints}
\vspace{0mm}
\end{equation*}

%


Note that $s_i$ equals 1 when $\bm{\hat{\omega}}_{i}^{layer}$ is a one-hot vector, and 0 when it is uniform.  In Figure \ref{fig:selectors}, we compare layer-selector trained with vs. without topological smooth constraints using this layer-to-brain color scheme. Topological smoothness significantly improved selection certainty.

\section{Results}
\label{sec:results}

For a fair comparison, we keep the same ViT network architecture while varying how the network is trained and the dataset used (Table \ref{tab:all_layer_selectors}). In Fig. \ref{fig:big_layer_selectors}, we display network layer-to-brain mapping for several popular pre-trained models.  An overview of our experiments:


\begin{enumerate}[noitemsep, nolistsep]
    \item \textit{What can relative brain prediction scores tell us? }
    \item  \textit{How do supervised and un-supervised training objectives change brain-network alignment?}
    \item  \textit{Do more data and larger model sizes lead to a more evident hierarchical structure?}
    \item \textit{What happens to a pre-trained network when fine-tuning to a new task with small samples?}
    \item \textit{Can the network channels be grouped to match well with brain functional units?}  
\end{enumerate}

\paragraph{Dataset}

We used Nature Scenes Dataset
(NSD) \cite{allen_massive_2022} for this study. Briefly, NSD provides 7T fMRI scan when watching COCO images \cite{lin_microsoft_2015}. A total of 8 subjects each viewed 3 repetitions of 10,000 images. We used the preprocessed and denoised single-trail data of the first 3 subjects \cite{prince_improving_2022}. We split 27,750 public trials into train validation and test sets (8:1:1) and ensured no data-leak of repeated trials. 

\subsection{Brain Score for Downstream Tasks Prediction}
\label{sec:brainscore}
The key finding is that 
a network with a high prediction score on a specific brain region is better suited for a relevant downstream task. CLIP, DiNOv2 and Stable Diffusion have overall high performance. 



Let $R^2 = 1 - \sum(y_{i,m} - \hat{y}_{i,m})^2/\sum(y_{i,m} - \bar{y}_{i,m})^2$ be the brain score metrics, $R^2$ is computed for each voxel $i$ over the test-set $m$. We report the raw score without dividing by noise ceiling or averaging repeated trials \cite{gifford_algonauts_2023}.  We compared the brain score for each model to the `max' model constructed by model-wise maximum for each voxel. We show the raw $R^2$ in Figure \ref{fig:brain_score}, and ROI-wise root sum squared difference to the `max' in Table \ref{tab:brain_score}.

\begin{table}[h]
\vspace{-2mm}


\centering
\resizebox{0.99\linewidth}{!}{
\begin{tblr}{
  row{4} = {fg=Gray},
  cell{1}{3} = {c=6}{c},
  cell{3}{1} = {c=2}{c},
  cell{3}{3} = {c=2}{c},
  hline{1,13} = {-}{0.1em},
  hline{3} = {3-8}{0.05em},
  hline{4} = {-}{0.08em},
}
\textbf{Model} & \textbf{Dataset} & \textbf{Root Sum Squared Difference $R^2$ $\downarrow$}              &                         &                         &                         &                         &                         \\
               &                  & \textbf{V1}             & \textbf{V2V3}           & \textbf{OPA}            & \textbf{EBA}            & \textbf{FFA}            & \textbf{PPA}            \\
Known Selectivity & & orientation & & navigate & body & face & scene \\
\textbf{max}     & &  0.237 &  0.215 &  0.097 &  0.185 &  0.186 &  0.134 \\
CLIP  \cite{radford_learning_2021}         & DC-1B \cite{gadre_datacomp_2023}          & \textbf{\textit{0.032}} & \textbf{0.023}          & \textbf{\textit{0.011}} & \textbf{\textit{0.015}} & \textbf{0.005}          & \textbf{0.006}          \\
DiNOv2 \cite{oquab_dinov2_2023}        & LVD-142M         & 0.033                   & 0.026                   & 0.021                   & \textbf{0.013}          & \textbf{\textit{0.008}} & \textbf{\textit{0.007}} \\
SAM  \cite{kirillov_segment_2023}          & SA-1B            & 0.037                   & 0.033                   & 0.025                   & 0.065                   & 0.056                   & 0.033                   \\
MAE  \cite{he_masked_2021}          & IN-1K            & \textbf{0.031}          & \textbf{\textit{0.025}} & \textbf{0.008}          & 0.029                   & 0.017                   & 0.009                   \\
MoCov3  \cite{chen_empirical_2021}       & IN-1K            & \textbf{\textit{0.032}} & 0.027                   & 0.014                   & 0.031                   & 0.015                   & 0.011                   \\
ImageNet \cite{deng_imagenet_2009}      & IN-1K            & 0.037                   & 0.032                   & 0.024                   & 0.028                   & 0.019                   & 0.015                   \\
SD (T20) \cite{rombach_high-resolution_2022}      & LAION-5B \cite{schuhmann_laion-5b_2022}        & 0.047                   & 0.050                   & 0.029                   & 0.056                   & 0.052                   & 0.032                   \\
SD (T40) \cite{rombach_high-resolution_2022}      & LAION-5B         & \textbf{0.031}          & 0.030                   & 0.021                   & 0.018                   & 0.019                   & 0.013                   
\end{tblr}
}
\vspace{-2mm}
\caption{\textbf{Brain Score}. Raw $R^2$ for max of all models and root sum squared difference for other models.}
\label{tab:brain_score}
\vspace{-4mm}
\end{table}

\definecolor{Jumbo}{rgb}{0.498,0.498,0.501}
\begin{figure*}[ht!]
    \vspace{-2mm}
    \centering
    \captionsetup{type=figure}
    \hspace{0mm}\includegraphics[width=1.0\linewidth]{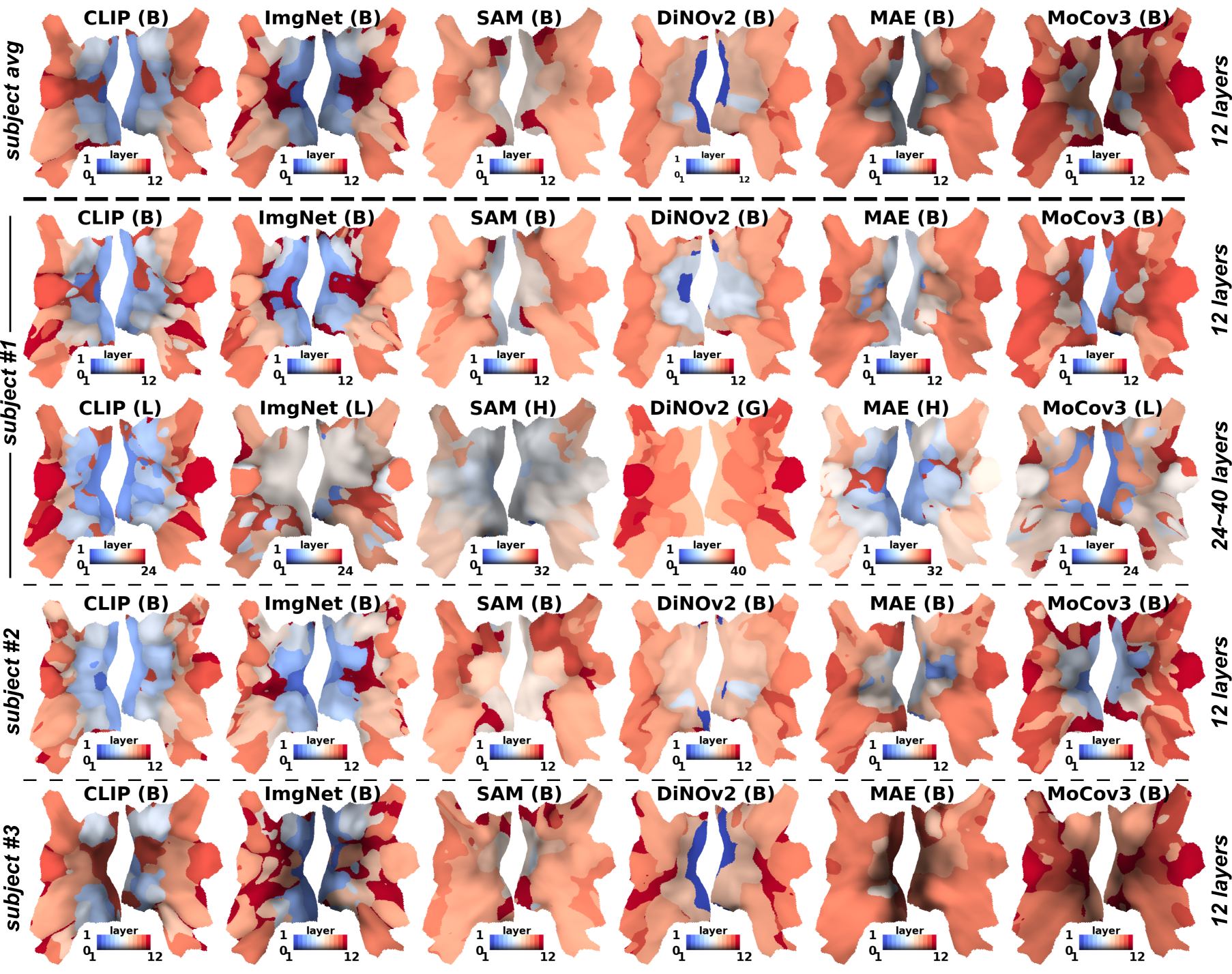}
    \vspace{-7mm}
    \captionof{figure}{\textbf{Layer Selectors, Brain-Network Alignment}. All models are ViT architecture, number of layers is marked in the colorbar x-axis. Brightness is confidence measurement (defined in Section \ref{sec:color}), and lower brightness means a \textit{softer} selection of multiple layers. \textit{\textbf{Top}}: average of three subjects, base size 12-layer model. \textit{\textbf{Middle}}: subject \#1, 12 layer small(S) and base(B) model, 24 layer large(L) model, 32 layer huge(H) model, 40 layer gigantic(G) model. \textit{\textbf{Bottom}}: subject \#2 and \#3, base size 12-layer model. \emph{\textbf{Insights}}: \textbf{1)} CLIP layers align best with the brain's hierarchical organization, \textbf{2)} ImageNet and SAM last layer align with mid-level in the brain, indicating their training objectives aimed at mid-level concept; \textbf{3)} DiNOv2: with a larger model, its hierarchy no longer align with the brain.} \label{fig:big_layer_selectors}
    
\vspace{4mm}
\resizebox{\linewidth}{!}{
\begin{tblr}{
  row{4} = {fg=Jumbo},
  cell{1}{2} = {c=4}{c},
  cell{1}{6} = {c=2}{c},
  cell{1}{8} = {c=3}{c},
  cell{1}{11} = {c=3}{c},
  cell{1}{14} = {c=3}{c},
  cell{1}{17} = {c=3}{c},
  cell{3}{2} = {c},
  cell{3}{3} = {c},
  cell{3}{4} = {c},
  cell{3}{5} = {c},
  cell{3}{6} = {c=2}{c},
  cell{3}{8} = {c=3}{c},
  cell{3}{11} = {c=3}{c},
  cell{3}{14} = {c=3}{c},
  cell{3}{17} = {c=3}{c},
  hline{1,8} = {-}{0.1em},
  hline{4} = {1}{},
  hline{2} = {2}{l},
  hline{2} = {3-4}{},
  hline{2} = {5}{r},
  hline{2} = {6}{l},
  hline{2} = {7}{r},
  hline{2} = {8}{l},
  hline{2} = {9}{},
  hline{2} = {10}{r},
  hline{2} = {11}{l},
  hline{2} = {12}{},
  hline{2} = {13}{r},
  hline{2} = {14}{l},
  hline{2} = {15}{},
  hline{2} = {16}{r},
  hline{2} = {17}{l},
  hline{2} = {18}{},
  hline{2} = {19}{r},
  hline{2} = {20}{l},
  hline{2} = {21}{},
  hline{2} = {22}{r},
  hline{4} = {2}{l},
  hline{4} = {3-4}{},
  hline{4} = {5}{r},
  hline{4} = {6}{l},
  hline{4} = {7}{r},
  hline{4} = {8}{l},
  hline{4} = {9}{},
  hline{4} = {10}{r},
  hline{4} = {11}{l},
  hline{4} = {12}{},
  hline{4} = {13}{r},
  hline{4} = {14}{l},
  hline{4} = {15}{},
  hline{4} = {16}{r},
  hline{4} = {17}{l},
  hline{4} = {18}{},
  hline{4} = {19}{r},
}
\textbf{Model}              & \textbf{CLIP} \cite{ilharco_openclip_2021} &                        &       &       & \textbf{ImageNet} \cite{deng_imagenet_2009} &                        & \textbf{SAM} \cite{kirillov_segment_2023} &       &                        & \textbf{DiNOv2} \cite{oquab_dinov2_2023} &       &                & \textbf{MAE} \cite{he_masked_2021} &               &                        & \textbf{MoCov3} \cite{chen_empirical_2021} &                &                        \\
\textbf{Size}               & L/14                                                          & B/16                   & B/32  & B/32  & L                                                              & B                      & H                                                            & L     & B                      & G                                                           & L     & B              & H                                                     & L             & B                      & L                                                             & B              & S                      \\
\textbf{Data}               & 1B \cite{gadre_datacomp_2023}              & 140M                   & 14M   & 1.4M  & IN-1K  \cite{deng_imagenet_2009}            &                        & SA-1B  \cite{kirillov_segment_2023}       &       &                        & LVD-142M  \cite{oquab_dinov2_2023}       &       &                & IN-1K  \cite{deng_imagenet_2009}   &               &                        & IN-1K \cite{deng_imagenet_2009}            &                &                        \\
\textbf{$R^2$ $\uparrow$}   & \textbf{0.132}                                                & 0.131                  & 0.117 & 0.083 & 0.117                                                          & \textbf{0.121}         & \textbf{0.120}                                               & 0.117 & 0.111                  & 0.123                                                       & 0.125 & \textbf{0.128} & \textbf{0.132}                                        & 0.129         & 0.128                  & 0.124                                                         & \textbf{0.127} & 0.126                  \\
\textbf{slope $\uparrow$}   & \textbf{0.53}                                                 & 0.50                   & 0.32  & 0.11  & 0.27                                                           & \textbf{0.39}          & 0.08                                                         & 0.10  & \textbf{0.15}          & 0.16                                                        & 0.25  & \textbf{0.41}  & 0.20                                                  & \textbf{0.37} & 0.32                   & 0.30                                                          & 0.33           & \textbf{0.40}          \\
\textbf{$b_0$ $\downarrow$} & \textit{\textbf{0.35}}                                                 & 0.38                   & 0.49  & 0.60  & \textit{\textbf{0.41}}                                                  & 0.45                   & \textit{\textbf{0.58}}                                                & 0.63  & 0.67                   & 0.76                                                        & 0.66  & \textit{\textbf{0.50}}  & \textit{\textbf{0.46}}                                         & 0.47          & 0.55                   & \textit{\textbf{0.40}}                                                 & 0.52           & 0.51                   \\
\textbf{$b_1$ $\uparrow$}   & \textbf{\textit{0.88}}                                        & \textbf{\textit{0.88}} & 0.82  & 0.71  & 0.68                                                           & \textbf{\textit{0.83}} & 0.66                                                         & 0.73  & \textbf{\textit{0.82}} & \textit{\textbf{0.92}}                                                        & \textit{\textbf{0.92}}  & 0.91           & 0.66                                                  & 0.84          & \textbf{\textit{0.87}} & 0.70                                                          & 0.85           & \textbf{\textit{0.91}} 
\end{tblr}
}
\vspace{-1mm}
\captionof{table}{\textbf{Layer Selectors, Brain-Network Alignment}. Brain-network alignment is measured by slope and intersection of linear fit (defined in Section \ref{par:slope}). Larger \textbf{slope} means generally better alignment with the brain, smaller \textbf{$b_0$} means better alignment of early layers, and larger \textbf{$b_1$} means better alignment of late layers. \textbf{$R^2$} is brain score. \textbf{Bold} marks the best within the same model. \emph{\textbf{Insights}}: \textbf{1)} CLIP's alignment to the brain improves with larger model capacity, \textbf{2)} for all others, bigger models decrease the brain-network hierarchy alignment.}\label{tab:all_layer_selectors}
\end{figure*}%

\definecolor{Grey}{rgb}{0.494,0.494,0.501}
\begin{figure*}[ht]
    \vspace{-2mm}
    \centering
    \includegraphics[width=0.95\linewidth]{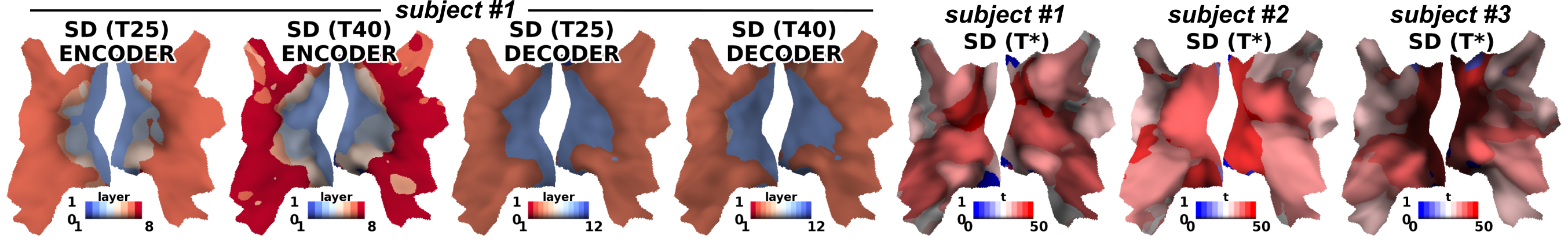}
    \vspace{-4mm}
    \caption{\textbf{Diffusion Models}. \textit{\textbf{Left}}: one time step (T=25 and T=40) layer selection, UNet encoder and decoder layers. \textit{\textbf{Right}}: fix layer (UNet decoder layer 6) time step selection. Color brightness is confidence measure (Section \ref{sec:color}). \textbf{\emph{Insights}}: \textbf{1)} Diffusion models have less delineation in brain-network mapping using fixed time-step encoder/decoder layers,  but more separation when using time steps, \textbf{2)} mid-late time steps align with higher level brain, late time step aligns with early brain region.}
    \label{fig:ddim_timesel}
    \vspace{-3mm}
\end{figure*}

Figure \ref{fig:brain_score} shows that the fovea regions of early visual cortex are highly predictable, and so are  higher regions of EBA and FFA, followed by PPA.  In Table \ref{tab:brain_score}, we found DiNOv2 and CLIP predict well on EBA and FFA but poorly for early visual regions; MAE and SAM are the opposite.   Stable Diffusion (SD) features, described in the next section, perform well in all regions.   This finding is consistent with recent works that show SD features are helpful for many visual tasks, from segmentation to semantic correspondence \cite{xu_open-vocabulary_2023, tang_emergent_2023}.  It could also explain why a combination of DiNOv2 for coarser semantic correspondence with SD for finer alignment could work well \cite{zhang_tale_2023, walmer_teaching_2023}.


\subsection{Training Objectives and Brain-Net Alignment}
\label{sec:training_objectives}

The key finding is training objective matters:
\textbf{1)} supervised methods show a more detailed delineation of network-to-brain mapping compared to self-supervised ones;
\textbf{2)}  ImageNet and SAM show the last layer mapped to the middle region of the brain;
\textbf{3)}  Stable Diffusion features show more detailed delineation between the time steps than between the UNet encoder or decoder layers.


The layer multi-selector output indicates, ``\emph{within one model, which layer best predicts this brain region?}''. 
Even though the mapping differs for subjects (Figure \ref{fig:big_layer_selectors}), the pattern of subject difference is consistent in both CLIP and ImageNet models: subject \#3 had considerably low confidence in early visual brain, and subjects \#2 and \#3 are missing the FFA (face) region that subject \#1 has.

For supervised pre-trained models, Figure \ref{fig:big_layer_selectors} shows CLIP's \cite{radford_learning_2021} last layer is close to EBA for subject average and EBA/FFA for subject \#1, probably because the training data contain languages related to body and face.   Surprisingly, ImageNet's \cite{deng_imagenet_2009} last layer is close to the mid-level lateral stream, suggesting that simple image labels are more primitive than text language.  SAM's \cite{kirillov_segment_2023} final layer is close to the mid-level ventral and parietal stream, indicating segmentation as a mid-level visual task.   These observations suggest a bottom-up feature computation and top-down task prediction in ImageNet and SAM. 

For the self-supervised models, the final layer of DiNOv2 (DiNOv1+iBOT) \cite{oquab_dinov2_2023, zhou_ibot_2022, caron_emerging_2021} and MAE \cite{he_masked_2021} is missing from the network-to-brain mapping, which indicates the last stage of un-supervised mask reconstruction deviates from the brain tasks.   For MoCov3 \cite{chen_empirical_2021}, there's a trend that the second-last layer matched more with the ventral stream (``what'' part of the brain) than the parietal stream (``where'' part), indicating self-contrastive learning is more focused on the semantics rather than spatial relationship \cite{walmer_teaching_2023}.

We also analyzed Stable Diffusion \cite{rombach_high-resolution_2022} by 1) fixing the time step and selecting layers, and 2) fixing the UNet decoder layer 6 and selecting time steps. We followed the ``inversion'' \cite{luo_diffusion_2023} time steps feature extraction and used a total of T$=$50 time steps.  In Figure \ref{fig:ddim_timesel}, layer selection showed that the diffusion model has less separation for early and late regions; this was true for both T$=$25, T$=$40, encoders and decoders.  Time step selection showed diffusion model early time steps (T$<$25) deviate from the brain tasks. The confidence (Section \ref{sec:color}) of time step selection was relatively high for EBA at (T$=$30) and for mid-level visual stream at (T$=$35, T$=$40). Overall, our results indicate that 1) the diffusion model has less feature separation across layers but instead is separated across time steps, and 2) global features are more in the middle-time steps, while local features are more aligned with the mid-to-late time steps.

\begin{figure*}[ht!]
    \vspace{2mm}
    \centering
    \includegraphics[width=\linewidth]{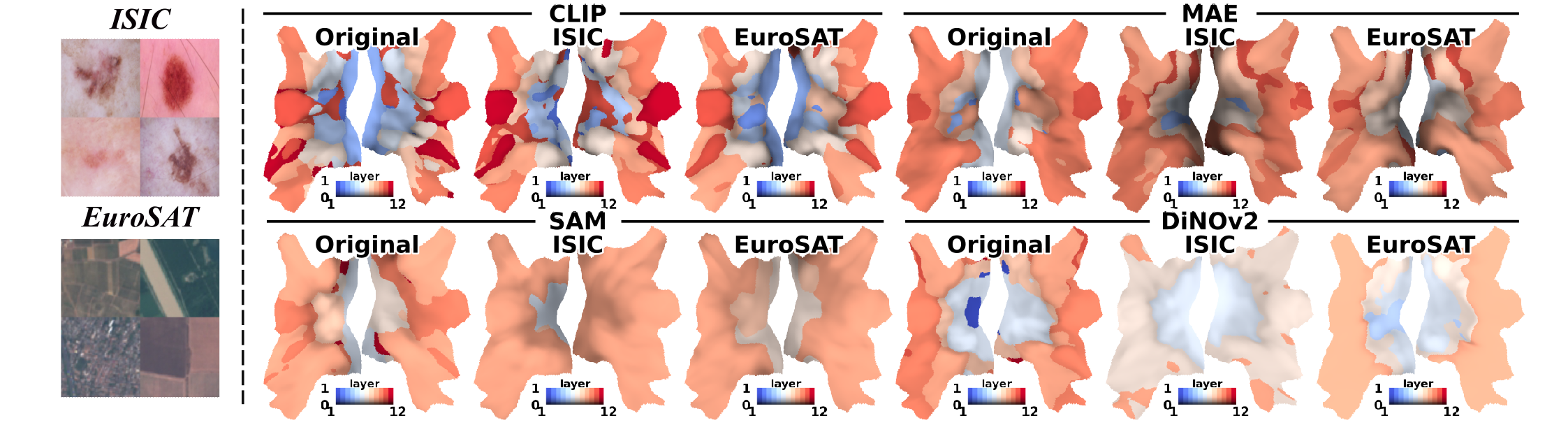}
    \vspace{-5mm}
    \caption{\textbf{fine-tuned to Small Datasets}. \textit{\textbf{Left}}: example images from ISIC and EuroSAT. \textit{\textbf{Right}}: layer selector (Colored as Section \ref{sec:color}) before and after fine-tuning. The whole network is fine-tuned. \textbf{\emph{Insights}}: CLIP fine-tunes with less change in the existing computation.}
    \label{fig:fine-tune_ls}
    \vspace{-0mm}
\end{figure*}

\subsection{Network Hierarchy and Model Sizes}
\label{sec:model_sizes}

The key findings are:
\textbf{1)} CLIP shows a more substantial alignment of hierarchical organization with the brain;
\textbf{2)} when scaling with more data and bigger model size, CLIP shows an improvement in its brain-hierarchical alignment, while others show a decrease. 


We propose a measure called \textbf{\emph{hierarchy slope}} by putting predefined brain ROI regions into a number-ordering and fitting a linear regression as a function of their layer selector output $\bm{\hat{\omega}}^{layer}$.  We used only coarse brain regions and did not consider feedback computation in the brain.

\noindent\textbf{\emph{Hierarchy slope}}
\label{par:slope}
Let $\hat{\iota}_i = \sum_{l=1}^{L} \frac{l-1}{L-1} \hat{\omega}_{i,l}^{layer}$ be a scalar that represents vector layer selector weights $\bm{\hat{\omega}}_i^{layer}$, such that $\hat{\iota}_i \in [0, 1]$. We pre-defined a four-level brain structure: 1) V1, 2) V2\&V3, 3) OPA, 4) EBA. Voxels inside these ROIs are assigned with an ideal value $\iota_i \in \{0, 0.33, 0.66, 1\}$. We fit a linear regression $\hat{\iota}_i = \beta \iota_i + \epsilon$, where slope $\beta$ measures brain-model alignment, $b_0 = \epsilon$ and $b_1 = \beta + \epsilon$ measures the proportion of early and late layer not being selected.

We found that both qualitatively (Figure \ref{fig:big_layer_selectors}) and quantitatively (Table \ref{tab:all_layer_selectors}), layer-to-brain alignment is best in the CLIP model.   Furthermore, the \emph{{hierarchy slope}} increases as CLIP scaled up both model size and data (slope 0.32 for M, 0.50 for L, 0.52 for XL).   CLIP M and S models were trained with the same model size but smaller data; the S model dropped \emph{{hierarchy slope}} significantly (0.11).   ImageNet, SAM, MoCov3, and DiNOv2 models show decreased \emph{{hierarchy slope}} when scaling up: their late \rev{(or early for DiNOv2)} layers were less selected for bigger models, indicating a decreasing hierarchical alignment with the brain.  

\subsection{Fine-tuned Model}
\label{sec:fine-tune}
The key findings are:
\textbf{1)} CLIP maintains a hierarchical structure and uses less re-wiring for downstream tasks;
\textbf{2)} DiNOv2 and SAM tend to re-wire their intermediate layers and lose their hierarchical structure rapidly when fine-tuned.

 
We fine-tuned on two small-scale downstream tasks, ISIC \cite{codella_skin_2019} skin cancer classification, and EuroSAT \cite{helber_eurosat_2019} satellite land-use classification. We used 50 training samples per class to train.  The pre-trained model is fine-tuned across all layers with AdamW optimizer lr=3e-5, weight decay of 0.01, batch size of 4, for 3,000 steps. We verified that the fine-tuned models reached maximum validation performance without significant overfitting. 

\begin{table}[h]
\centering
\vspace{-2mm}
\resizebox{0.99\linewidth}{!}{
\begin{tblr}{
  cell{1}{2} = {c=3}{c},
  hline{1,7} = {-}{0.1em},
  hline{2} = {2-4}{0.05em},
  hline{3} = {-}{0.05em}
}
                       & \textbf{Brain Score $R^2$ $\uparrow$} &       &         \\
\textbf{Model\space/\space Fine-tune dataset } & \textbf{Original}             & \textbf{ISIC}\quad  & \textbf{EuroSAT} \\
\textbf{CLIP}                   & 0.131                & 0.115 & 0.112   \\
\textbf{MAE}                    & 0.128                & 0.117 & 0.113   \\
\textbf{SAM}                    & 0.111                & 0.086 & 0.087   \\
\textbf{DiNOv2}                 & 0.128                & 0.085 & 0.082   
\end{tblr}
}
\vspace{-2mm}
\caption{\rev{Brain score dropped after fine-tuning on small datasets.}}
\label{tab:finetune_brain_score}
\vspace{-4mm}
\end{table}

We apply brain-to-network mapping to visualize the fine-tuned networks. The first dataset ISIC skin cancer classification relies on low-level features. Figure \ref{fig:fine-tune_ls} shows ImageNet/CLIP's last layer aligned with V1 after ISIC fine-tuning, potentially indicating the usage of top-down information for low-level vision tasks. The second dataset, EuroSAT, requires less fine-tuning on low-level features; V1 is still aligned to early layers for CLIP. 
\rev{After fine-tuning, qualitative results in Figure \ref{fig:fine-tune_ls} showed CLIP and MAE maintained a strong hierarchical structure, while SAM and DiNOv2 largely lost their hierarchy; quantitative results in Table \ref{tab:finetune_brain_score} showed brain score of CLIP and MAE dropped less compare to SAM and DiNOv2. 
Overall, CLIP and MAE adapt to dynamic tasks with less catastrophic forgetting and re-wiring of existing computation.
}

\begin{figure*}[ht!]
    \centering
    \includegraphics[width=\linewidth]{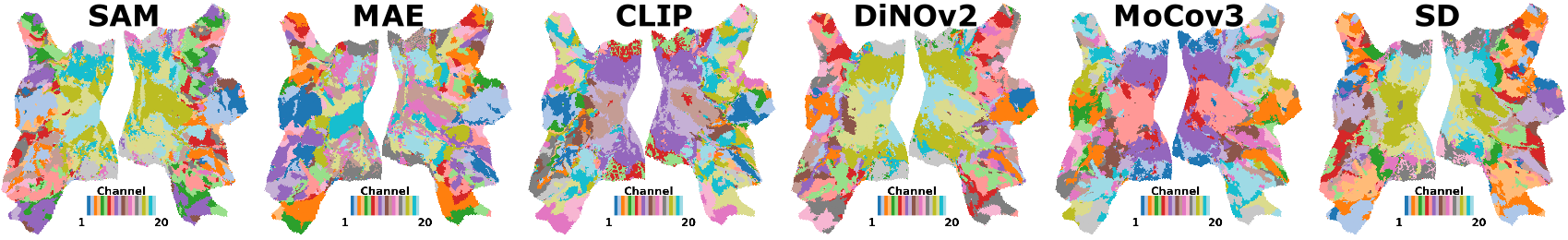}
    \vspace{-6mm}
    \caption{\textbf{Channel Clustering}. Brain voxels clustered by channel selection weight $\bm{w}_i$. \textbf{\emph{Insights}}: early visual brain uses less diverse channels but more diverse spatial locations (Figure \ref{fig:selectors}), higher level brain is the opposite. }
    \label{fig:channel_clustering}
    \vspace{-2mm}
\end{figure*}

\begin{figure*}[ht!]
    \centering
    \includegraphics[width=\linewidth]{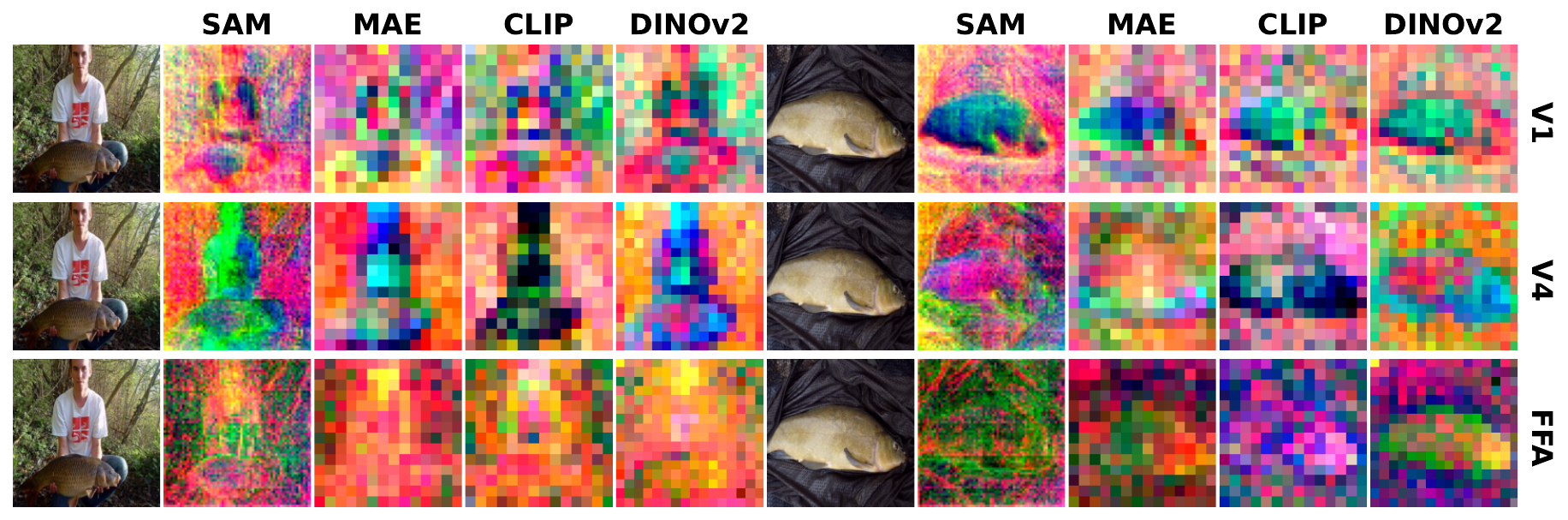}
    \vspace{-6mm}
    \caption{{\textbf{Image features (selected channels).} Image RGB value corresponds to top-3 principal components (PCA) of $n$ brain voxels' channel selection weights inside each brain ROI. The top-3 PCA channel selection weights are multiplied with channel-aligned image features and summed at every image pixel. \textbf{V1}: early visual brain, \textbf{V4}: mid-level visual brain, \textbf{FFA}: face-selective brain region.}}
    \label{fig:channel_rgb}
    \vspace{-2mm}
\end{figure*}

\subsection{Channels and Brain ROIs}
\label{sec:channel}





The key findings are:
\textbf{1)} we can cluster brain voxels using the co-occurrence of brain voxels with channels, and the clusters largely align well with known brain ROIs;  
\textbf{2)} we can compute brain ROI/cluster-specific responses on images to reveal the ROI functionality. 

Recall our factorized multi-selector method compresses information across 4D network features into a channel-wise vector of $\bm{v}_i$ for each brain voxel $i$.   Furthermore, channels across all the layers are aligned (Methods \ref{sec:channel_align}),  resulting in a layer-agnostic channel representation. From that, a linear regression weight vector $\bm{w}_i$ acts as a \emph{channel selector} to determine ``\emph{which feature channels best predict this brain voxel?}''   We can view this channel feature selector, $\bm{w(k)}_i$, as co-occurrence between brain voxel $i$ and channel elements $k$, which can be used to cluster the brain voxels: linking two voxels, $i,j$ if they share similar channel selectors $\bm{w}_i$, $\bm{w}_j$. Figure \ref{fig:channel_clustering} shows the result of clustering brain voxels into 20 clusters.  



The higher-level brain utilized diverse channels across the brain areas; there is a consistent pattern that the face and body region use the same channel in CLIP, DiNOv2, MoCov3, and SD. The early visual brain used similar channels across the visual cortex; there is a consistent pattern that the left and right brain are symmetrical, as well as the ventral and parietal streams. SAM and MAE early visual brain-selected channels are non-symmetrical, indicating shift variant properties \cite{walmer_teaching_2023}.


Furthermore, the selected channels reveal brain ROI’s functionality. We visualized image feature response produced by the top-3 PCA components of channel weights within the selected ROIs (in Figure \ref{fig:channel_rgb}), which shows the brain ROIs encode low-level edge information in V1, mid-level semantic segmentation in V4, and face-selective features in FFA. 
Interestingly, DiNOv2 generalizes face across humans and fish \cite{walmer_teaching_2023, zhang_tale_2023}.

\section{Discussion and Limitations}

 We have developed a visualization tool, \emph{FactorTopy}, by training a robust brain encoding model.  It allows us to see the internal working mechanism of any deep network.  With this visualization and known functionality of brain ROIs, we can predict the network's downstream task performance, and diagnose their behavior when scaling up with a larger model or fine-tuning to a small dataset.  

\vspace{2mm}
\noindent
\textbf{Limitations}  
High-quality brain-encoding data of input images paired with brain fMRI responses is needed. NSD is the only such data publicly available. 
Over time, this situation might improve. 
Comparing brain-to-network alignments is less informative if networks' computation differs entirely from the brain.  
It is possible to achieve efficiency and generalization in a non-brain-like way; therefore, our tool is not universally applicable to all network designs. 

\vspace{5mm}
\noindent
\textbf{Acknowledgement} 
This work is supported by the funds provided by the National Science Foundation and by DoD OUSD (R\&E) under Cooperative Agreement PHY-2229929 (The NSF AI Institute for Artificial and Natural Intelligence). Huzheng Yang and James Gee are supported in part by R01-HL133889, R01-EB031722, and RF1-MH124605.


  \clearpage
\newpage
{\small
\bibliographystyle{ieee_fullname}
\bibliography{main}
}

\clearpage
\appendix
\twocolumn[{%
\renewcommand\twocolumn[1][]{#1}%

    \centering
    \captionsetup{type=figure}
    \includegraphics[width=\linewidth]{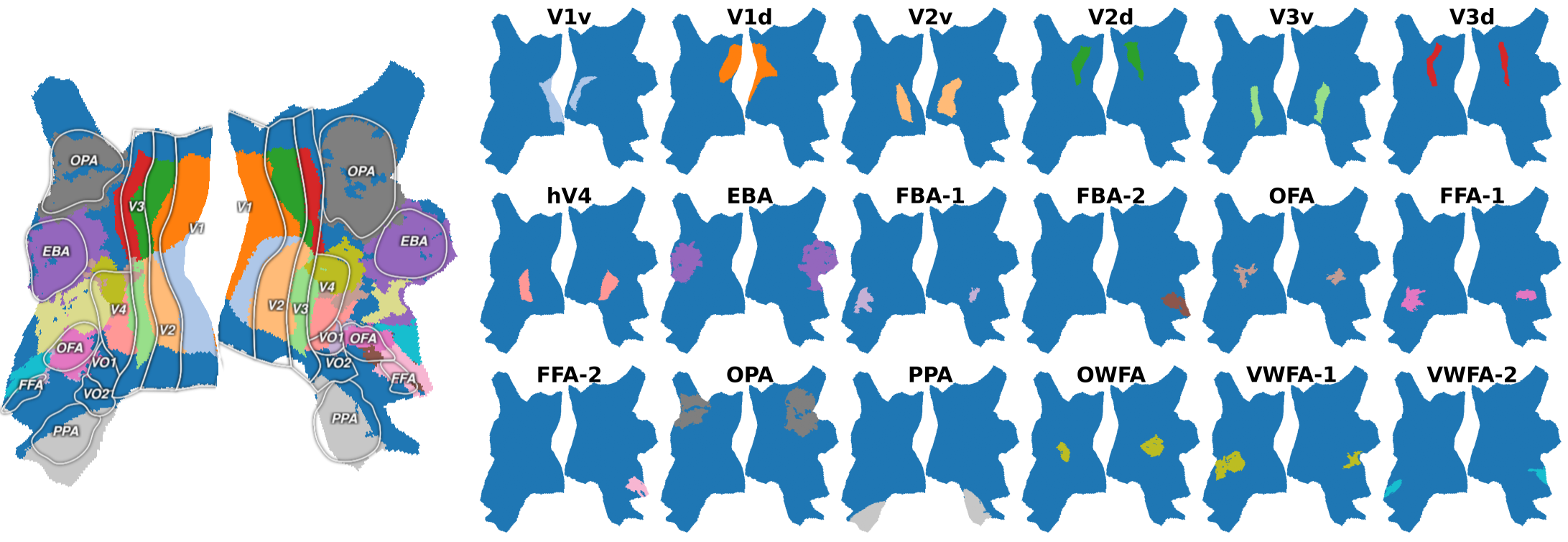}
    \captionof{figure}{\textbf{Brain Region of Interests (ROIs)}. \textit{\textbf{Left}}: color is subject-specific ROI, and border overlay is subject-average common template ROIs. \textbf{\textit{Right}}: subject-specific ROIs. V1v: ventral stream, V1d: dorsal stream.}
    \label{fig:supp_rois}

\vspace{4mm}
    
    \resizebox{0.99\linewidth}{!}{
    \begin{tblr}{
      hline{1,3} = {-}{0.08em},
    }
    \textbf{ROI name}  \quad\quad\quad & V1 V2 V3 & V4 \quad & EBA FBA & OFA FFA & OPA \quad & PPA \quad & OWFA VWFA \\
    \textbf{Known Function/Selectivity} \quad\quad & primary visual & mid-level & body & face & navigation & scene & words \\
    \end{tblr}
    }
    \captionof{table}{Known function and selectivity of brain region of interests (ROIs).}\label{tab:supp_rois}
\vspace{10mm}
}]


\section{Appendix Overview}

\begin{enumerate}
\item Table \ref{tab:supp_tab_overview} is an overview of key findings in this work.
\item \cref{sec:brain_details} summarizes known function of brain ROIs. 
\item \cref{sec:model_details} lists details of pre-trained models. 
\item \cref{sec:supp_results} is extended results with more ViT sizes, diffusion time steps, and more subjects.
\item \cref{sec:supp_methods} is implementation details of data processing and model training, pseudocode for visualization.
\item \cref{sec:sota} summarizes state-of-the-art methods, ablation study of our methods.
\item \cref{sec:supp_less_data} demonstrates the resulting brain-to-network mapping when trained with less data samples.
\end{enumerate}


\newpage
\section{Brain Region Details}
\label{sec:brain_details}
This section briefly summarizes the known functionality of brain regions of interest (ROI). In our primary result, we included numerical results for V1, V2, V3, OPA, PPA, EBA, and FFA. In this appendix, we further report numerical results on FBA, OFA, OWFA, and VWFA.

 \Cref{fig:supp_rois} is an overview of brain ROIs. We used subject-specific ROIs provided by NSD \cite{allen_massive_2022}, NSD defined subject-specific ROIs by population receptive field (prf) and functional localizer (floc) experiments. It's worth noting that common template ROIs are different from subject-specific ROIs.

\Cref{tab:supp_rois} is known function and selectivity for each ROI. Briefly, V1 to V3 is the primary visual stream, they are further divided into ventral (lower) and dorsal (upper) streams. V4 is the mid-level visual area. EBA (extrastriate body area) and FBA (fusiform body area) are body-selective regions, FFA (fusiform face area) and OFA (occipital face area) are face-selective, OWFA (occipital word form area) and VWFA (visual word form area) are words selective. PPA (parahippocampal place area) is scene and place selective, and OPA (occipital place area) is related to navigation and spatial reasoning.

\begin{table*}[ht!]
\setlength{\cmidrulewidth}{0.01em}
\renewcommand{\tabcolsep}{6pt}
\renewcommand{\arraystretch}{1}
\centering\renewcommand\cellalign{lc}

\resizebox{0.99\linewidth}{!}{
\begin{tabular}{@{}lcc@{}}
\toprule
 \textbf{Key Observations} &  \textbf{Sections} & \textbf{Figures \& Tables} \\
 \midrule \\[-8pt]

\textbf{\textit{Brain Score}} \\
\makecell*[{{p{13cm}}}]{MAE and SAM are relatively better for the early visual brain, \\CLIP and DiNOv2 are relatively better for high-level brain regions.} & \ref{sec:brainscore}, \ref{sec:supp_brainscore} & Fig. \ref{fig:brain_score}; Tab. \ref{tab:brain_score}, \ref{tab:supp_brainscore} \\
\makecell*[{{p{13cm}}}]{SD late time-steps are uniformly good for all brain regions.} & \ref{sec:brainscore}, \ref{sec:supp_brainscore} & Fig. \ref{fig:brain_score}; Tab. \ref{tab:brain_score}, \ref{tab:supp_brainscore}\\
\makecell*[{{p{13cm}}}]{SD late time-steps are better for early brain, mid-to-late time steps are better for late brain.} & \ref{sec:supp_brainscore} & \cref{tab:supp_brainscore}\\
\midrule
\textbf{\textit{Brain-Net Alignment}} \\
\makecell*[{{p{13cm}}}]{Across all models included in this study, CLIP has the best brain-alignment.} & \ref{sec:training_objectives}, \ref{sec:model_sizes} & Fig. \ref{fig:big_layer_selectors}; Tab. \ref{tab:all_layer_selectors} \\
\makecell*[{{p{13cm}}}]{ImageNet and SAM last layer align to the mid-level visual brain, classification and segmentation are mid-level brain tasks.} & \ref{sec:training_objectives}, \ref{sec:supp_align} & Fig. \ref{fig:big_layer_selectors}, \ref{fig:supp_vit_raw_p2} \\
\makecell*[{{p{13cm}}}]{DiNOv2 and MAE last layer does not align to any brain region, mask reconstruction deviates from the brain's task.} & \ref{sec:training_objectives}, \ref{sec:supp_align} & Fig. \ref{fig:big_layer_selectors}, \ref{fig:supp_vit_raw_p2}  \\
\makecell*[{{p{13cm}}}]{MoCov3 last layers align better with the late ventral stream (`what' part) than the dorsal stream (`what' part), self contrastive learning is more on semantics than spatial relationship.} & \ref{sec:training_objectives}, \ref{sec:supp_align} & Fig. \ref{fig:big_layer_selectors}, \ref{fig:supp_vit_raw_p2} \\
\makecell*[{{p{13cm}}}]{CLIP and ImageNet early layers align with the early visual brain, SAM DiNOv2 MAE MoCov3 early layers deviate from the brain.} & \ref{sec:training_objectives}, \ref{sec:supp_align} & Fig. \ref{fig:big_layer_selectors}, \ref{fig:supp_vit_raw_p1}  \\
\makecell*[{{p{13cm}}}]{SD have less separation in layers but more in time steps. \\SD encoder layers have more separation than decoder layers. \\SD's brain-alignment is more `soft' compared to ViT models. \\SD final time steps align to early brain regions, SD mid-late time steps align to the late brain.} & \ref{sec:training_objectives}, \ref{sec:supp_align} & Fig. \ref{fig:ddim_timesel}, \ref{fig:supp_sd_raw_p3}\\
\midrule
\textbf{\textit{Model Sizes}} \\
\makecell*[{{p{13cm}}}]{CLIP's brain-net alignment improved as CLIP scaled up size and training data. In bigger CLIP models, both early and late layers become more aligned with the brain.} & \ref{sec:model_sizes}, \ref{sec:sizes} & Fig. \ref{fig:big_layer_selectors}, \ref{fig:supp_clip_size}  \\
\makecell*[{{p{13cm}}}]{SAM, ImageNet, DiNOv2, MoCov3, and MAE's brain-net alignment decreased as they scaled up sizes. ImageNet and DiNOv2 bigger models' early layers deviate from the brain; SAM, MAE, and MoCov3 bigger models' late layers deviate from the brain.} & \ref{sec:model_sizes}, \ref{sec:sizes} & Fig. \ref{fig:big_layer_selectors}, \ref{fig:supp_sam_size}-\ref{fig:supp_moco_size} \\
\midrule
\textbf{\textit{Fine-tuning}} \\
\makecell*[{{p{13cm}}}]{CLIP maintained brain-alignment after fine-tuning, DiNOv2 and MAE re-wired late layers.} & \ref{sec:fine-tune} & Fig. \ref{fig:fine-tune_ls}, Tab. \ref{tab:finetune_brain_score} \\
\makecell*[{{p{13cm}}}]{Fine-tuning performance does not correlate to change of computation layout, CLIP had the best fine-tuning performance but DiNOv2 and MAE also had competitive performance.} & \ref{sec:supp_finetune} & Tab. \ref{tab:supp_finetune} \\
\midrule
\textbf{\textit{Channels and Brain ROIs}} \\
\makecell*[{{p{13cm}}}]{Early visual brain uses similar channels but diverse spatial tokens. \\Late visual brain use diverse channels and global token.} & \ref{sec:channel} & Fig. \ref{fig:channel_clustering}, \ref{fig:selectors} \\
\makecell*[{{p{13cm}}}]{The top selected channels reveal brain ROIs' function. \\Image space features also reveal differences in various pre-trained models.} & \ref{sec:supp_channel} & Fig. \ref{fig:channel_rgb}, \ref{fig:supp_top_channels1}-\ref{fig:supp_top_channels6}\\
\midrule
\textbf{\textit{Methods and Consistency}} \\
\makecell*[{{p{13cm}}}]{Consistent subject difference exists in both brain prediction score and brain-net alignment.} & \ref{sec:supp_brainscore}, \ref{sec:training_objectives} & Fig. \ref{fig:supp_brain_score}, \ref{fig:big_layer_selectors}; Tab. \ref{tab:supp_brainscore}\\
\makecell*[{{p{13cm}}}]{Brain-Net mapping is consistent across random seeds within the same subject.} & \ref{sec:supp_align} & Fig. \ref{fig:supp_random_seed_consistency} \\ 
\makecell*[{{p{13cm}}}]{Brain-Net mapping can be trained with limited training data samples. 3K data samples is a good trade-off for speed and quality.} & \ref{sec:supp_less_data} & Fig. \ref{fig:num_data}, \ref{fig:supp_num_data}\\ 

\bottomrule
\end{tabular}
}
\caption{Overview of key observations in this work.}
\label{tab:supp_tab_overview}
\end{table*}

\newpage
\section{Pre-trained Model Details}
\label{sec:model_details}
This section briefly summarizes the models included in this study. All the models are ViT architecture except for U-Net Stable Diffusion. We primarily used models released by their original authors, we used models from third-party releases when size variants are unavailable from the official release. We did not run any pre-training ourselves.



\begin{table}[H]
\centering
\resizebox{0.99\linewidth}{!}{
\begin{tblr}{
  hline{1,6} = {-}{0.08em},
  hline{2} = {-}{0.05em},
}
\textbf{Model} & \textbf{Layers} & \textbf{Width} & \textbf{Input Size} & \textbf{Patch Size} & \textbf{Training Data} \\
CLIP XL & 24 & 1024 & 224x224 & 14x14 & DataComp-1B \\
CLIP L & 12 & 768 & 224x224 & 16x16 & DataComp-140M \\
CLIP M & 12 & 768 & 224x224 & 32x32 & DataComp-14M \\
CLIP S & 12 & 768 & 224x224 & 32x32 & DataComp-1.4M \\
\end{tblr}
}
\caption{CLIP Models.}
\label{tab:supp_clip}
\end{table}

\paragraph{CLIP} 
The objective of CLIP \cite{radford_learning_2021} (Contrastive Language-Image Pre-Training) is to match images with their corresponding text captions. The training objective is to minimize a contrastive loss that increases the similarity of paired images and text but decreases for unpaired ones. CLIP has two branches, one for vision and one for text, we only used the vision branch. We used a model released from the OpenCLIP \cite{ilharco_openclip_2021} repository, models are pre-trained on data from DataComp \cite{gadre_datacomp_2023}. Size variants of CLIP were trained on different sub-samples of data from 1B to 1.4M samples.

\begin{table}[H]
\centering
\resizebox{0.99\linewidth}{!}{
\begin{tblr}{
  hline{1,5} = {-}{0.08em},
  hline{2} = {-}{0.05em},
}
\textbf{Model} & \textbf{Layers} & \textbf{Width} & \textbf{Input Size} & \textbf{Patch Size} & \textbf{Training Data} \\
SAM H & 32 & 1280 & 1024x1024 & 16x16 & SA-1B \\
SAM L & 24 & 1024 & 1024x1024 & 16x16 & SA-1B \\
SAM B & 12 & 768 & 1024x1024 & 16x16 & SA-1B \\
\end{tblr}
}
\caption{SAM Models.}
\label{tab:supp_sam}
\end{table}

\paragraph{SAM} The objective of the Segment Anything Model (SAM) \cite{kirillov_segment_2023} is interactive segmentation with points, boxes, or text prompts as additional input. SAM was trained without the class label of the objects, but the text prompts (CLIP embeddings) enhanced SAM's understanding of the semantics. SAM was initialized from the MAE H model. Training was done on the SA-1B dataset, which was built by the SAM authors. SAM is an encoder-decoder design, we only took features from the encoder part. SAM's ViT architecture does not have a class token, we used global averaging pooling to replace the global token. We used the officially released model weights for SAM.

\paragraph{ImageNet} This fully supervised model was trained to predict ImageNet \cite{deng_imagenet_2009} labels, the training was done on ImageNet-1K from scratch without any pre-training. We used model weights released by PyTorch Hub. We used a model from the improved training recipe that covers state-of-the-art training tricks and augmentations. Specifically, we used the IMAGENET1K\_V1 weights, the base size model has 81.9 ImageNet accuracy, large size model has 79.7 accuracy.

\begin{table}[H]
\centering
\resizebox{0.99\linewidth}{!}{
\begin{tblr}{
  hline{1,4} = {-}{0.08em},
  hline{2} = {-}{0.05em},
}
\textbf{Model} & \textbf{Layers} & \textbf{Width} & \textbf{Input Size} & \textbf{Patch Size} & \textbf{Training Data} \\
ImageNet L & 24 & 1024 & 224x224 & 16x16 & IN-1K \\
ImageNet B & 12 & 768 & 224x224 & 16x16 & IN-1K \\
\end{tblr}
}
\caption{ImageNet Models.}
\label{tab:supp_imagenet}
\end{table}

\paragraph{DiNOv2} The authors describe DiNOv2 \cite{oquab_dinov2_2023} as DiNOv1 \cite{caron_emerging_2021} plus iBOT \cite{zhou_ibot_2022} with the centering of SwAV \cite{caron_unsupervised_2021-1}. DiNOv2 was trained with momentum self-distillation and mask reconstruction of latent tokens. The training was done on LVD-142M, which is a custom dataset made by the DiNOv2 authors. One notable difference to other models is that DiNOv2 smaller models were distilled from bigger models. We used the officially released model weights for DiNOv2.

\begin{table}[H]
\centering
\resizebox{0.99\linewidth}{!}{
\begin{tblr}{
  hline{1,5} = {-}{0.08em},
  hline{2} = {-}{0.05em},
}
\textbf{Model} & \textbf{Layers} & \textbf{Width} & \textbf{Input Size} & \textbf{Patch Size} & \textbf{Training Data} \\
DiNOv2 G & 40 & 1536 & 224x224 & 14x14 & LVD-142M \\
DiNOv2 L & 24 & 1024 & 224x224 & 14x14 & LVD-142M \\
DiNOv2 B & 12 & 768 & 224x224 & 14x14 & LVD-142M \\
\end{tblr}
}
\caption{DiNOv2 Models.}
\label{tab:supp_dinov2}
\end{table}

\paragraph{MoCov3} The Momentum Contrastive (MoCo) \cite{chen_empirical_2021} method trains contrastive loss with a momentum teacher encoder, which is an exponential moving average of the previous iteration models. The constrastive objective is to enforce the encoder to generate a similar representation to the momentum model. The training was done with the ImageNet-1K dataset. We used MoCov3 model weights released by MMPreTrain.

\begin{table}[H]
\centering
\resizebox{0.99\linewidth}{!}{
\begin{tblr}{
  hline{1,5} = {-}{0.08em},
  hline{2} = {-}{0.05em},
}
\textbf{Model} & \textbf{Layers} & \textbf{Width} & \textbf{Input Size} & \textbf{Patch Size} & \textbf{Training Data} \\
MoCov3 L & 24 & 1024 & 224x224 & 16x16 & IN-1K \\
MoCov3 B & 12 & 768 & 224x224 & 16x16 & IN-1K \\
MoCov3 S & 12 & 384 & 224x224 & 16x16 & IN-1K \\
\end{tblr}
}
\caption{MoCov3 Models.}
\label{tab:supp_mocov3}
\end{table}

\paragraph{MAE} The Mask Autoencoder (MAE) \cite{he_masked_2021} objective is to reconstruct the masked patches of input images given the un-masked patches, reconstruction is in the image space. The training was done on the ImageNet-1K dataset. MAE used an encoder and decoder design, we only studied the encoder part. We used the official release from the original authors.

\begin{table}[H]
\centering
\resizebox{0.99\linewidth}{!}{
\begin{tblr}{
  hline{1,5} = {-}{0.08em},
  hline{2} = {-}{0.05em},
}
\textbf{Model} & \textbf{Layers} & \textbf{Width} & \textbf{Input Size} & \textbf{Patch Size} & \textbf{Training Data} \\
MAE H & 32 & 1280 & 224x224 & 16x16 & IN-1K \\
MAE L & 24 & 1024 & 224x224 & 16x16 & IN-1K \\
MAE B & 12 & 768 & 224x224 & 16x16 & IN-1K \\
\end{tblr}
}
\caption{MAE Models.}
\label{tab:supp_mae}
\end{table}

\paragraph{SD} The Stable Diffusion (SD) \cite{rombach_high-resolution_2022} model's objective is to generate photo-realistic images. Although SD was trained without supervision on the loss term, the content of the generated image is controlled by a text prompt (CLIP embeddings), and the text prompt enhanced the semantic understanding of the features. SD is a U-Net and ResNet design with cross-attention to CLIP embeddings. There are 8 layers in the U-Net encoder and 12 layers in the decoder, skip connection connects the encoder and decoder blocks. There's no class token in SD, we used global averaging pooling to replace it. In the feature extraction, we used an empty text prompt, we followed the `inversion' time steps that chain the features of different time steps.
SD was trained on LAION-5B \cite{schuhmann_laion-5b_2022} dataset. We used the Huggingface release of the SD version 1.5 model. 

\begin{table}[H]
\centering
\resizebox{0.99\linewidth}{!}{
\begin{tblr}{
  hline{1,6} = {-}{0.08em},
  hline{2} = {-}{0.05em},
}
\textbf{\shortstack{Encoder\\ Layers}} & \textbf{\shortstack{Decoder\\ Layers}} & \textbf{Width} & \textbf{\shortstack{Feature\\ Size}} & \textbf{\shortstack{Input\\ Size}} & \textbf{Training Data} \\
1,2 & 10,11,12 & 320 & 64x64 & 512x512 & LAION-5B \\
3,4 & 7,8,9 & 640 & 32x32 & 512x512 & LAION-5B \\
5,6 & 4,5,6 & 1280 & 16x16 & 512x512 & LAION-5B \\
7,8 & 1,2,3 & 1280 & 8x8 & 512x512 & LAION-5B \\
\end{tblr}
}
\caption{Stable Diffusion Layers.}
\label{tab:supp_sd}
\end{table}

\newpage
\section{Extended Results}
\label{sec:supp_results}
In addition to the main results in Section 
4
, this appendix presents extended results that cover more brain ROIs, more ViT model sizes, more diffusion time steps, and more subjects. The structure of this appendix section follows the main results:

\begin{enumerate}
    \item \textbf{Brain Score}. Results on three subjects. Numerical results on more ROIs, all diffusion time steps.
    \item \textbf{Training Objectives and Brain-Net Alignment}. Consistency check. Display of raw layer selector weights. 
    \item \textbf{Network Hierarchy and Model Sizes}. Layer selector results in more ViT model size variants.
    \item \textbf{Fine-tuned Models}. Fine-tune performance score.
    \item \textbf{Channels and Brain ROIs}. Top channel image feature display on more brain ROIs.
\end{enumerate}

\subsection{Brain Score}
\label{sec:supp_brainscore}

\begin{figure*}[ht]
    \centering
    \includegraphics[width=0.99\linewidth]{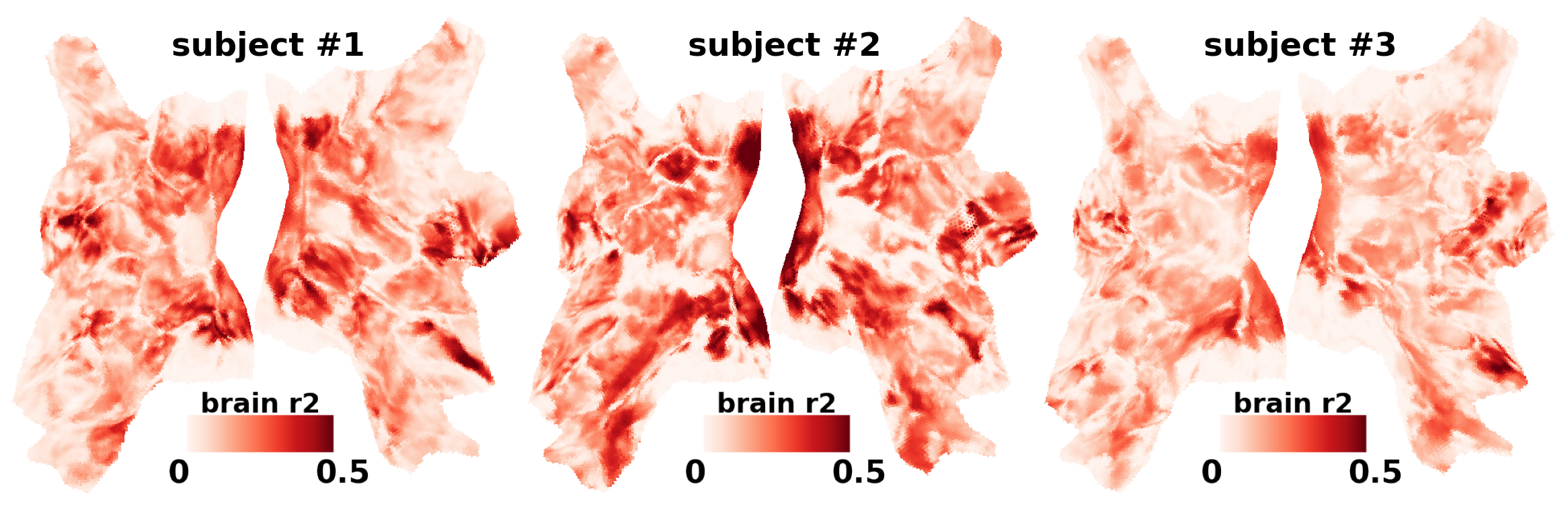}
    \captionof{figure}{\textbf{Brain Score}. Three subjects, CLIP L (base size 12 layer) model.}
    \label{fig:supp_brain_score}

\vspace{4mm}

\centering
\resizebox{0.99\linewidth}{!}{
    \begin{tblr}{
      column{even} = {r},
      column{3} = {r},
      column{5} = {r},
      column{7} = {r},
      column{9} = {r},
      column{11} = {r},
      column{13} = {r},
      cell{1}{2} = {c=13}{c},
      cell{3}{1} = {c=14}{},
      cell{7}{1} = {c=14}{},
      cell{14}{1} = {c=14}{},
      hline{1,26} = {-}{0.1em},
      hline{2} = {2-14}{},
      hline{3,7,14} = {-}{},
    }
                                             & \textbf{ROI Brain Score $R^2$ $\uparrow$ ($\pm$ 0.001)}            &                         &                         &                         &                         &                         &                         &                         &                         &                         &                         &                         &                         \\
    \textbf{Model}                                    & \textbf{all}                     & \textbf{V1}                      & \textbf{V2}                      & \textbf{V3}                      & \textbf{V4}                      & \textbf{EBA}                     & \textbf{FBA}                     & \textbf{OFA}                     & \textbf{FFA}                     & \textbf{OPA}                     & \textbf{PPA}                     & \textbf{OWFA}                    & \textbf{VWFA}                    \\
    \textit{CLIP model, three subjects}               &                         &                         &                         &                         &                         &                         &                         &                         &                         &                         &                         &                         &                         \\
    CLIP (subject \#1)                                    & 0.132                   & 0.216                   & 0.209                   & 0.185                   & 0.139                   & 0.176                   & 0.157                   & 0.129                   & 0.182                   & 0.091                   & 0.130                   & 0.121                   & 0.092                   \\
    CLIP (subject \#2)                                    & 0.154                   & 0.247                   & 0.183                   & 0.192                   & 0.188                   & 0.182                   & 0.134                   & 0.098                   & 0.136                   & 0.126                   & 0.199                   & 0.083                   & 0.140                   \\
    CLIP (subject \#3)                                    & 0.104                   & 0.155                   & 0.128                   & 0.108                   & 0.105                   & 0.134                   & 0.137                   & 0.080                   & 0.151                   & 0.081                   & 0.125                   & 0.104                   & 0.087                   \\
    \textit{ViT models, subject \#1}                  &                         &                         &                         &                         &                         &                         &                         &                         &                         &                         &                         &                         &                         \\
    CLIP                                     & \textbf{0.132}          & \textbf{\textit{0.216}} & \textbf{\textit{0.209}} & \textbf{\textit{0.185}} & \textbf{0.139}          & \textbf{0.176}          & \textbf{0.157}          & \textbf{0.129}          & \textbf{0.182}          & \textbf{\textit{0.091}} & \textbf{0.130}          & \textbf{\textit{0.121}} & \textbf{0.092}          \\
    SAM                                      & 0.110                   & 0.212                   & 0.197                   & 0.172                   & 0.113                   & 0.127                   & 0.120                   & 0.104                   & 0.142                   & 0.074                   & 0.104                   & 0.105                   & 0.066                   \\
    ImageNet                                 & 0.120                   & 0.205                   & 0.202                   & 0.174                   & 0.127                   & 0.159                   & 0.143                   & 0.117                   & 0.169                   & 0.076                   & 0.121                   & 0.109                   & 0.077                   \\
    DiNOv2                                     & 0.126                   & 0.208                   & 0.202                   & 0.175                   & 0.127                   & \textbf{\textit{0.174}} & \textbf{\textit{0.152}} & 0.122                   & \textbf{\textit{0.178}} & 0.083                   & \textbf{\textit{0.126}} & 0.111                   & \textbf{\textit{0.088}} \\
    MAE                                      & \textbf{\textit{0.129}} & \textbf{0.219}          & \textbf{0.210}          & \textbf{0.186}          & \textbf{\textit{0.135}} & 0.165                   & 0.148                   & \textbf{\textit{0.127}} & 0.173                   & \textbf{0.093}          & \textbf{\textit{0.126}} & \textbf{0.124}          & 0.086                   \\
    MoCov3                                     & 0.126                   & 0.214                   & 0.208                   & 0.181                   & 0.134                   & 0.163                   & 0.150                   & 0.120                   & 0.176                   & 0.086                   & 0.124                   & 0.115                   & 0.081                   \\
    \textit{Stable Diffusion time steps, subject \#1} &                         &                         &                         &                         &                         &                         &                         &                         &                         &                         &                         &                         &                         \\
    T0                                       & 0.048                   & 0.135                   & 0.112                   & 0.088                   & 0.057                   & 0.036                   & 0.033                   & 0.046                   & 0.036                   & 0.024                   & 0.041                   & 0.049                   & 0.025                   \\
    T5                                       & 0.062                   & 0.151                   & 0.130                   & 0.103                   & 0.070                   & 0.053                   & 0.050                   & 0.055                   & 0.056                   & 0.039                   & 0.055                   & 0.058                   & 0.033                   \\
    T10                                      & 0.077                   & 0.161                   & 0.146                   & 0.119                   & 0.078                   & 0.085                   & 0.079                   & 0.068                   & 0.091                   & 0.050                   & 0.071                   & 0.068                   & 0.044                   \\
    T15                                      & 0.095                   & 0.187                   & 0.169                   & 0.141                   & 0.097                   & 0.111                   & 0.105                   & 0.085                   & 0.123                   & 0.063                   & 0.090                   & 0.083                   & 0.055                   \\
    T20                                      & 0.106                   & 0.195                   & 0.184                   & 0.155                   & 0.110                   & 0.135                   & 0.120                   & 0.100                   & 0.142                   & 0.071                   & 0.104                   & 0.096                   & 0.063                   \\
    T25                                      & 0.112                   & 0.199                   & 0.191                   & 0.161                   & 0.109                   & 0.149                   & 0.127                   & 0.109                   & 0.151                   & 0.076                   & 0.112                   & 0.103                   & 0.068                   \\
    T30                                      & 0.121                   & 0.207                   & 0.202                   & 0.177                   & \textbf{\textit{0.129}} & 0.163                   & 0.138                   & 0.118                   & 0.163                   & 0.080                   & 0.121                   & 0.114                   & 0.076                   \\
    T35                                      & \textbf{0.125}          & 0.212                   & 0.205                   & 0.178                   & 0.128                   & \textbf{0.170}          & \textbf{0.145}          & \textbf{\textit{0.123}} & \textbf{\textit{0.169}} & \textbf{0.084}          & \textbf{0.126}          & 0.118                   & \textbf{0.083}          \\
    T40                                      & 0.123                   & \textbf{0.215}          & \textbf{\textit{0.207}} & 0.177                   & 0.123                   & \textbf{\textit{0.169}} & 0.143                   & 0.120                   & \textbf{\textit{0.169}} & 0.080                   & 0.123                   & 0.116                   & 0.075                   \\
    T45                                      & \textbf{0.125}          & \textbf{0.215}          & \textbf{0.208}          & \textbf{0.181}          & \textbf{0.130}          & \textbf{0.170}          & \textbf{0.145}          & \textbf{0.124}          & \textbf{0.170}          & \textbf{\textit{0.082}} & \textbf{\textit{0.125}} & \textbf{\textit{0.119}} & 0.078                   \\
    T50                                      & \textbf{\textit{0.124}} & \textbf{\textit{0.213}} & \textbf{\textit{0.207}} & \textbf{\textit{0.179}} & 0.124                   & \textbf{\textit{0.169}} & \textbf{\textit{0.144}} & \textbf{\textit{0.123}} & 0.168                   & \textbf{\textit{0.082}} & 0.124                   & \textbf{0.120}          & \textit{\textbf{0.081}} 
    \end{tblr}
    }
    \captionof{table}{\textbf{Brain Score}. ViT models are base size 12-layer. \textbf{Bold} marks best within each category, \textbf{\textit{bold italic}} marks the second best. \textbf{\textit{Top}}: CLIP model on three subjects. \textit{\textbf{Middle}}: ViT models on subject \#1. \textit{\textbf{Bottom}}: Stable Diffusion model time steps on subject \#1. \textit{\textbf{Insights}}: 1) individual difference exists, subject \#3's early visual cortex is significantly less predictable. 2) CLIP and DiNOv2 are better for late brain regions, and MAE is better for the early visual cortex. 3) Stable Diffusion T35 is better for late brain regions, and T45 is better for early visual cortex.}
    \label{tab:supp_brainscore}
    
\end{figure*}

In addition to the brain score reported in main results Section 
4.1
, we report 1) CLIP brain score on three subjects, 2) Numerical brain score results of ViT base size model on more ROIs, and 3) Stable Diffusion brain score results that cover the full time-step range. Also, in main results we reported the root summed square difference of brain score, in this appendix, we report the raw ROI-averaged brain score.

\paragraph{Three subjects} In \Cref{fig:supp_brain_score} and \Cref{tab:supp_brainscore}, subject \#2 has a more predictable V1 while subject \#3 has a least predictable early visual cortex. Subject \#1 has a most predictable FFA and FBA. The prediction score difference matches the brain-to-network mapping results that subject \#3 has large uncertainty in the early visual cortex (Figure \ref{fig:supp_random_seed_consistency}), and subject \#2 and \#3 are missing the FFA region that subject \#1 has. Overall, individual difference is expected and consistent.

\paragraph{ViT models} In \Cref{tab:supp_brainscore}, we report the raw ROI-average brain score. Among the ViT models, MAE has the best prediction power in early visual (V1 to V3) and navigation and spatial-relation region OPA. Interestingly, MAE has the best score in word and letter region OWFA but not for VWFA. CLIP has the best score in face, body, and scene-related regions (EBA, FBA, OFA, FFA, PPA) followed by DiNOv2. 


\paragraph{Diffusion time steps} In \Cref{tab:supp_brainscore}, we report brain score fixing each diffusion time step. $T<25$ showed a sub-optimal performance score in all regions. $T=35$ showed the best performance on high-level regions (EBA, FBA, OPA, PPA), and $T=45$ showed good performance for all regions from early visual to high-level. Surprisingly, $T=0$ achieved relatively good brain score for the early visual ROIs.

\subsection{Training Objectives and Brain-Net Alignment}
\label{sec:supp_align}

In this section, we show: 1) consistency of brain-net alignment across random seeds, and 2) expanded raw layer selector weights.

\begin{figure}[h]
  \centering
  \begin{subfigure}[b]{0.48\textwidth}
    \includegraphics[width=\textwidth]{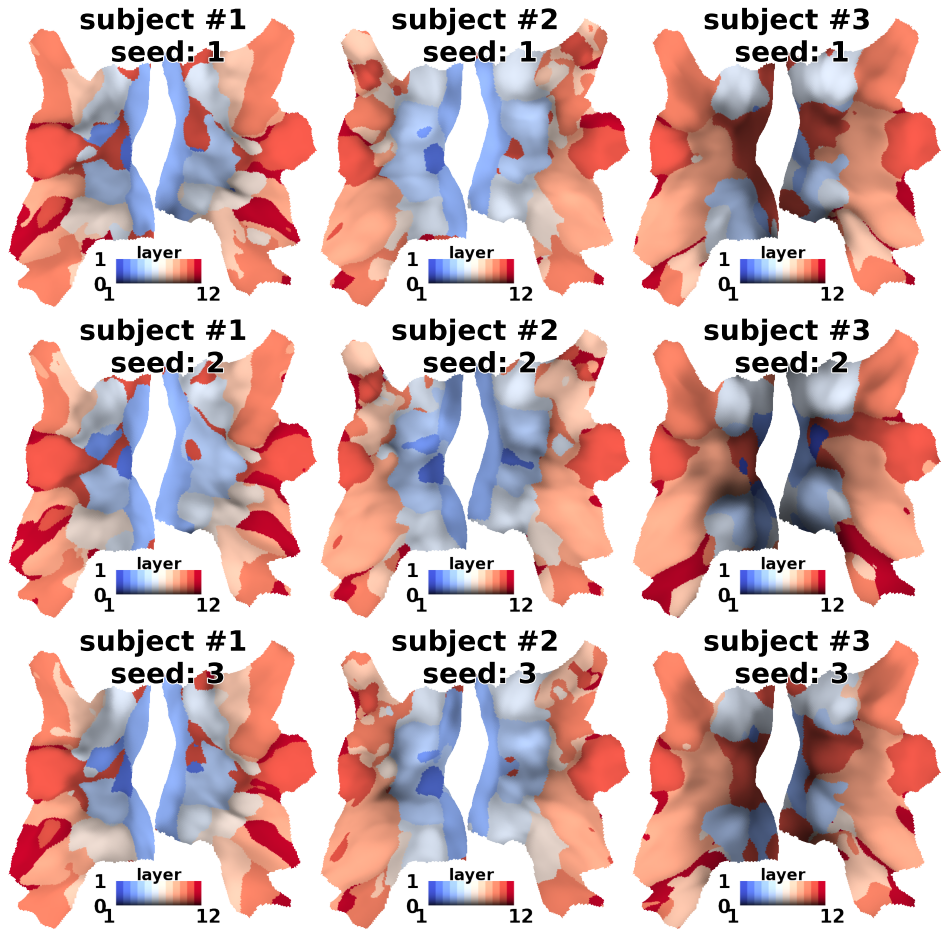}
    \caption{CLIP}
    \label{fig:baa}
  \end{subfigure}
  \hfill
  \vspace{2mm}
  \begin{subfigure}[b]{0.48\textwidth}
    \includegraphics[width=\textwidth]{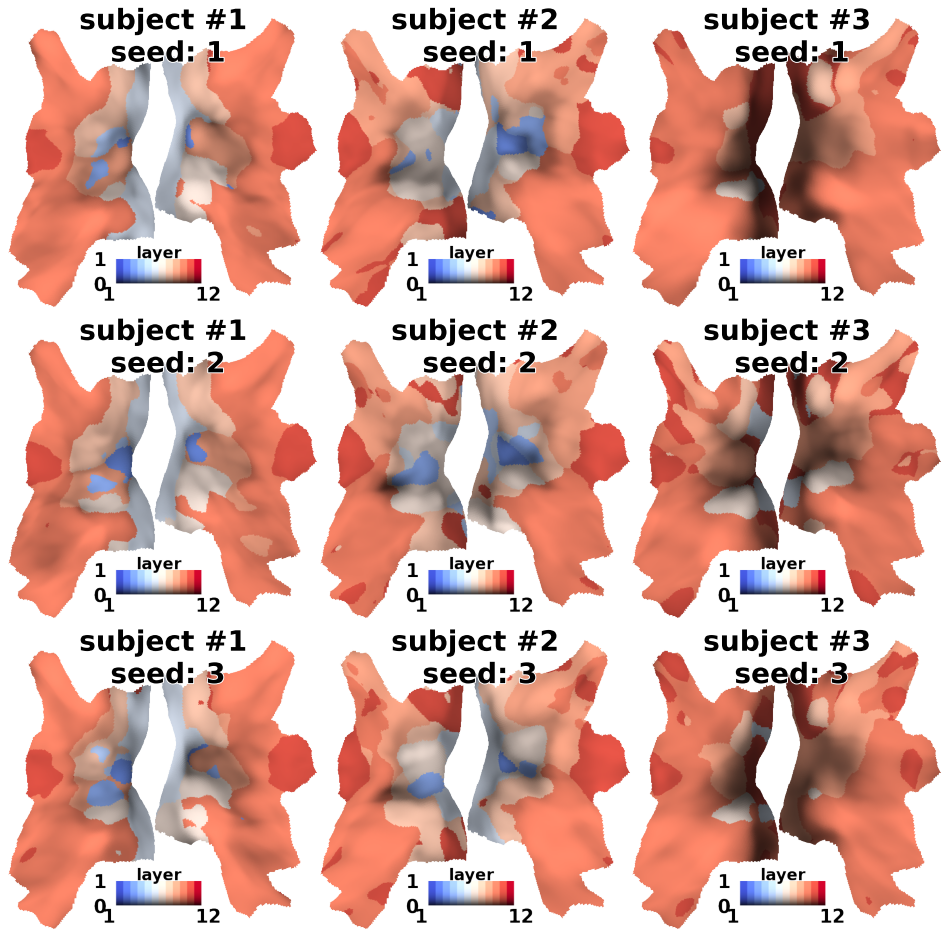}
    \caption{MAE}
    \label{fig:bab}
  \end{subfigure}
  
  \caption{\textbf{Random Seed Consistency}. CLIP (up) and MAE (down) model, 3 subjects (columns) and 3 random seeds (rows).}
  \label{fig:supp_random_seed_consistency}
\end{figure}

\paragraph{Random seed consistency} In the main results Section 
4.2
we found consistent differences across subjects. In this experiment, we repeated the same model and subject for 3 different random seeds. Results are in \Cref{fig:supp_random_seed_consistency}, we found subjects \#1 and \#2 had consistent brain-to-layer mapping across random seeds. Subject \#3 was less consistent across random seeds, note that subject \#3 also had the lowest data quality (brain score, \Cref{tab:supp_brainscore}).

\begin{figure*}[ht]
    \centering
    \includegraphics[width=\linewidth]{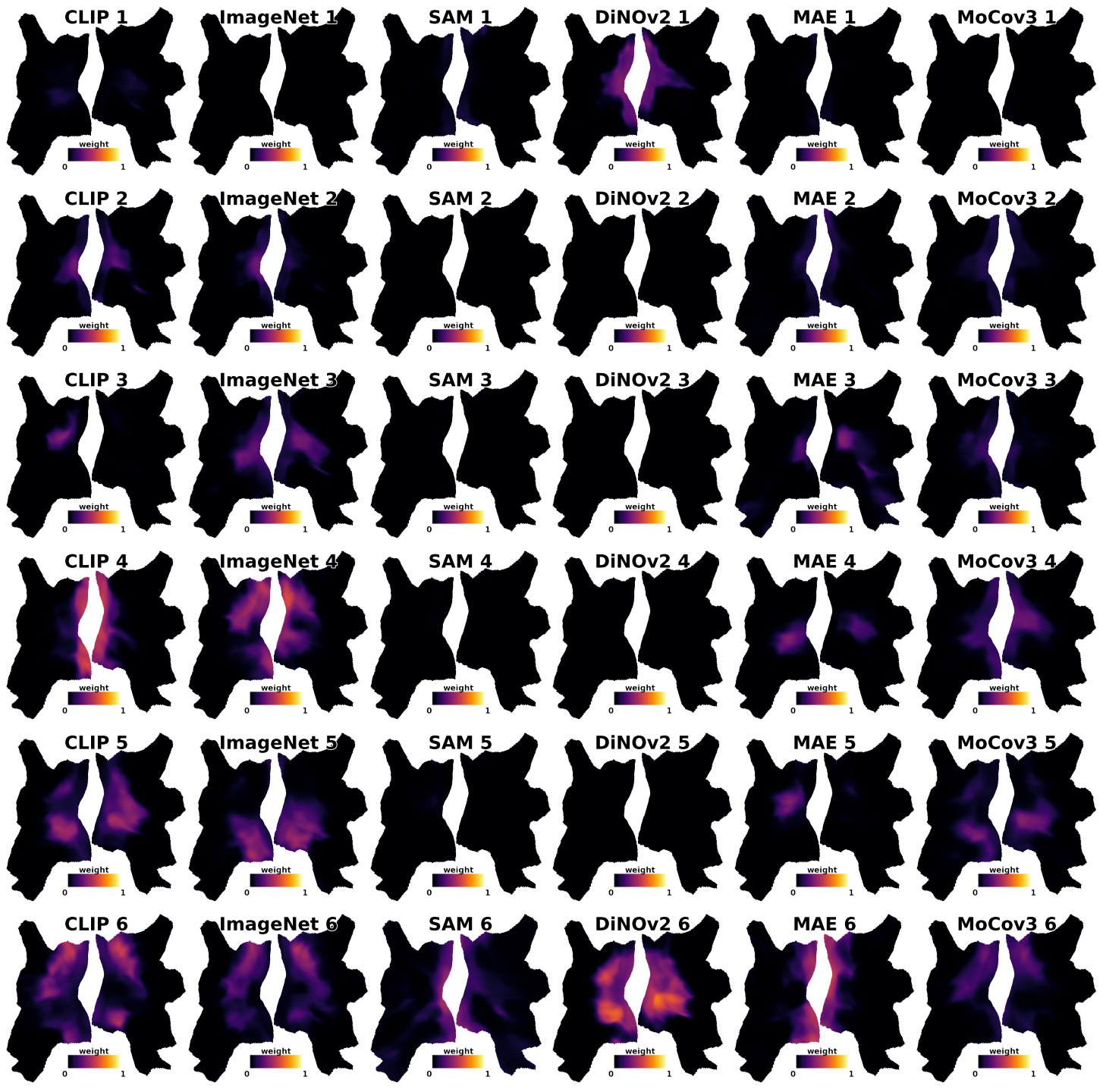}
    \captionof{figure}{\textbf{Raw Layer Selector weights (Part 1)}. Layer 1 to 6 of  ViT base size 12-layer models. The number tailing model name is the layer index.}
    \label{fig:supp_vit_raw_p1}
\end{figure*}

\begin{figure*}[ht]
    \centering
    \includegraphics[width=\linewidth]{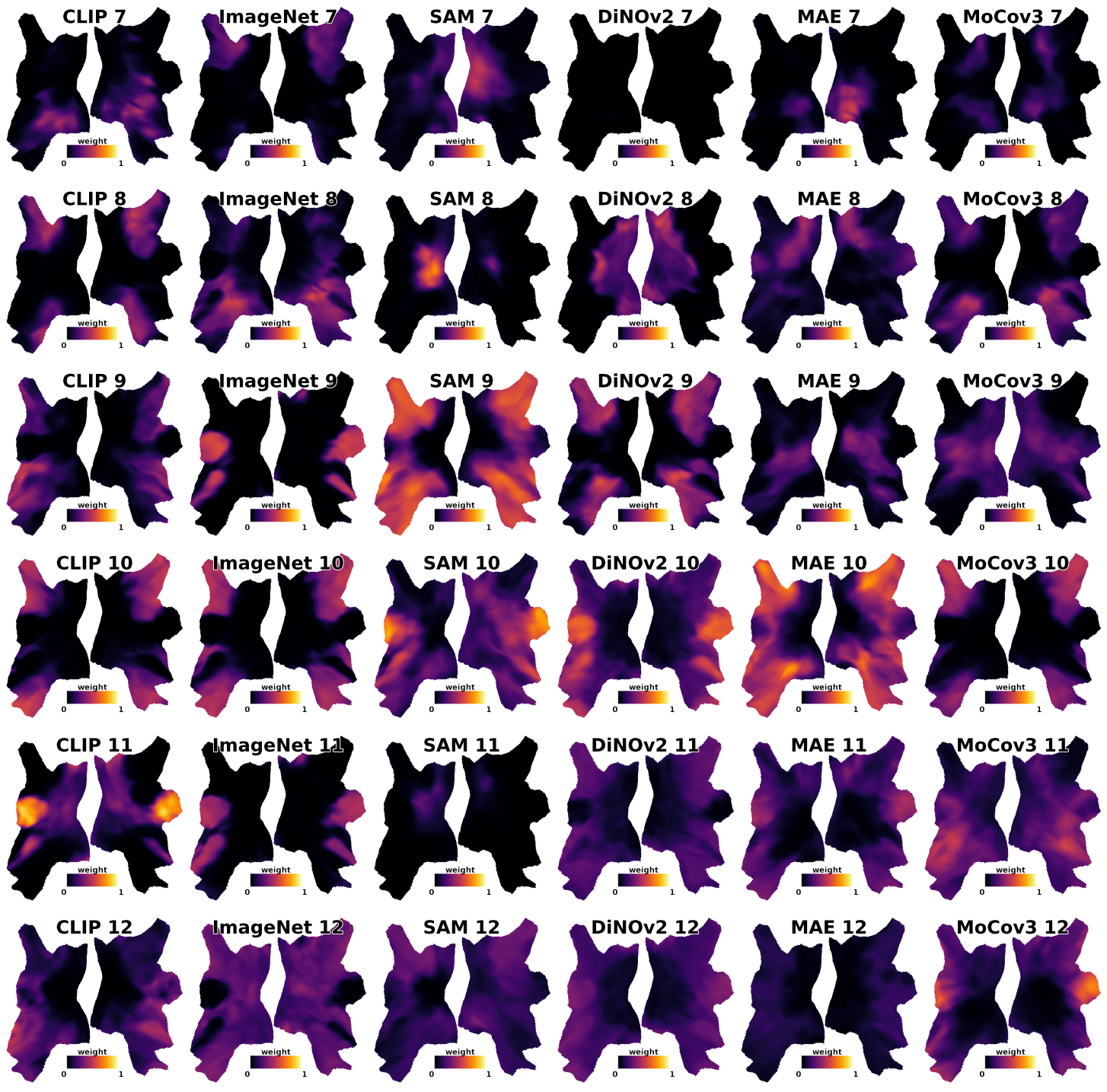}
    \captionof{figure}{\textbf{Raw Layer Selector weights (Part 2)}. Layer 7 to 12 of ViT base size 12-layer models. The number tailing model name is the layer index.}
    \label{fig:supp_vit_raw_p2}
\end{figure*}

\begin{figure*}[ht]
    \centering
    \includegraphics[width=\linewidth]{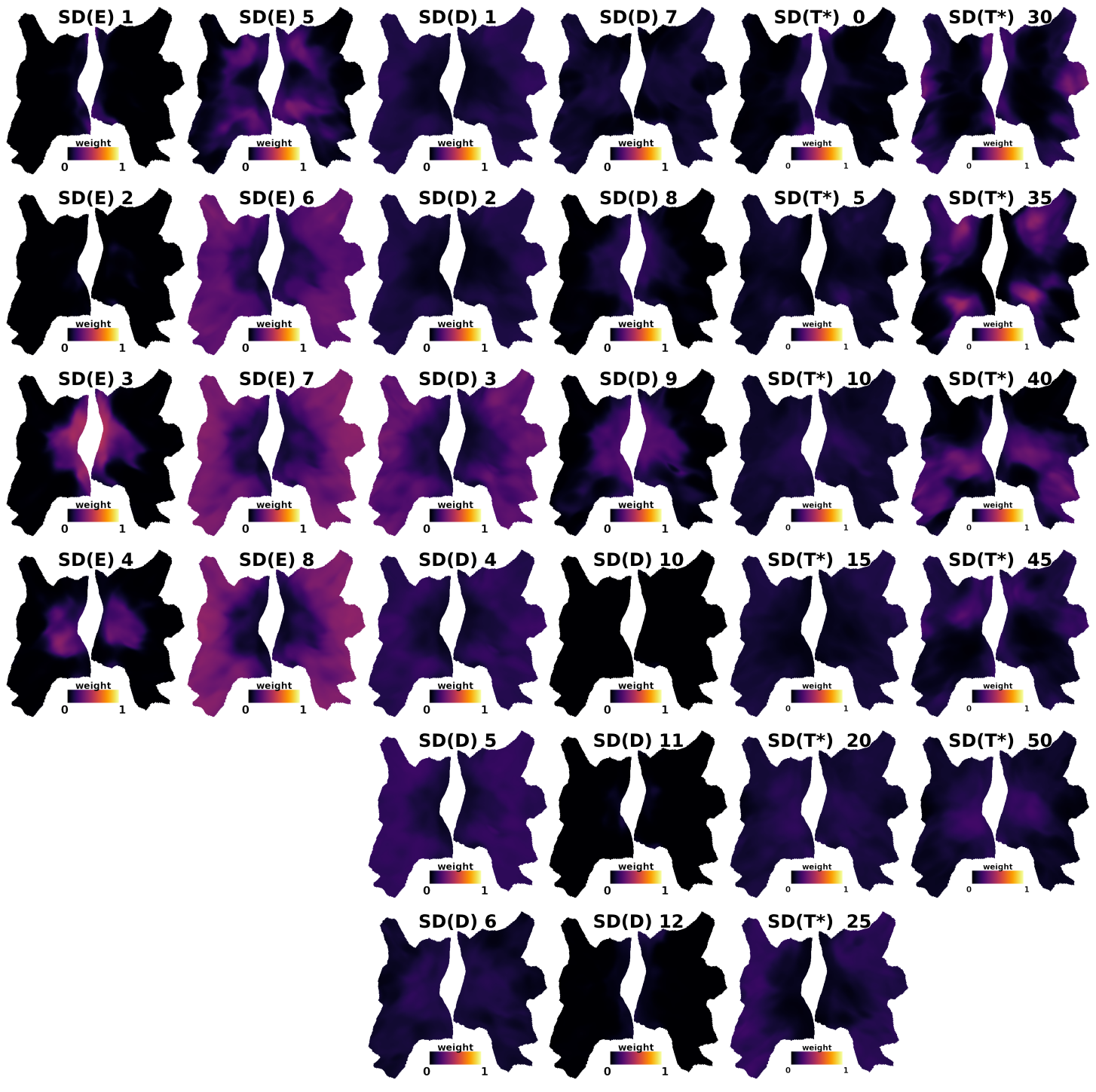}
    \captionof{figure}{\textbf{Raw Layer Selector weights (Part 3)}. Stable Diffusion model. SD(E): 8 encoder layers (fixed at T=40). SD(D): 12 decoder layers (fixed at T=40). SD(T*): 50 time steps (fixed at decoder layer no. 6). The number tailing model name is the layer or time step index.}
    \label{fig:supp_sd_raw_p3}
\end{figure*}

\paragraph{Raw layer selector weights} In our main results Sections 
3.2
and 
4.2
, we displayed argmax and confidence of selected layers. In \Cref{fig:supp_vit_raw_p1}-\ref{fig:supp_sd_raw_p3}, we display the raw output of layer selector weights for 1) 6 ViT base size 12-layer models, and 2) Stable Diffusion model fix time step T40 layer selection and fix decoder layer 6 time-step selection. 

There are some interesting observations that are hard to conclude from the argmax plot but more visible in the raw weights: 1) CLIP layer 11 is strongly aligned to EBA but also weakly aligned to the mid-level dorsal stream. 2) ImageNet's last layer is weakly aligned to all regions expect EBA and FFA. 3) SAM's last layer is weakly aligned to the mid-to-high level dorsal stream and mid-level ventral stream. 4) DiNOv2's last two layers' alignment weakly follows layer 10. 5) MAE layer 10 strongly aligns to mid-to-high level dorsal and ventral stream, MAE last layer does not align to any brain regions. 6) MoCov3 layer 11 aligns with the late ventral stream but not the dorsal stream, and MoCov3's layer 12 aligns with EBA.

\clearpage

\clearpage

\clearpage



\subsection{Network Hierarchy and Model Sizes}
\label{sec:sizes}

In this section, we expand the main results Section 
4.3
brain-layer alignment display to include more size variants. Details for pre-trained models, including layer, width, input size, patch size, and training data, are in \Cref{sec:model_details}.

\paragraph{CLIP} CLIP models showed increasing brain-net alignment as they scaled up both data and size. Both early and late layers in larger CLIP models are more selected by the brain. Notable, CLIP (M) and CLIP (S) were trained with the same model size but $\times10$ smaller training samples, CLIP (S) showed low confidence selection for the whole visual brain and only the late layers were more selected. 

\begin{figure}[H]
    \centering
    \includegraphics[width=0.76\linewidth]{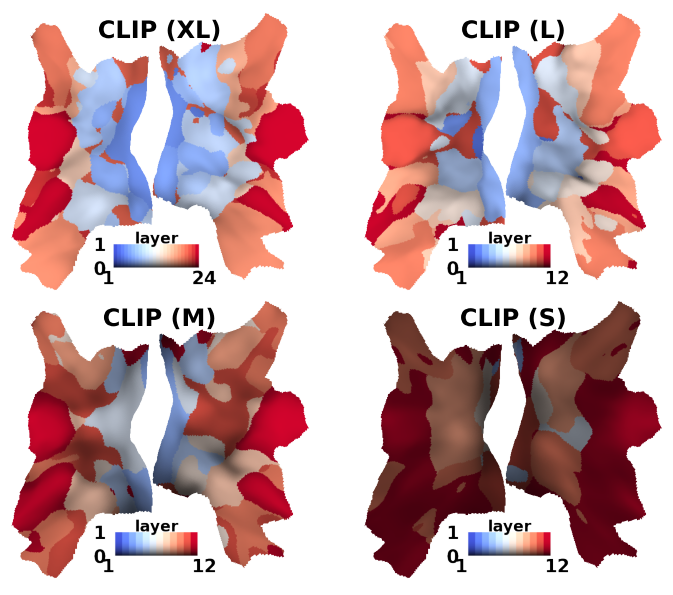}
    \captionof{figure}{\textbf{CLIP Brain-Net Alignment}. XL to M are size and data variants, M and S are the same size but have smaller training data.}
    \label{fig:supp_clip_size}
\end{figure}

\paragraph{SAM} SAM models showed decreasing brain-net alignment as they scaled up sizes. Larger SAM models' late layers were not selected; SAM's early layers were not selected in all model sizes. The uncertainty of selection went up in the early visual cortex for larger SAM models. 

\begin{figure}[H]
    \centering
    \includegraphics[width=0.99\linewidth]{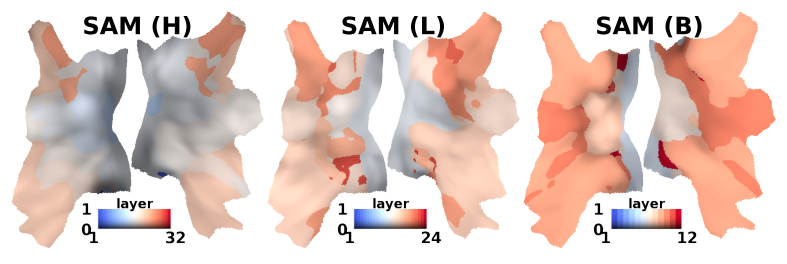}
    \captionof{figure}{\textbf{SAM Brain-Net Alignment}. Size variants, same training data.}
    \label{fig:supp_sam_size}
\end{figure}

\paragraph{ImageNet} ImageNet models showed decreasing brain-net alignment as the size scales up. Base size ImageNet model's early and late were both selected, larger size ImageNet model's early layers were not selected.

\begin{figure}[H]
    \centering
    \includegraphics[width=0.8\linewidth]{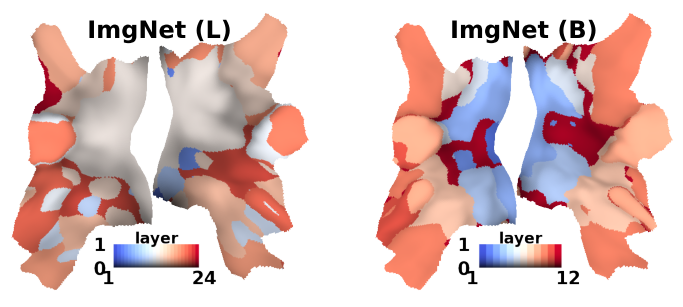}
    \captionof{figure}{\textbf{ImageNet Brain-Net Alignment}. Size variants, same training data.}
    \label{fig:supp_imagenet_size}
\end{figure}

\paragraph{DiNOv2} DiNOv2 models showed decreasing brain-net alignment when scaled up. Larger DiNOv2 models' early layers were less selected, only the last 1/4 of the layers were selected for the gigantic model. The first 1/2 of the layers were not selected for DiNOv2 models of all sizes. 

\begin{figure}[H]
    \centering
    \includegraphics[width=0.99\linewidth]{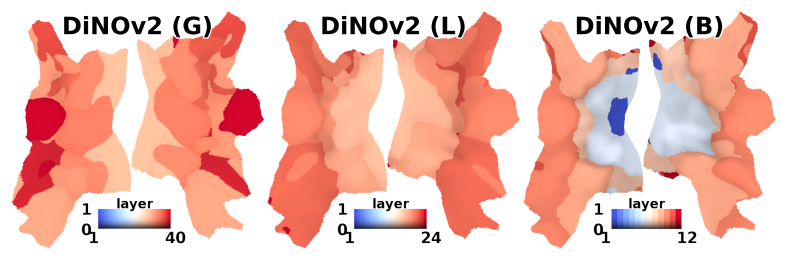}
    \captionof{figure}{\textbf{DiNOv2 Brain-Net Alignment}. Size variants, same training data.}
    \label{fig:supp_dino_size}
\end{figure}

\paragraph{MAE} MAE models showed increasing brain-net alignment from base to large, decreasing from large to huge. MAE's early layers were not selected for the base size model, selected for the large and huge size models. MAE's late layers were not selected for the huge size model, selected for the base and large models. The huge model had more separation of semantic brain regions.

\begin{figure}[H]
    \centering
    \includegraphics[width=0.99\linewidth]{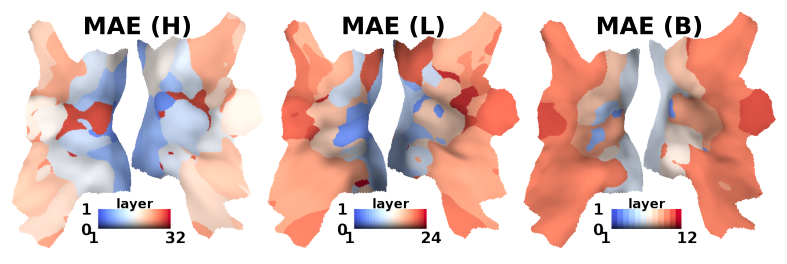}
    \captionof{figure}{\textbf{MAE Brain-Net Alignment}. Size variants, same training data.}
    \label{fig:supp_mae_size}
\end{figure}

\paragraph{MoCov3} MoCov3 showed decreasing brain-net alignment as size scales up. MoCov3's late layers were more selected for small and base size models, and MoCov3's late layers were significantly less selected for large size models.

\begin{figure}[H]
    \centering
    \includegraphics[width=0.99\linewidth]{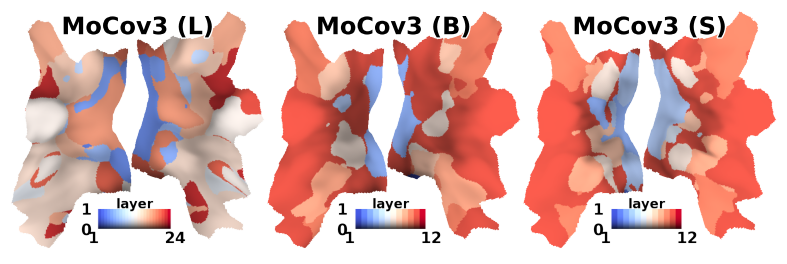}
    \captionof{figure}{\textbf{MoCov3 Brain-Net Alignment}. Size variants, same training data.}
    \label{fig:supp_moco_size}
\end{figure}

\subsection{Fine-tuned Model}
\label{sec:supp_finetune}
In the main results Section 
4.4 
, we attached an MLP prediction head to the last layer class token and fine-tuned the whole model to ISIC and EuroSAT tasks. In our main results, we found CLIP to maintain its computation layouts after fine-tuning while SAM and DiNOv2 re-wired their late layers and surfer from catastrophic forgetting.

\begin{figure}[h]
    \centering
    \includegraphics[width=\linewidth]{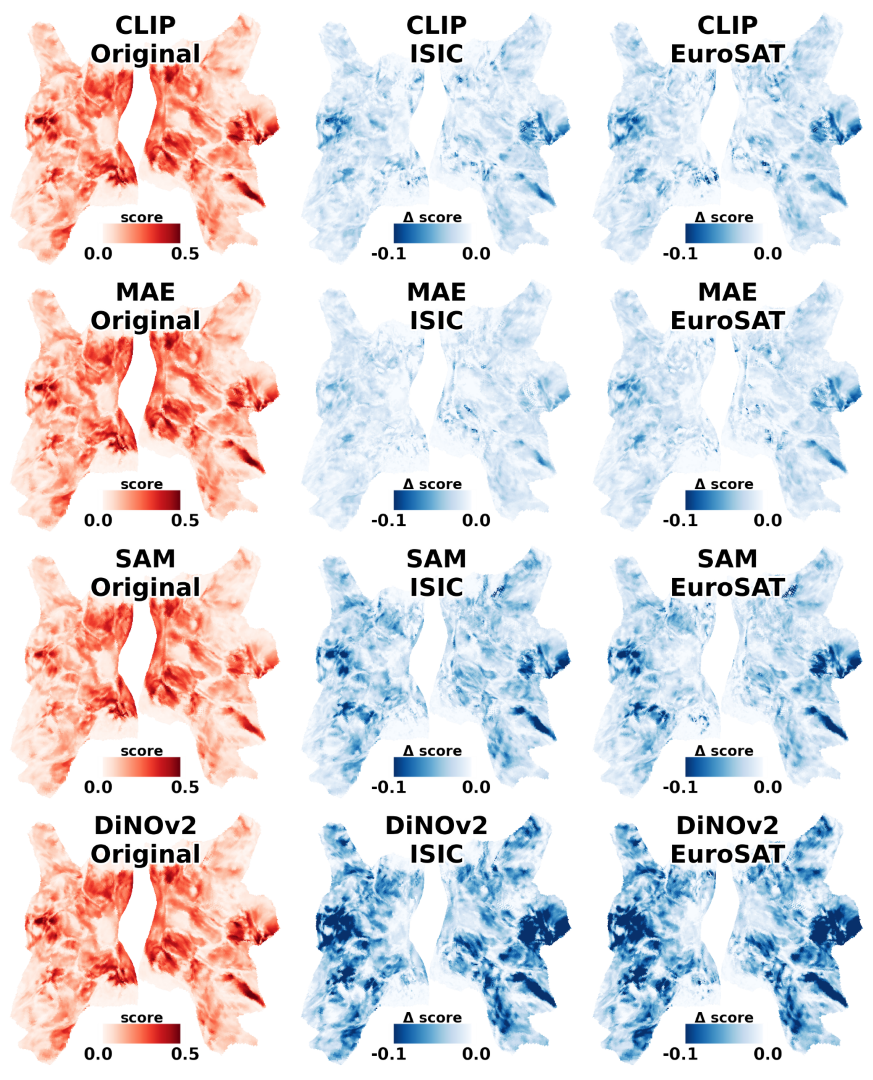}
    \captionof{figure}{Brain score before and after fine-tuning on small datasets (ISIC, EuroSAT). Brain score of CLIP dropped less compare to DiNOv2 and SAM. CLIP suffer less from catastrophic forgetting.}
    \label{fig:supp_ft_brain_score}
\end{figure}

\paragraph{Brain score after fine-tuning} 
In this appendix, we quantitatively compare brain score before and after fine-tuning. In Figure \ref{fig:supp_ft_brain_score}. Brain score of CLIP dropped from 0.131 to 0.115 after fine-tuning, DiNOv2 dropped from 0.128 to 0.085, SAM dropped from 0.111 to 0.086. The fact that CLIP dropped less brain score further support the observation that CLIP maintain computation layouts.

\paragraph{Fine-tune last layer performance}  
In this appendix, we further reported the fine-tuning performance score in \Cref{tab:supp_finetune}. CLIP had the best performance overall. Interestingly, SAM and DiNOv2 also had competitive performance despite their late layers being mostly re-wired (Section 4.4). We found the fine-tuning performance score does not correlate to the changes in brain alignments.

\begin{table}[H]
\centering
\resizebox{0.99\linewidth}{!}{
\begin{tblr}{
    hline{1,5} = {-}{0.1em},
    hline{2} = {2-5}{0.05em},
    hline{3} = {0}{0.05em},
    cell{1}{2} = {c=4}{c},
}
        & \textbf{Fine-tuned Accuracy} $\uparrow$ &       &                         &                         \\
\textbf{Dataset} \quad\quad\quad\quad\quad\quad & CLIP                & MAE   & SAM                     & DiNOv2                  \\
ISIC ($\pm 0.008$)    & \textbf{0.640}      & 0.589 & \textbf{\textit{0.627}} & 0.622                   \\
EuroSAT ($\pm 0.004$) & \textbf{0.954}      & 0.936 & 0.885                   & \textbf{\textit{0.946}} 
\end{tblr}
}
\caption{Fine-tuned performance score. Average of 10 runs. The whole model is fine-tuned with the prediction head attached to the last layer class token.}
\label{tab:supp_finetune}
\end{table}

\begin{figure}[H]
    \centering
    \includegraphics[width=0.99\linewidth]{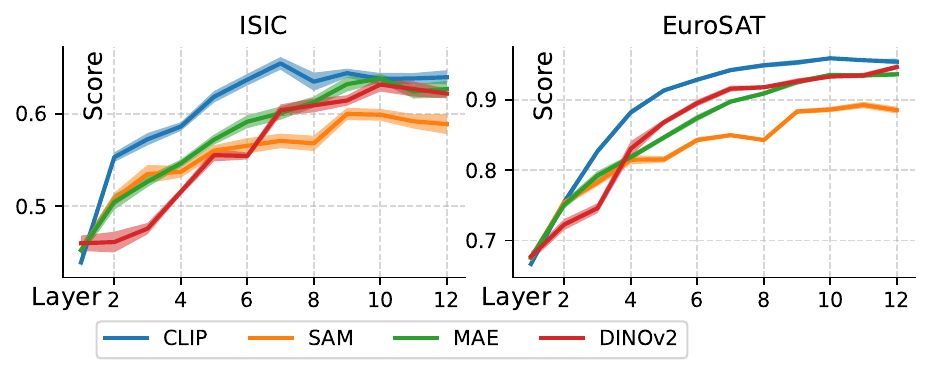}
    \captionof{figure}{Grid search of fine-tuned layer. Average of 10 runs. The prediction head is attached to one layer, later layers are discarded.}
    \label{fig:supp_fintune_grid}
\end{figure}

\paragraph{Grid search on which layer to fine-tune} In the main results, we stated that ``ISIC requires low-level features'', we verify this statement in this appendix. In this experiment, we ran a grid search that attached the prediction head to each layer, layers before the prediction layer are trained, and layers after the prediction layer are discarded. In \Cref{fig:supp_fintune_grid}, on ISIC, we found CLIP layer 7 reached peak performance, and other models also peaked at mid-to-late layers; on EuroSAT, all models' performance peaked at the last or second-last layer. Overall, the ISIC task relies on low-level features, EuroSAT task relies on high-level features.

\newpage
\subsection{Channels and Brain ROIS}
\label{sec:supp_channel}

\begin{figure*}[t]
    \centering

    \begin{subfigure}[b]{\textwidth}
        \includegraphics[width=\textwidth]{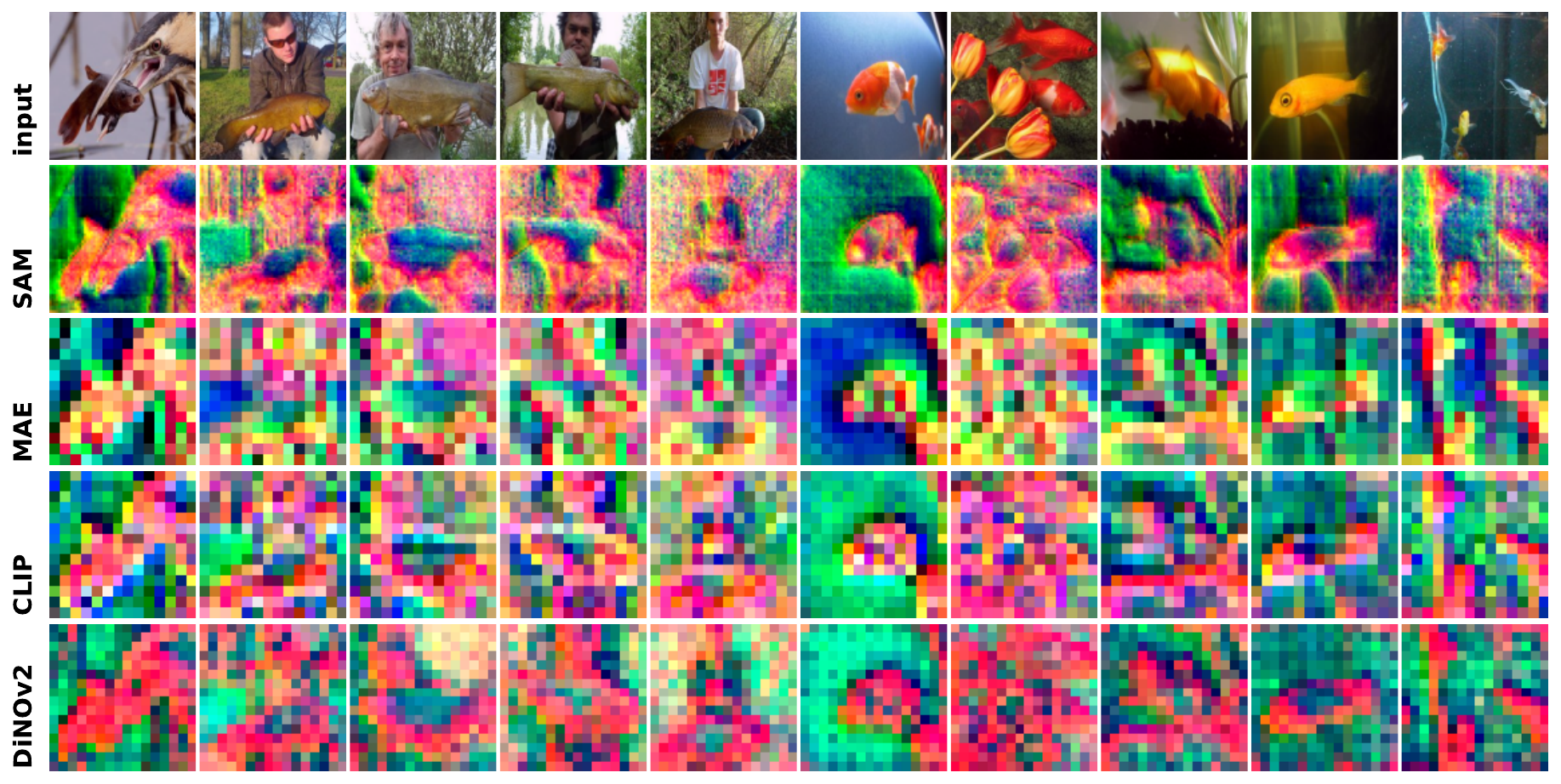}
        \caption{\textbf{V1} (early visual)}
    \end{subfigure}

    \begin{subfigure}[b]{\textwidth}
        \includegraphics[width=\textwidth]{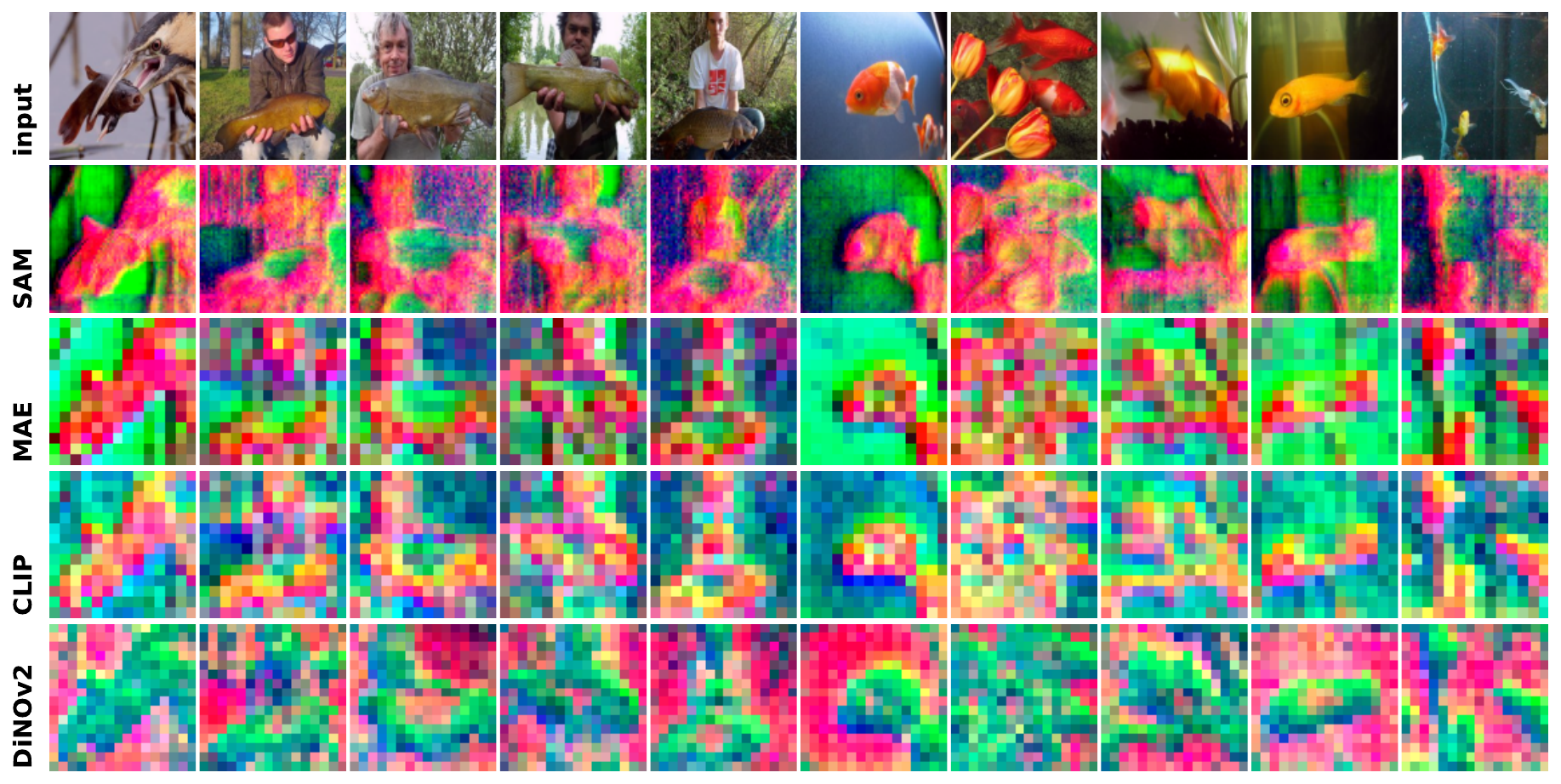}
        \caption{\textbf{V2} (early visual)}
    \end{subfigure}

\caption{Top 3 selected channels for voxels in one brain ROI (methods in \Cref{sec:code}, findings in \Cref{sec:supp_channel}).}
\label{fig:supp_top_channels1}
\end{figure*}

\begin{figure*}[t]
    \centering

    \begin{subfigure}[b]{\textwidth}
        \includegraphics[width=\textwidth]{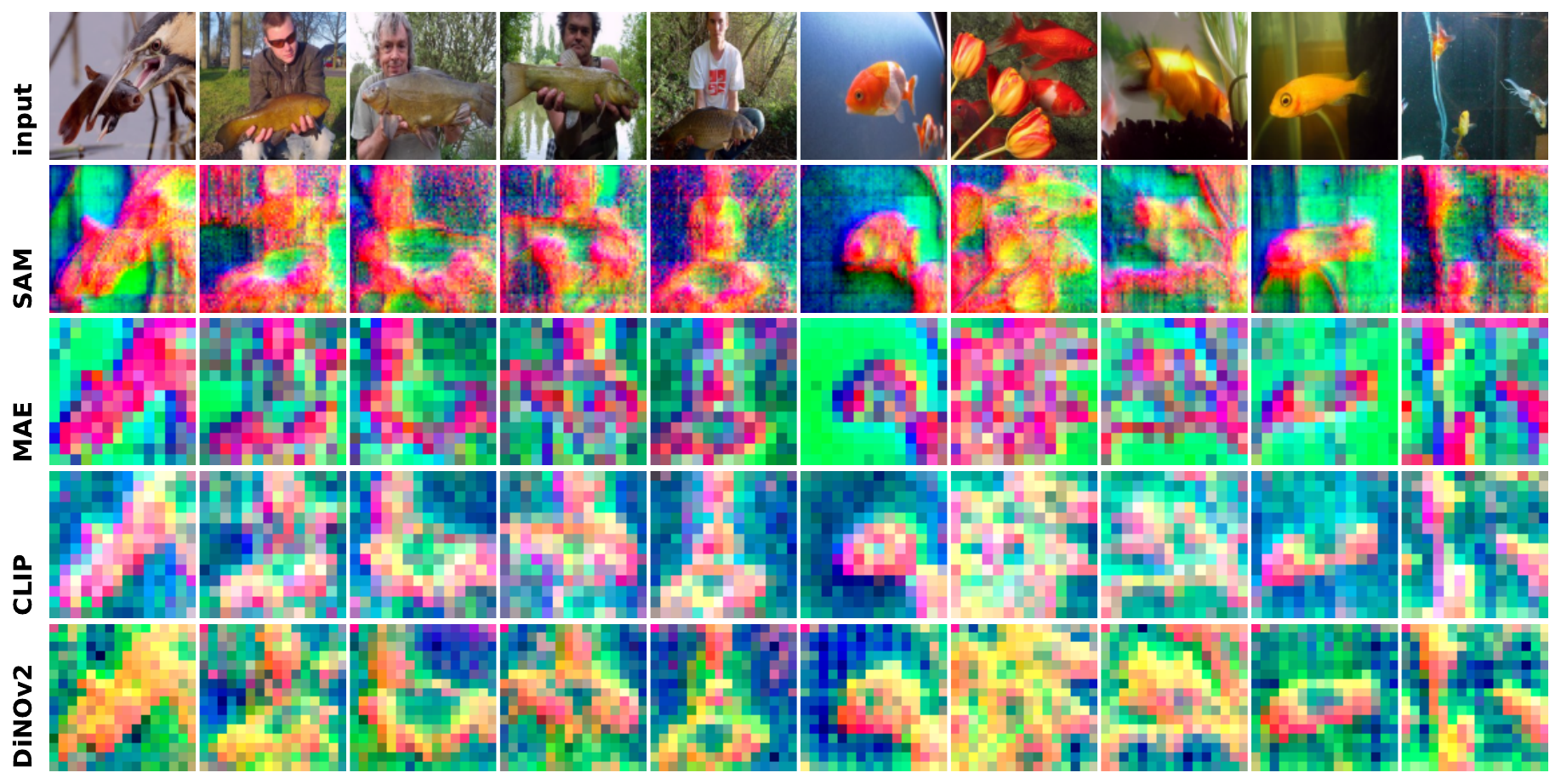}
        \caption{\textbf{V3} (early visual)}
    \end{subfigure}

    \begin{subfigure}[b]{\textwidth}
        \includegraphics[width=\textwidth]{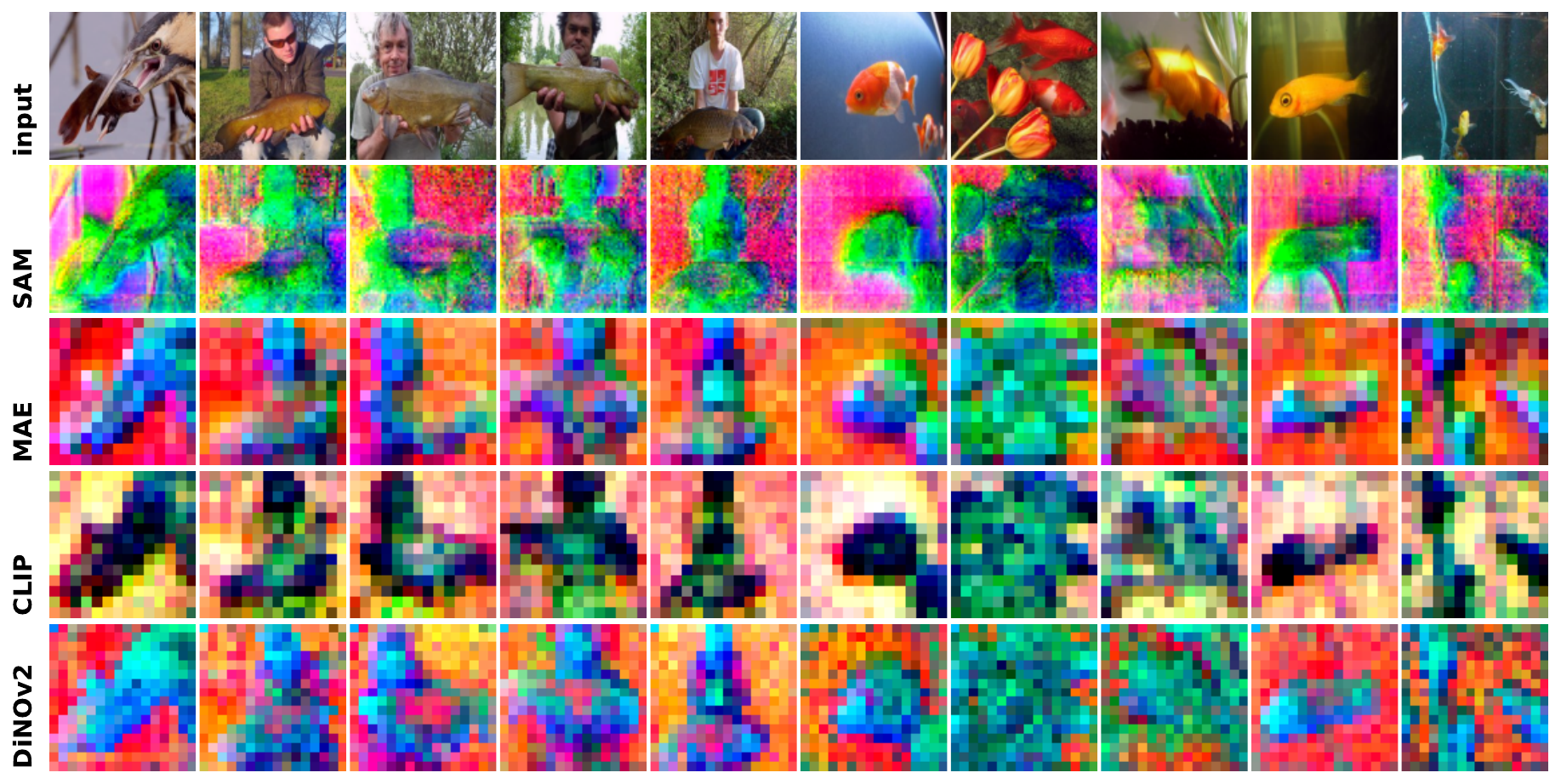}
        \caption{\textbf{V4} (mid-level)}
    \end{subfigure}

\caption{Top 3 selected channels for voxels in one brain ROI (methods in \Cref{sec:code}, findings in \Cref{sec:supp_channel}).}
\label{fig:supp_top_channels2}
\end{figure*}

\begin{figure*}[t]
    \centering

    \begin{subfigure}[b]{\textwidth}
        \includegraphics[width=\textwidth]{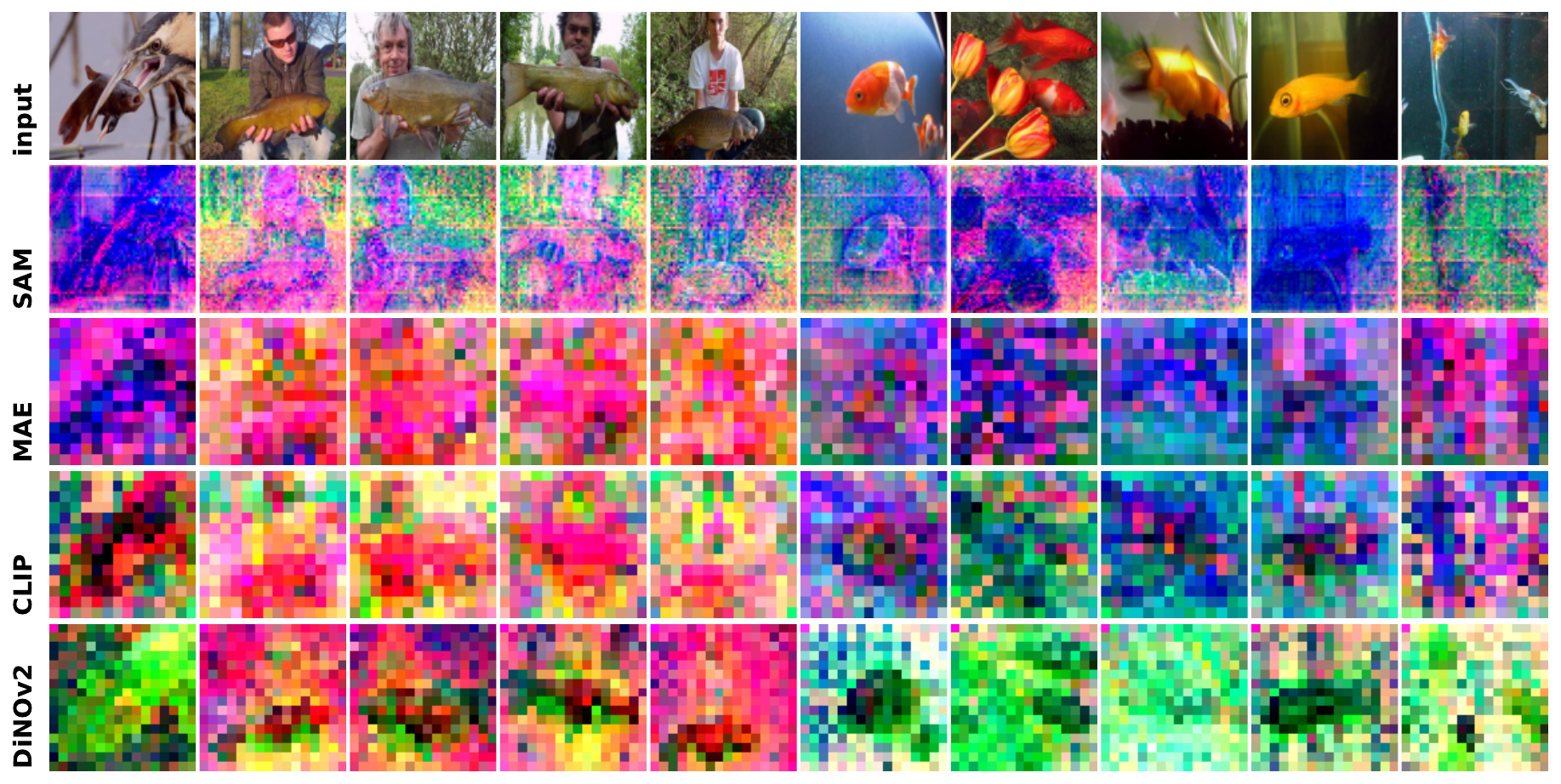}
        \caption{\textbf{EBA} (body)}
    \end{subfigure}

    \begin{subfigure}[b]{\textwidth}
        \includegraphics[width=\textwidth]{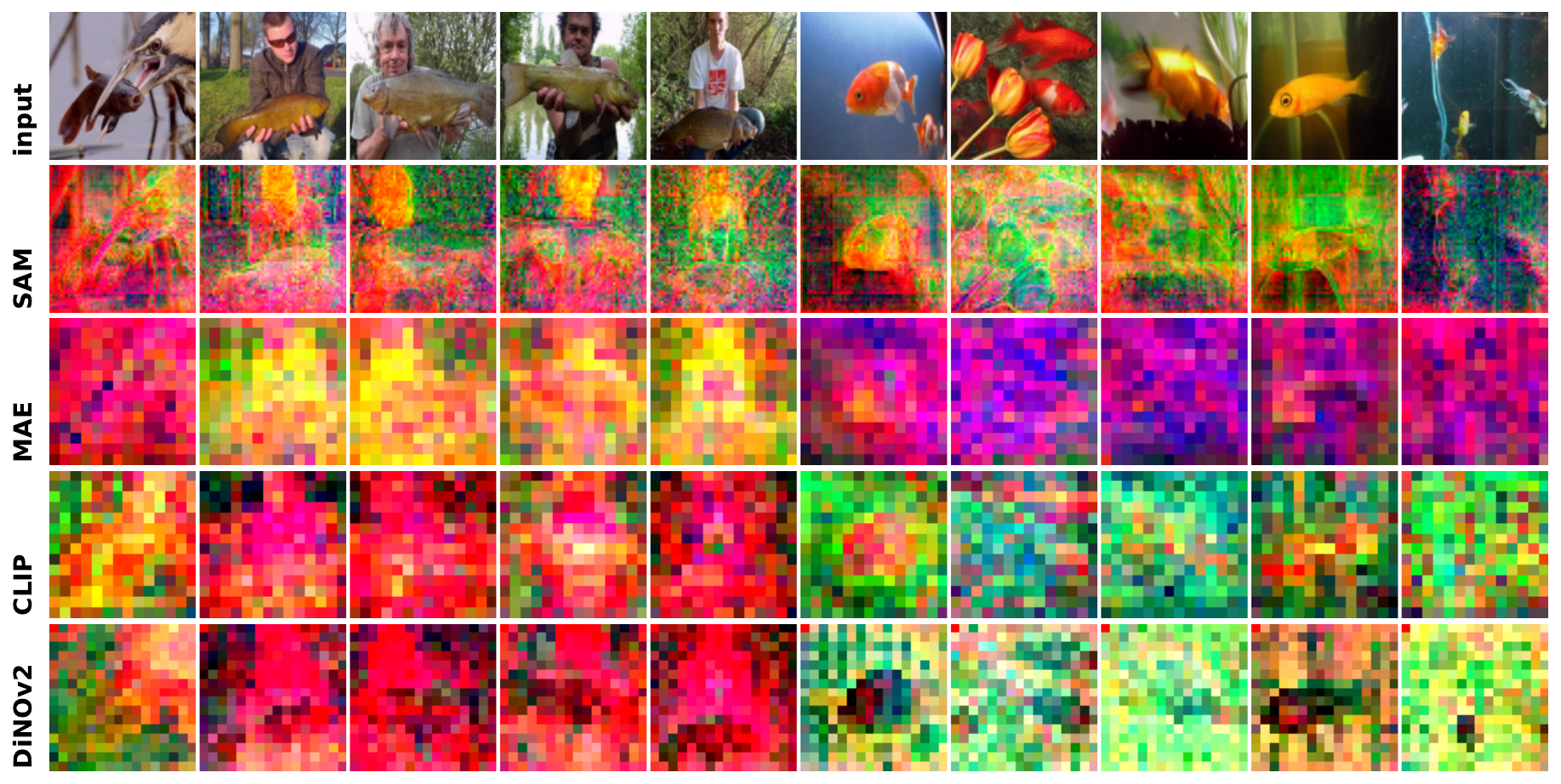}
        \caption{\textbf{FBA} (body)}
    \end{subfigure}

\caption{Top 3 selected channels for voxels in one brain ROI (methods in \Cref{sec:code}, findings in \Cref{sec:supp_channel}).}
\label{fig:supp_top_channels3}
\end{figure*}

\begin{figure*}[t]
    \centering

    \begin{subfigure}[b]{\textwidth}
        \includegraphics[width=\textwidth]{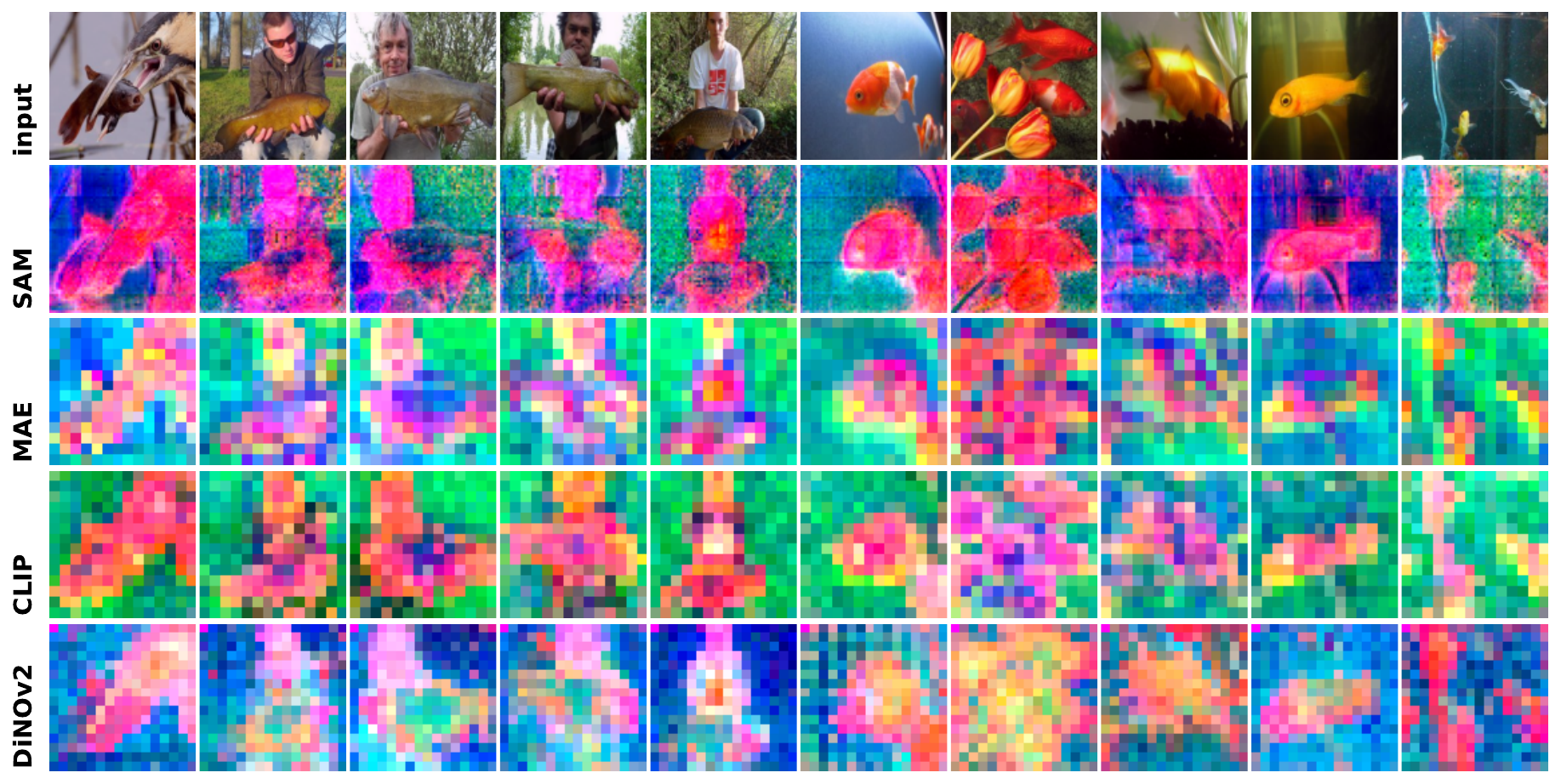}
        \caption{\textbf{OFA} (face)}
    \end{subfigure}

    \begin{subfigure}[b]{\textwidth}
        \includegraphics[width=\textwidth]{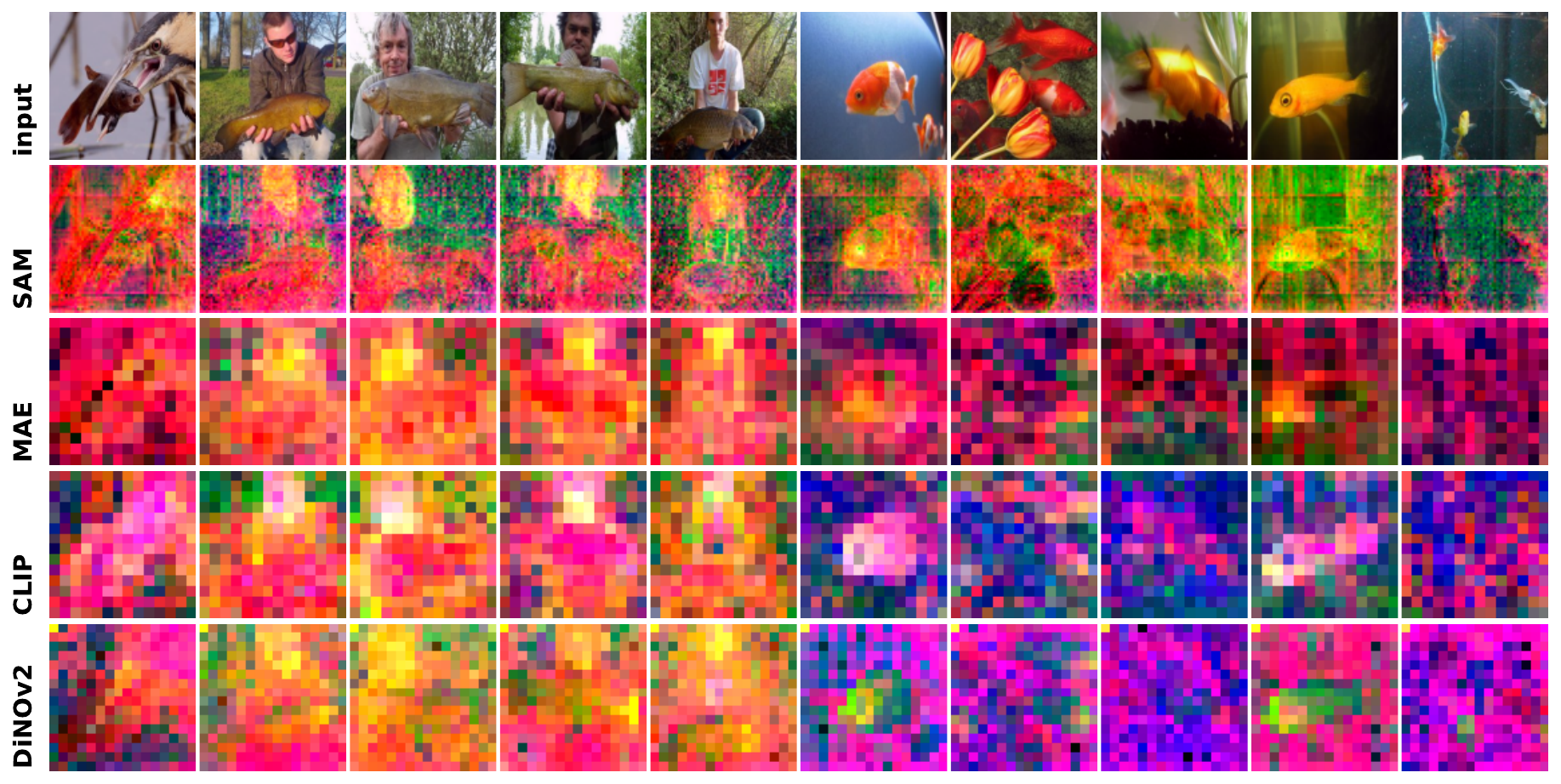}
        \caption{\textbf{FFA} (face)}
    \end{subfigure}

\caption{Top 3 selected channels for voxels in one brain ROI (methods in \Cref{sec:code}, findings in \Cref{sec:supp_channel}).}
\label{fig:supp_top_channels4}
\end{figure*}

\begin{figure*}[t]
    \centering

    \begin{subfigure}[b]{\textwidth}
        \includegraphics[width=\textwidth]{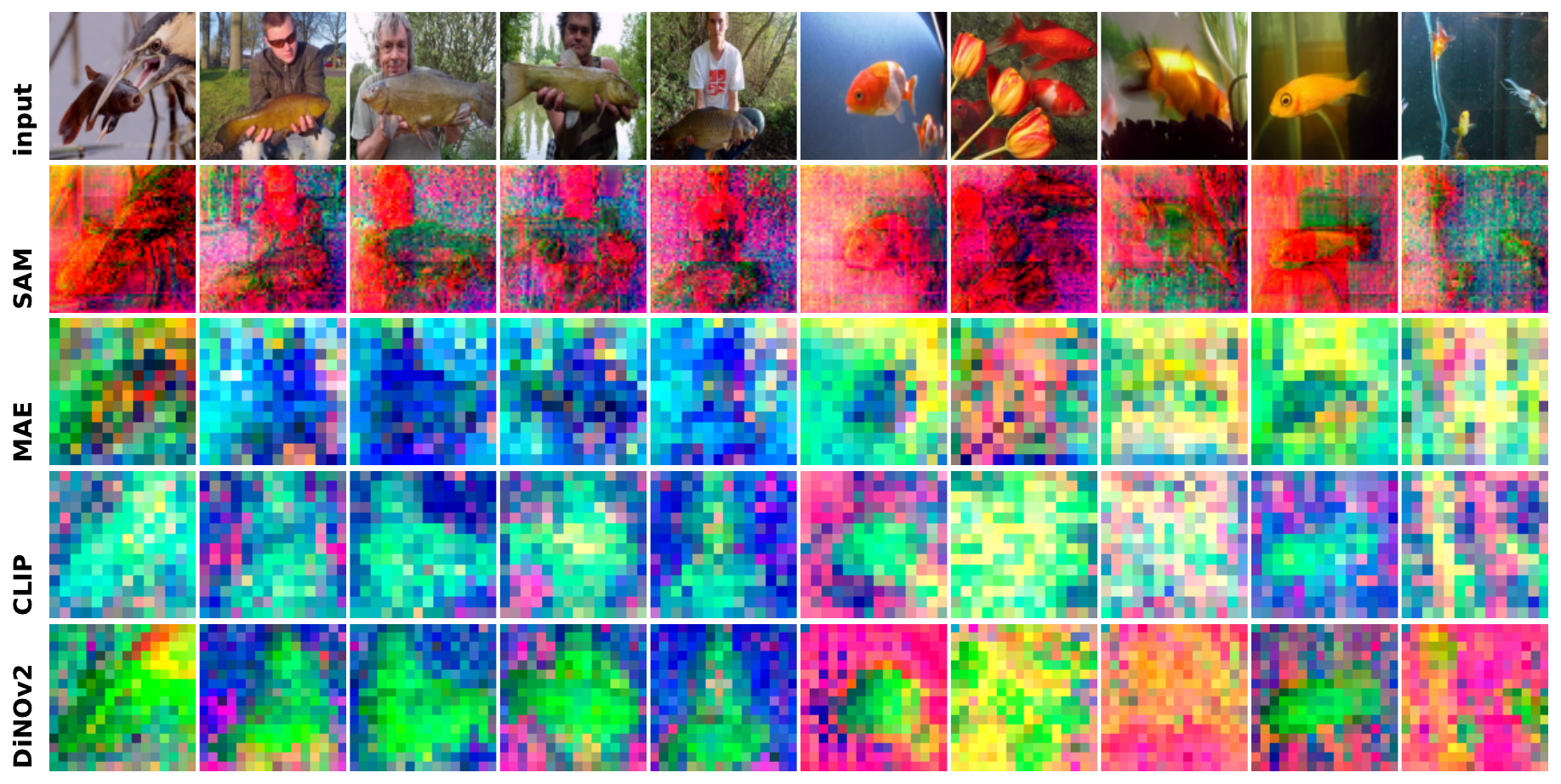}
        \caption{\textbf{OPA} (navigation)}
    \end{subfigure}

    \begin{subfigure}[b]{\textwidth}
        \includegraphics[width=\textwidth]{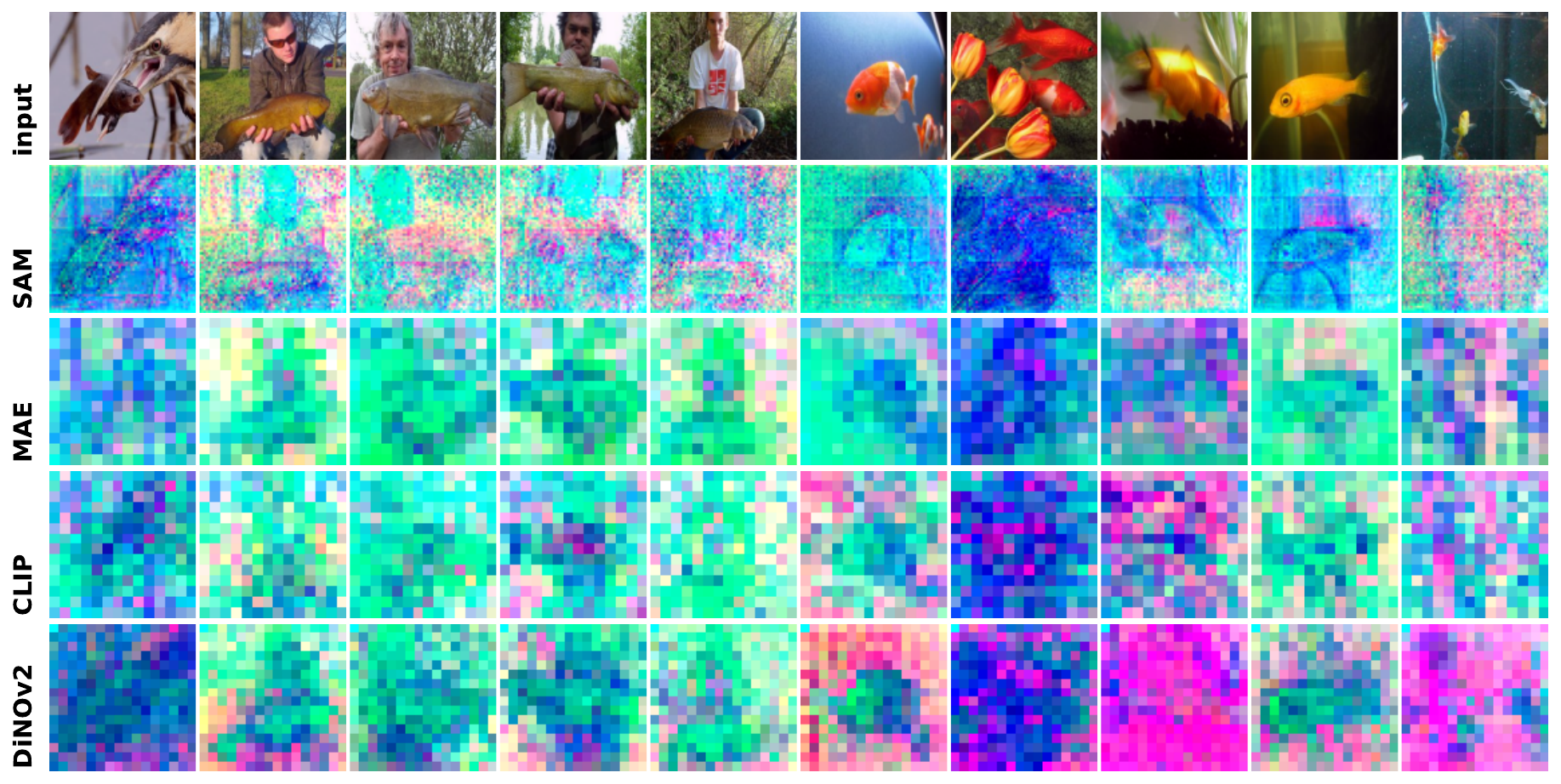}
        \caption{\textbf{PPA} (scene)}
    \end{subfigure}

\caption{Top 3 selected channels for voxels in one brain ROI (methods in \Cref{sec:code}, findings in \Cref{sec:supp_channel}).}
\label{fig:supp_top_channels5}
\end{figure*}

\begin{figure*}[t]
    \centering

    \begin{subfigure}[b]{\textwidth}
        \includegraphics[width=\textwidth]{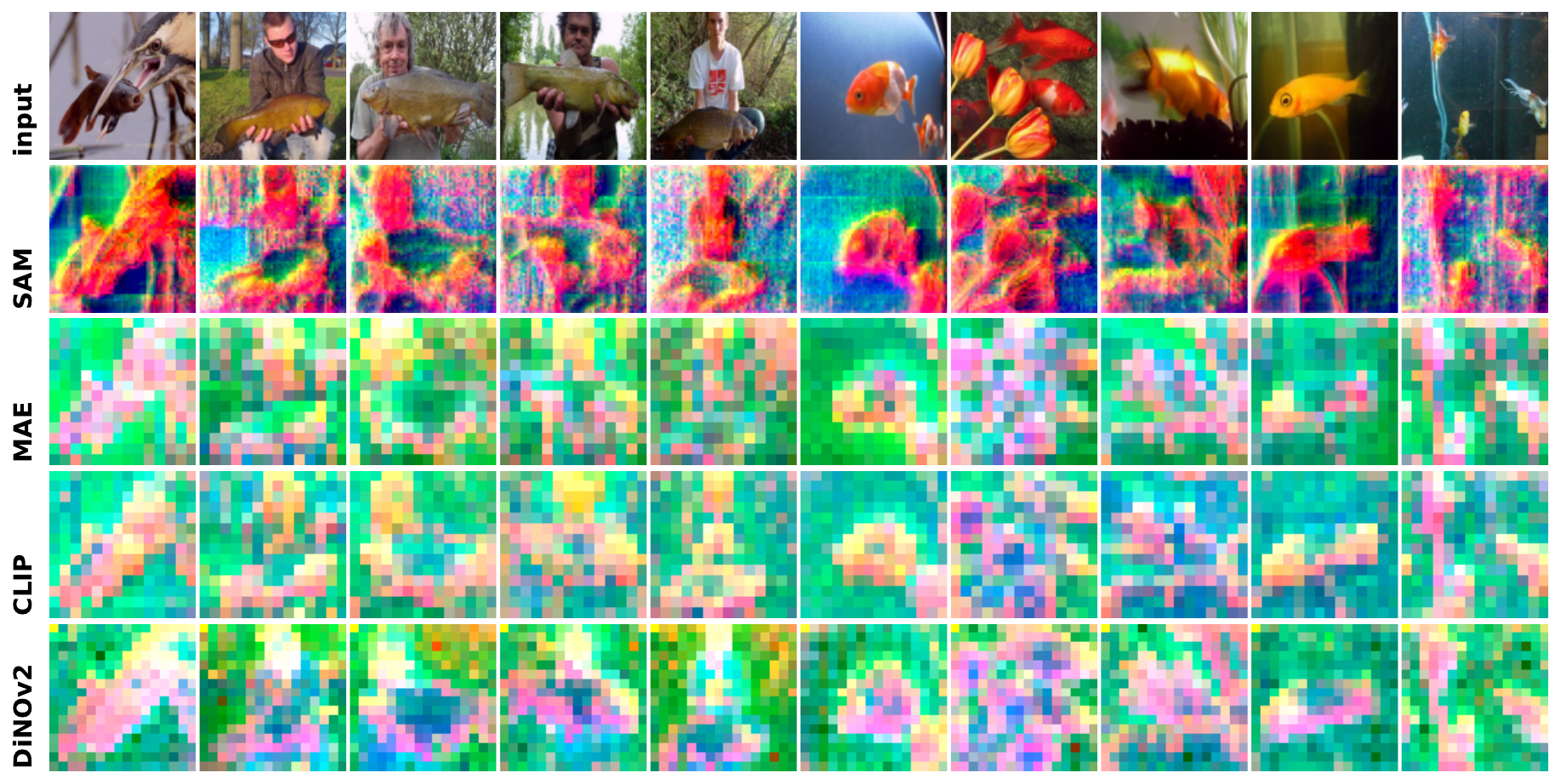}
        \caption{\textbf{OWFA} (words)}
    \end{subfigure}

    \begin{subfigure}[b]{\textwidth}
        \includegraphics[width=\textwidth]{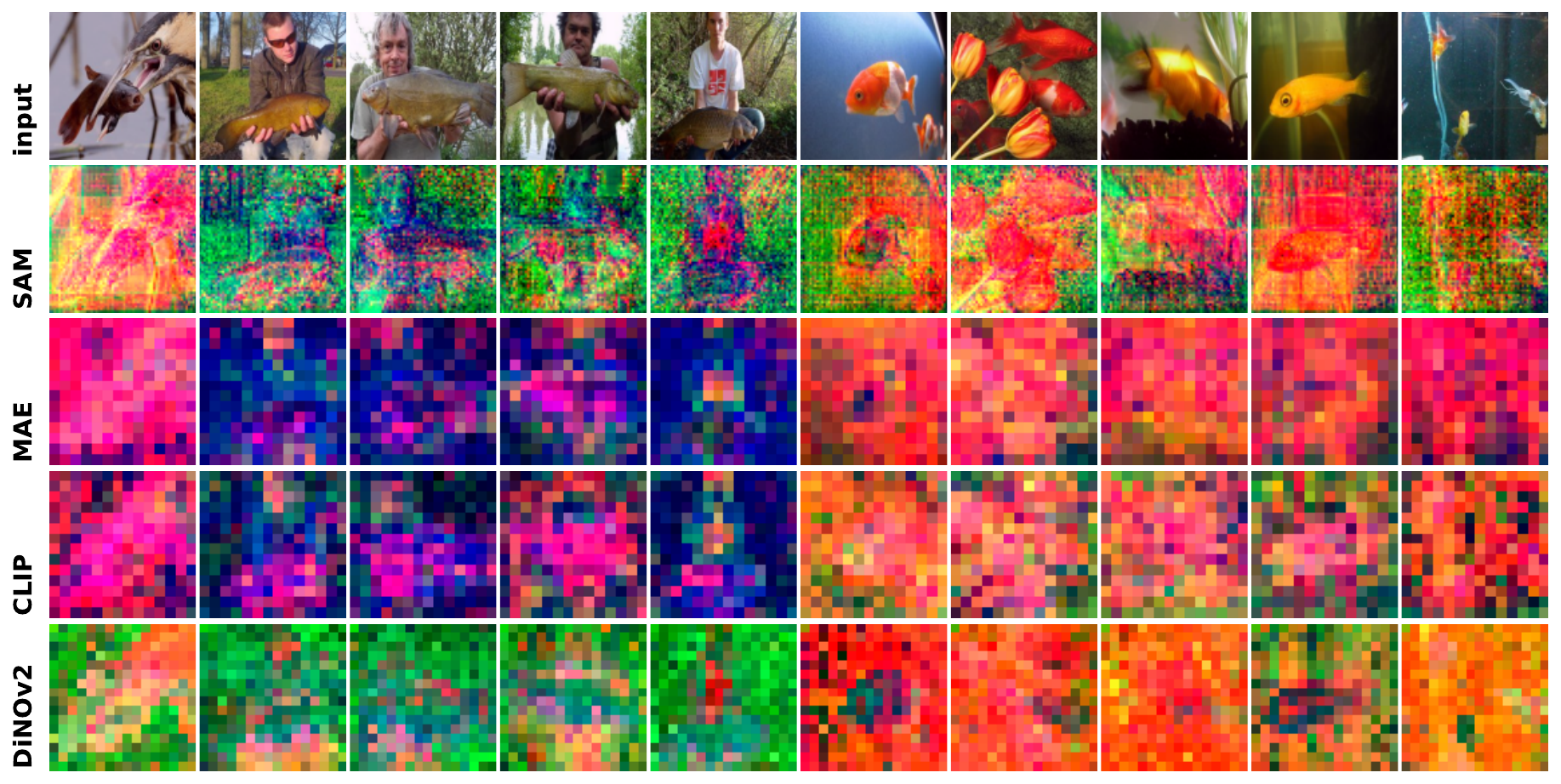}
        \caption{\textbf{VWFA} (words)}
    \end{subfigure}

\caption{Top 3 selected channels for voxels in one brain ROI (methods in \Cref{sec:code}, findings in \Cref{sec:supp_channel}).}
\label{fig:supp_top_channels6}
\end{figure*}

In the main results Section 
4.5
, we displayed the top selected channel in latent image space for V1 and FFA. In this appendix, we further display all ROIs: V1, V2, V3, V4, EBA, FBA, OFA, FFA, OPA, PPA, OWFA, and VWFA. Results are in \Cref{fig:supp_top_channels1} to \ref{fig:supp_top_channels6}. Methods and pseudocode are in \Cref{sec:code}.

\textbf{Comparing across all ROIs, we found}:
\begin{itemize}
    \item from V1 to V4, features become increasingly abstract
    \item EBA captures the body but not including face, EBA has two global patterns for whether a human is present
    \item FBA captures body including face, FBA has two global patterns for whether a human is present
    \item OFA segments out object and background
    \item FFA reacts to face centered at the eyeball, FFA has two global patterns for whether a human is present
    \item OPA segments out the central object but not peripheral objects
    \item PPA reacts uniformly to the whole image
    \item OWFA segments out the object and background
    \item VWFA has two global patterns for whether a human is present
\end{itemize}

\textbf{Comparing across all models, we found}: 
\begin{itemize}
    \item SAM's V1 to V4 features have finer segmentation of objects, SAM's EBA activates less on bodies, SAM's OPA does not capture the global layout, and SAM's OWFA does not capture abstract representation.
    \item MAE's EBA activates less on bodies, MAE's FBA activates more on bodies and faces. 
    \item CLIP's V4 has less activation on the central object, CLIP's EBA reacts to human bodies FBA reacts to animal bodies.
    \item DiNOv2's V3 showed grid structure, DiNOv2's EBA and FBA react to human bodies but less to animal bodies.
\end{itemize}

\newpage
\section{Methods Details}
\label{sec:supp_methods}
\subsection{NSD Data Processing Details}

We used the officially released GLMsingle \cite{prince_improving_2022} beta3 preparation of the data, the pre-processing pipeline consists of motion correlation, hemodynamic response function (HRF) selection for each voxel, nuisance regressor estimation via PCA, and finally, a general linear model (GLM) is fit independently for each voxel with selected HRF and nuisance regressor. In addition to the officially released pre-processing, we applied session-wise z-score to each voxel independently \cite{gifford_algonauts_2023}. We used the official release of the data on FreeSurfer average (brain surface) space. There's a total of 327,684 vertices for the whole cerebral cortex and sub-cortical regions, we only used 37,984 vertices in the visual cortex defined by the `nsdgeneral' ROI. We used coordinates of vertices in inflated brain surface space.

\subsection{Model and Visualization Pseudocode Code}
\label{sec:code}

\paragraph{Model (FactorTopy) Pseudocode} \Cref{list:code1} presents a PyTorch-style pseudocode for our main \emph{FactorTopy} model. The factorized selectors in Equation 1 are implemented as separate MLPs with \texttt{tanh}, \texttt{softmax}, and \texttt{sigmoid} activation functions, respectively. \texttt{pe} is sinusoidal positional encoding.

\definecolor{LightGray}{gray}{0.95}
\usemintedstyle{vs}
\begin{listing}[!ht]
\begin{minted}[
frame=lines,
framesep=2mm,
baselinestretch=1.0,
bgcolor=LightGray,
fontsize=\footnotesize,
]{python}
### FactorTopy model ###
# x: Tensor, [B, 3, 224, 224], B := batch size
# coord: Tensor, [N, 3], N := number of voxels

## 1. backbone
local_tokens, global_tokens = backbone(x)
# local_tokens: dict, {layer: [B, C, H, W]}
# global_tokens: dict, {layer: [B, C]}

## 2a. downsample, (H, W) -> (8, 8)
local_tokens = downsample(local_tokens)

## 2b. layer-unique bottleneck, C -> D
for layer in layers:
    local_tokens[layer] = bottle_neck[layer](
        local_tokens[layer]) # [B, D, 8, 8]
    global_tokens[layer] = bottle_neck[layer](
        global_tokens[layer]) # [B, D]

## 3. multi-selectors
space = tanh(space_mlp(pe(coord))) # [N, 2]
layer = softmax(layer_mlp(pe(coord))) # [N, L]
scale = sigmoid(scale_mlp(pe(coord))) # [N, 1]

## 4. get v
# get v_local
v_local = bilinear_interpolate(
    local_tokens, space) # [B, N, D, L]
# sum v_local and v_global
v_global = stack(global_tokens).repeat(1, N) 
    # [B, N, D, L]
v = v_local * (1-scale) + v_global * scale 
    # [B, N, D, L]
# sum over layers
v = (v * layer).sum(dim=-1) # [B, N, D]

## 5. voxel-specific linear regression
y = (v * w).mean(dim=-1) + b # [B, N]
\end{minted}
\caption{PyTorch-style pseudocode of our methods FactorTopy.}
\label{list:code1}
\end{listing}

\paragraph{Channel Clustering} We clustered selected channels (linear regression weights $\bm{w}$) into 20 clusters in primary results Section 
4.5
and \Cref{sec:supp_channel}. The procedure for clustering is: 1) use kernel trick $\bm{\bar{w}} = \bm{w}^{T} \bm{w}$, $\bm{w} \in \mathbb{R}^{D \times N}$ where $D$ is channel dimension, $N$ is the number of voxels. 2) use k-means clustering on $\bm{\bar{w}}$ with euclidean distance, k=1000. 3) use Agglomerative Hierarchical Clustering on the k-means centroids, euclidean distance, and Ward's method, iterative merge until resulting in 20 clusters.

\paragraph{Channel Visualization Pseudocode} In the main results Section 
4.5
and \Cref{sec:supp_channel}, we visualized the top selected channel in image space for brain ROIs. The motivation for the image space visualization is to plot the top selected channel for an ROI of voxels; voxels' linear regression weights are functioning as `\textit{channel selection}' that answers ``\textit{which channels best predict my brain response?}''. 

In a single voxel case, we can 1) obtain local tokens $\mathbb{R}^{D \times H \times W}$ by summing features $\mathbb{R}^{L \times D \times H \times W}$ with layer selector weight $\bm{\hat{\omega}}_i \in \mathbb{R}^{L}$, 2) sum local tokens $\mathbb{R}^{D \times H \times W}$ with regression weight $w_i \in \mathbb{R}^{D}$, output a greyscale image $\mathbb{R}^{1 \times H \times W}$. 

To extend to an ROI of voxels, we 1) summed local tokens from all layer $\mathbb{R}^{L \times D \times H \times W}$ by ROI-average layer selector weights $\bm{\hat{\omega}}_* = \frac{1}{|roi|} \sum_{i\in roi} \bm{\hat{\omega}}_i$, where $|roi| = N'$, output $\mathbb{R}^{D \times H \times W}$, 2) applied PCA to reduce linear regression weights $\mathbb{R}^{D \times N'}$ along the dimension of number of voxels $N'$, output $\mathbb{R}^{D \times 3}$, 3) applied top 3 PC weights to local tokens to reduce the channel dimension $D$ of local tokens, output RGB image $\mathbb{R}^{3 \times H \times W}$. A complete pseudocode is in \Cref{list:code2}.

\definecolor{LightGray}{gray}{0.95}
\usemintedstyle{vs}
\begin{listing}[!ht]
\begin{minted}[
frame=lines,
framesep=2mm,
baselinestretch=1.0,
bgcolor=LightGray,
fontsize=\footnotesize,
]{python}
### top channel visualization ###
# x: Tensor, [3, 224, 224], batch size is 1 
# coord: Tensor, [N, 3], N := number of voxels
# roi_mask: Tensor, [N], boolean, sum = N'

## 1. backbone
local_tokens, global_tokens = backbone(x)
# local_tokens: dict, {layer: [C, H, W]}

## 2b. layer-unique bottleneck, C -> D
for layer in layers:
    local_tokens[layer] = bottle_neck[layer](
        local_tokens[layer]) # [D, H, W]
local_tokens = stack(local_tokens) 
    # [L, D, H, W]

## 3. multi-selectors
layer = softmax(layer_mlp(pe(coord[roi_mask]))) 
    # [N', L]

## 4. sum local_tokens by ROI
layer_weights = layer.mean(0) # [L]
local_tokens = sum(layer_weights * local_tokens) 
    # [D, H, W]

## 5. PCA on linear regression weights
_w = w[:, roi_mask] # [D, N']
_pc_w = pca(_w) # [D, 3]

## 6. RGB image
image = _pc_w.t() @ local_tokens # [3, H, W]
\end{minted}
\caption{PyTorch-style pseudocode for channel visualization.}
\label{list:code2}
\end{listing}







    

\subsection{Training Details}
 
\paragraph{Hardware and Wall-clock} We conducted experiments on a mixture of Nvidia A6000 and RTX4090 GPUs. Features of the pre-trained model are pre-computed and cached. We used bottleneck dimension $D=128$; increasing $D$ will significantly increase computation intensity as the number of brain voxels (vertices) is large (37,984). A full data (22K data samples) model converges in 1 to 3 hours for 12 to 40 layer models respectively. A partial data (3K data samples) 12-layer model converges in 30 minutes.

\paragraph{Optimizer and Training Recipe} For training brain encoding models, we used the AdamW optimizer, batch size 8, learning rate 1e-3, betas (0.9, 0.999), and weight decay 1e-2. We trained for 1,000 steps per epoch, with an early stopping of 20 epochs. Models reached maximum validation score at 40,000 to 60,000 steps, and the multi-selectors in our methods became stable after 10,000 steps. For each model, we saved the top 10 validation checkpoints and used ModelSoup \cite{wortsman_model_2022-1} to average the best validation checkpoints and greedily optimize the score on the test set. We did not apply any data augmentation, existing data augmentation is not useful for brain encoding because the prediction target (brain) is not transformed alongside the input image. 

\begin{figure}[H]
    \centering
    \includegraphics[width=0.99\linewidth]{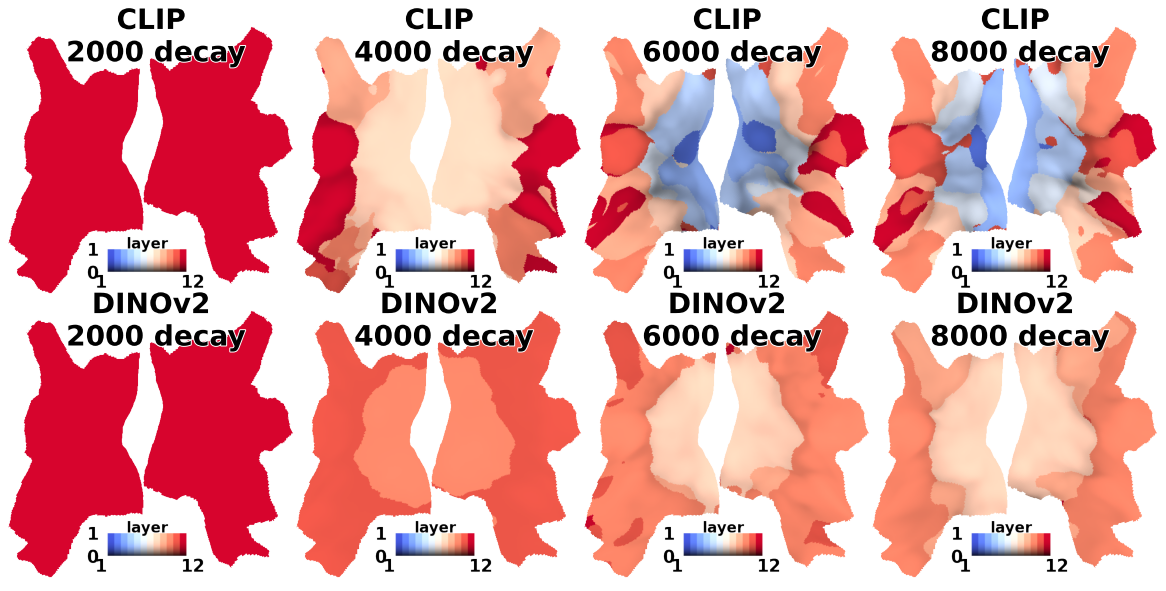}
    \captionof{figure}{Layer selector output w.r.t. regularization decay total steps, number is total steps.}
    \label{fig:supp_decay}
\end{figure}

\paragraph{Loss and Regularization} We used smooth L1 loss (beta=0.1) with an additional decaying regularization term on layer selector $\bm{\hat{\omega}}^{layer}$. The motivation for regularization is the use of softmax activation function in layer selector MLP leads to vanishing gradient at one-hot output, layer selector converges to a singular selection for all voxels (\Cref{fig:supp_decay}) if with insufficient regularization,

\begin{equation}
\begin{aligned}
    loss_{reg} &= - \frac{1}{N} \sum_{i=1}^{N} (\frac{\sum_{l=1}^{L} \hat{\omega}_{i,l}^{layer} \log \hat{\omega}_{i,l}^{layer}}{\sum_{l=1}^{L} \frac{1}{L} \log \frac{1}{L}}) \\
    decay &= max(0, 1-\frac{step_{i}}{step_{total}}) \\
    loss &= loss_{l1} + \lambda * loss_{reg} * decay
\end{aligned} 
\end{equation}

\noindent
$N$ is number of voxels, $L$ is number of layers, $\lambda$ is set to 0.1, $step_i$ is the current training step, $step_{total}$ is total steps of linear decaying. In \Cref{tab:supp_decay_steps}, we ran a grid search of $step_{total}$ and concluded to use a total decay step of 6000; the same total decay step is set for all models. \Cref{fig:supp_decay} shows the resulting brain-to-layer mapping when trained with less regularization decay steps. When trained with insufficient regularization, layer selection converges to a local minimum (\Cref{tab:supp_decay_steps}) of selecting only the last layer (\Cref{fig:supp_decay}).

It's worth noting that it's possible to optimize the performance score by searching optimal decay steps for every model. However, we use entropy as a confidence measurement (Equation 2) in our experiments. The regularization term impacts the resulting confidence value, thus, we set the same total decay step (6000) for all models to avoid unfair comparison of confidence measurement.

\begin{table}[H]
\centering
\resizebox{0.99\linewidth}{!}{
\begin{tblr}{
  cell{1}{2} = {c=4}{c},
  hline{1,5} = {-}{0.08em},
  hline{3} = {-}{0.05em},
}
 & \textbf{Decay Steps, Brain Score $R^2$ $\uparrow$ ($\pm$ 0.001)} & & & \\
\textbf{Model} \quad\quad\quad & 2000 \quad\quad & 4000 \quad\quad & 6000 \quad\quad & 8000 \\
CLIP & 0.093 & 0.128 & \textbf{\textit{0.131}} & \textbf{0.132} \\
DiNOv2 & 0.113 & \textbf{{0.126}} & \textbf{{0.126}} & \textbf{\textit{0.125}} \\
\end{tblr}
}
\caption{Performance score w.r.t. total decay steps for regularization term. Grid search with CLIP and DiNOv2 base size 12-layer model. Average of 3 runs.}
\label{tab:supp_decay_steps}
\end{table}


\clearpage
\newpage
\section{Related Work: State-of-the-art Methods}
\label{sec:sota}
In the main text Section \ref{sec:performance}, we compared our methods against the state-of-the-art methods' most salient design, but not their original methods. In this appendix, we discuss the competition-winning approaches in detail and explain the motivation for comparing their most salient design but not their original methods.

\paragraph{Experiment Setting} Our experiment setting is different from the competition-winning methods. They build an ensemble of ROI-unique models, there's less demand for voxel-wise feature selection in ROI-unique models because voxels in the same ROI select similar features. However, we build one all-ROI model that covers all visual brain voxels, and the local similarity and global diversity of voxels emphasized the importance of factorized and topology-constrained feature selection introduced in this work. Overall, existing work use pre-defined ROIs and ensemble of ROI-unique models, we build one all-ROI model.

\begin{figure}[h]
\vspace{-0mm}
    \centering
    \includegraphics[width=0.99\linewidth]{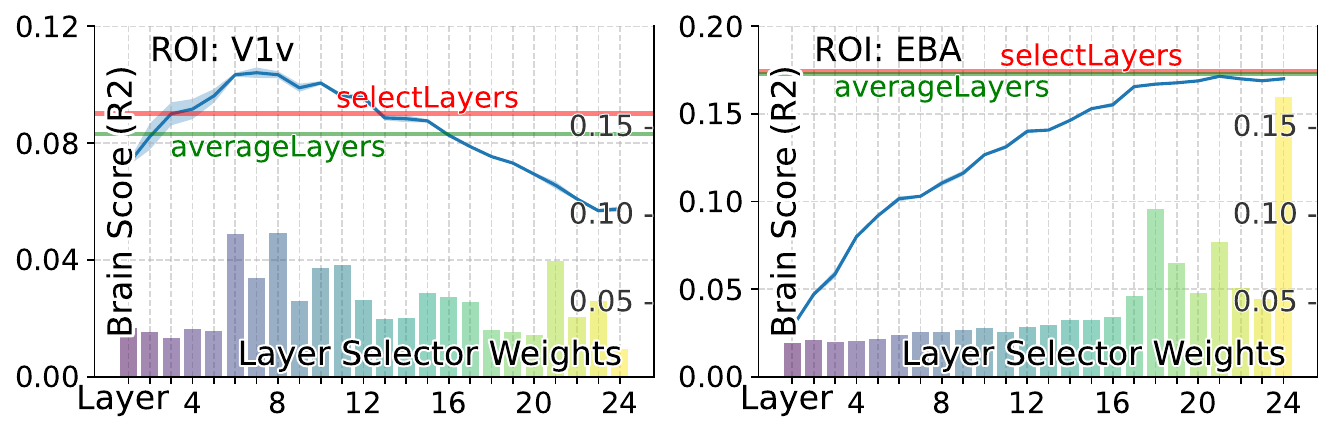}
    \vspace{-3mm}
    \caption{Grid search matched layer selector weights. Right y-axis is selector weights, and left y-axis is prediction score. Blue curve: per-layer model grid search score; Red line: score with layer selection; Green line: score with uniform layer average. Grid search with class token only.}
    \label{fig:grid_layers}
\vspace{-2mm}
\end{figure}

Past Algonauts competition-winning methods used an ensemble with a grid search of layers \cite{cichy_algonauts_2021, gifford_algonauts_2023}.   The best single-layer model outperforms averaging or concatenating multiple layers.  We aim to build a single all-ROI model that dynamically selects layers for voxels in every ROI.   In Figure \ref{fig:grid_layers}, we verified that our layer selector weights matched the grid search score of single-layer models. 

\newpage
 
\paragraph{Algonauts 2021 winner, patchToken} The Algonauts 2021 challenge was hosted with another 3T fMRI dataset \cite{lahner_bold_2023} but not NSD. The Algonauts 2021 dataset lacks the high data quality that NSD has, the lower data quality limited the effect of novel model building. The Algonauts 2021 competition winners (the top 3 methods are summarized in supplementary of \cite{lahner_bold_2023}) used primitive methods that do not select unique features for each voxel but compress the flattened patches into one feature vector for all voxels. All voxels use the same feature vector.
The most salient design in Algonauts 2021 is the patch compression module, patchToken, we re-implemented their methods (Table 3, patchToken) and found that patchToken methods achieved sub-optimal performance on the high-quality dataset NSD.

\vspace{-2mm}
\paragraph{GNet by NSD} Before the Algonauts 2023 challenge, for the NSD dataset, the commonly used state-of-the-art method is GNet introduced by the NSD authors \cite{allen_massive_2022}. GNet introduced a layer-specific spatial pooling field for each voxel, which is a non-factorized and non-topology-constrained feature selection for each voxel. The original GNet was an end-to-end CNN trained from scratch, later studies \cite{conwell_what_2023} swapped the image backbone model with frozen state-of-the-art pre-trained ViT models to increase the performance. 
In our comparison we used a frozen CLIP XL model for all the models, so it's not the original GNet. 
The most salient design in GNet is the layer-specific spatial pooling field, we re-implemented the spatial pooling field design and compared it with our methods (Table 3, GNetViT). Notably, GNetViT requires quadratic memory and computation because of the unique $L\times H\times W$ spatial pooling field for each voxel.

\vspace{-2mm}
\paragraph{Algonauts 2023 first place} Our methods is an extension of the Algonauts 2023 winning methods Memory Encoding Model (Mem) \cite{yang_memory_2023}. Mem used topology constraints but only partially factorized the feature selection (they are missing the scale axis). In our methods, we further introduced fully factorized feature selection. There are some major differences between our work and their settings: 1) We only consider one image as input, Mem used extra information including past 32 images, behavior response, and time information, extra information led to a shocking 10\% challenge score boost. 2) We build one single all-ROI model, Mem builds an ensemble of ROI-unique models. 3) We ran the training only once, Mem used dark knowledge distillation and ran the training twice. 4) We only used voxels in the visual brain, Mem additionally used voxels outside the visual brain to increase data samples. 5) We only used one subject for training, Mem trained a shared backbone for all 8 subjects. Mem's is partially factorized (without scale axis) and topology-constrained feature selection, we included Mem's most salient design in our ablation study in Table 3 ``- no scale sel''. 

\paragraph{Algonauts 2023 second place} For the second place winning methods of the Algonauts 2023 challenge \cite{adeli_predicting_2023}, the most important factors to their winning are: 1) they used extra information including behavior response and time information, which led to a 4\% challenge score boost, 2) they built ROI-unique models ensemble, and 3) they trained on 8 subjects. For the methods, they used a Detection Transformer (DETR) style attention mask with ROIs as queries. Their feature selection is voxel-shared but ROI-specific and also image-specific, not factorized or topology constrained. The DETR-style attention mask requires quadratic computation resources, their methods is possible to run for 36 ROIs as queries but impossible for 37,984 voxels as queries. We did not include the transformer methods in the comparison because their methods fundamentally rely on pre-defined ROIs and are unable to scale up to voxels. The closest comparison to this transformer method is GNetViT. 

\paragraph{Algonauts 2023 third place} For the third place winning methods of the Algonauts 2023 challenge \cite{nguyen_algonauts_2023}, the most important factors that contributed to their winning are: 1) they ensembled 6 backbone models, 2) they pre-trained models on all ROI and all subjects, then fine-tuned ROI-unique models for each subject, and 3) they used a bag of training tricks. Their method used the same feature vector for voxels in the same ROI, similar to the patchToken methods in Algonauts 2021. However, it remains unclear how they compressed the $L\times C \times H \times W$ feature into one feature vector. We did not include this method in the comparison because the feature compression module is unclear, the closest comparison is classToken and patchToken.

\subsection{Performance and Complexity} 
\label{sec:performance}

Previous state-of-the-art brain encoding approaches made diverse choices on image encoders and feature selections. We re-implemented them to avoid unfair comparison by keeping their most salient design choices but swapping them in standard components. We used CLIP-XL \cite{ilharco_openclip_2021, gadre_datacomp_2023} backbone for the image encoder for all methods.

There are three distinct types of feature selections. \textbf{1)} The simplest way is to leverage the class tokens, classToken, by taking it from each layer, $\mathbb{R}^{L \times C}$, applies a layer-unique transformation to $\mathbb{R}^{L \times D}$, and average pools across the layers to obtain a $\mathbb{R}^{D}$ feature vector. \textbf{2)} The second way, patchComp, extracts information from the patch image token, allowing finer pixel region selection: flattened features first along the spatial dimension $H \times W$ for each layer and fed $\mathbb{R}^{C \times H \times W}$ to a layer-unique-MLP that compressed it to a $\mathbb{R}^{D}$ feature vector.  \textbf{3)} Finally, in the style of GNet \cite{allen_massive_2022}, we construct a layer-specific 2D selection mask to pool $\mathbb{R}^{L \times D \times H \times W}$ into a vector of $\mathbb{R}^{L \times D}$, followed by pooling layers to obtain a $\mathbb{R}^{D}$ feature vector.  In the ablation study of our network (\emph{FactorTopy}), we created several versions each by replacing one of the factorized selectors in layer, space, and scale and with average pooling.  We also created a more robust version by sampling three times in the space selection.
Comparison results are reported in Table \ref{tab:space_selection}.

\begin{table}[h]
\vspace{1mm}
\begin{center}
\resizebox{0.99\linewidth}{!}{
\begin{tblr}{
  row{11} = {fg=Gray},
  cell{1}{4} = {c=4}{},
  cell{1}{1} = {r=2}{},
  cell{1}{2} = {r=2}{},
  cell{1}{3} = {r=2}{},
  cell{3}{2} = {fg=Gray},
  cell{3}{3} = {fg=Gray},
  cell{4}{2} = {fg=Gray},
  cell{4}{3} = {fg=Gray},
  cell{6}{2} = {fg=Gray},
  cell{6}{3} = {fg=Gray},
  cell{7}{2} = {fg=Gray},
  cell{7}{3} = {fg=Gray},
  cell{8}{2} = {fg=Gray},
  cell{8}{3} = {fg=Gray},
  cell{9}{2} = {fg=Gray},
  cell{9}{3} = {fg=Gray},
  cell{10}{2} = {fg=Gray},
  cell{10}{3} = {fg=Gray},
  cell{11}{2} = {fg=Gray},
  cell{11}{3} = {fg=Gray},
  cell{12}{2} = {fg=Gray},
  cell{12}{3} = {fg=Gray},
  hline{1,12} = {-}{0.1em},
  hline{2} = {4-7}{0.05em},
  hline{3} = {-}{0.05em},
}
\textbf{Method}      & \textbf{Time$^{\ddag}$} & \textbf{MACs} & \textbf{Brain Score $R^2$ $\uparrow$ ($\pm$ 0.001)} &                &                &                \\
                     &               &                 & \textbf{all}                 & \textbf{V1v}   & \textbf{V3v}   & \textbf{EBA}   \\
classToken                  & $\times$1          & $\times$1            & 0.100                        & 0.085          & 0.075          & 0.173          \\
patchToken \cite{lahner_bold_2023}                 & $\times$1          & $\times$1            & 0.122                        & 0.176          & 0.163          & 0.165          \\
GNetViT  \cite{allen_massive_2022}                & $\times$94          & $\times$17             & 0.124                        & 0.174          & 0.146          & 0.174          \\
\textbf{FactorTopy (Ours)} & $\times$3 & $\times$1.2 & \textbf{0.132} & \textbf{0.205} & \textbf{0.179} & \textbf{0.175} \\
 - w/o topology & $\times$3 & $\times$1.4 & 0.130 & 0.197 & 0.176 & 0.174 \\
 - no layer sel & $\times$3 & $\times$1.2 & 0.125 & 0.181 & 0.162 & 0.174 \\
 - no space sel & $\times$3 & $\times$1.2 & 0.117 & 0.094 & 0.089 & 0.175 \\
 - no scale sel & $\times$3 & $\times$1.2 & 0.131 & 0.201 & 0.177 & 0.175 \\
 + multiple sample & $ \times$7 &$\times$1.6 & 0.134 & 0.207 & 0.182 & 0.176 \\
\end{tblr}
}
\end{center}
\vspace{-6mm}
\caption{\textbf{Performance Ablation}. Average of 3 runs. $^\ddag$: wall-clock.}
\label{tab:space_selection}
\vspace{-4mm}
\end{table}

\newpage

\twocolumn[{%
\renewcommand\twocolumn[1][]{#1}%
\vspace{-4mm}

    \centering
    \captionsetup{type=figure}
    \includegraphics[width=\linewidth]{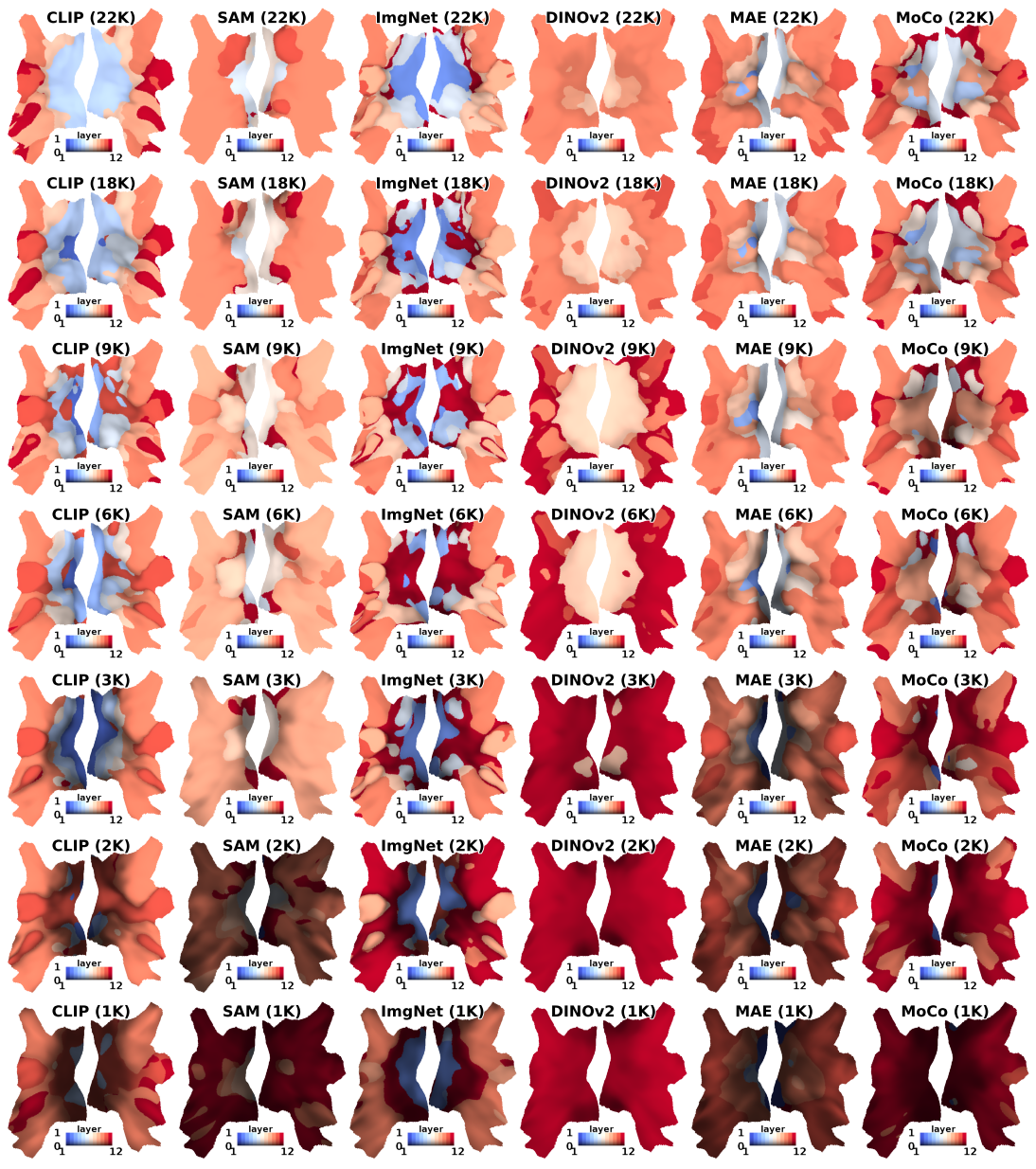}
    \captionof{figure}{Brain-to-Network alignment trained with limited data samples. Base size models, number of samples marked in brackets.}
    \label{fig:supp_num_data}

\vspace{8mm}
}]

\section{Limited Training Samples}
\label{sec:supp_less_data}

\label{sec:num_data}

\begin{figure}[h]
    \vspace{-2mm}
    \centering
    \includegraphics[width=0.9\linewidth]{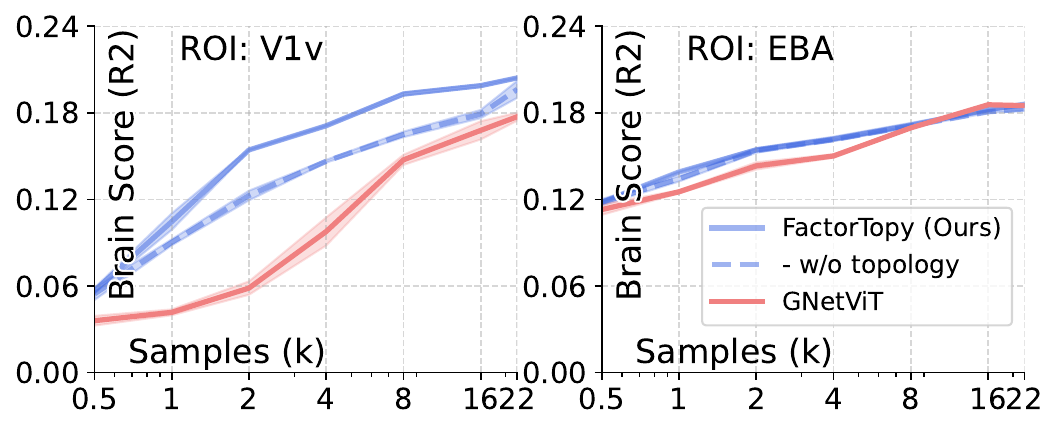}
    \vspace{-3mm}
    \caption{Performance w.r.t. training data sample, in log scale.}
    \label{fig:num_data}
    \vspace{-3mm}
\end{figure}

Practical use of our brain-to-network mapping tool for network visualization requires our network to be trained efficiently.  Using data scaling experiments shown in Figure \ref{fig:num_data} and \Cref{fig:supp_num_data}, we conclude that teaching our model with 3K sample images (30 minutes on RTX4090) offers a good trade-off.   Our topological constraints and factorized feature selection (\emph{FactorTopy}) scales better to less training data.

\newpage
\clearpage
\pagestyle{fancy}
\fancyhead{}
\fancyhead[RO,LE]{\textbf{Top Channels for V1 V2}}

\clearpage
\pagestyle{fancy}
\fancyhead{}
\fancyhead[RO,LE]{\textbf{Top Channels for V3 V4}}

\clearpage
\pagestyle{fancy}
\fancyhead{}
\fancyhead[RO,LE]{\textbf{Top Channels for EBA FBA}}

\clearpage
\pagestyle{fancy}
\fancyhead{}
\fancyhead[RO,LE]{\textbf{Top Channels for OFA FFA}}

\clearpage
\pagestyle{fancy}
\fancyhead{}
\fancyhead[RO,LE]{\textbf{Top Channels for OPA PPA}}

\clearpage
\pagestyle{fancy}
\fancyhead{}
\fancyhead[RO,LE]{\textbf{Top Channels for OWFA VWFA}}

\end{document}